\documentclass[10pt,journal,compsoc]{IEEEtran}
\usepackage{amsmath,amsfonts}
\usepackage{algorithmic}
\usepackage{algorithm}
\usepackage{array}
\usepackage[caption=false,font=normalsize,labelfont=sf,textfont=sf]{subfig}
\usepackage{textcomp}
\usepackage{stfloats}
\usepackage{url}
\usepackage{verbatim}
\usepackage{graphicx}
\usepackage{cite}

\usepackage{graphicx}
\usepackage{amsthm}
\usepackage{multicol}
\usepackage{multirow}
\usepackage{booktabs}
\usepackage{pifont}
\usepackage{xcolor}
\usepackage{color}
\usepackage[backref]{hyperref}
\usepackage{ragged2e}
\hypersetup{
hidelinks,
colorlinks=true,
linkcolor=black,
citecolor=black
}
\begin{document}

\title{On the Element-Wise Representation and Reasoning in Zero-Shot Image Recognition: \\
A Systematic Survey}

\author{
Jingcai~Guo,~\IEEEmembership{Member,~IEEE,}
Zhijie~Rao,
Zhi~Chen,~\IEEEmembership{Member,~IEEE,}
Song~Guo,~\IEEEmembership{Fellow,~IEEE,}
Jingren~Zhou,~\IEEEmembership{Fellow,~IEEE,}
and Dacheng~Tao,~\IEEEmembership{Fellow,~IEEE}
        % <-this % stops a space
%\thanks{Manuscript received October 12, 2024.}
\thanks{Jingcai Guo and Zhijie Rao are with Department of Computing, The Hong Kong Polytechnic University, Hong Kong SAR, China (e-mail: jc-jingcai.guo@polyu.edu.hk, zhijie96.rao@connect.polyu.hk).}
%\thanks{Zhijie Rao is with Department of Computing, The Hong Kong Polytechnic University, Hong Kong SAR, China (e-mail: zhijie96.rao@connect.polyu.hk).}
\thanks{Zhi Chen is with School of Electrical Engineering and Computer Science, The University of Queensland, Brisbane QLD 4072, Australia (e-mail: zhi.chen@uq.edu.au).}
\thanks{Song Guo is with Department of Computer Science and Engineering, Hong Kong University of Science and Technology, Hong Kong SAR, China (e-mail: songguo@cse.ust.hk).}
\thanks{Jingren Zhou is with Alibaba Group, Hangzhou 311121, China (e-mail: jingren.zhou@alibaba-inc.com).}
\thanks{Dacheng Tao is with College of Computing and Data Science, Nanyang Technological University, Singapore (e-mail: dacheng.tao@ntu.edu.sg).}
}

% The paper headers
\markboth{Under Review}%
{Guo \MakeLowercase{\textit{et al.}}: On the Element-Wise Representation and Reasoning in Zero-Shot Image Recognition: A Systematic Survey}

%\IEEEpubid{0000--0000/00\$00.00~\copyright~2021 IEEE}
% Remember, if you use this you must call \IEEEpubidadjcol in the second
% column for its text to clear the IEEEpubid mark.

%\maketitle
\IEEEtitleabstractindextext{%
\begin{abstract}\justifying
Zero-shot image recognition (ZSIR) aims to recognize and reason in unseen domains by learning generalized knowledge from limited data in the seen domain. The gist of ZSIR is constructing a well-aligned mapping between the input visual space and the target semantic space, which is a bottom-up paradigm inspired by the process by which humans observe the world. In recent years, ZSIR has witnessed significant progress on a broad spectrum, from theory to algorithm design, as well as widespread applications. However, to the best of our knowledge, there remains a lack of a systematic review of ZSIR from an element-wise perspective, \textit{i.e.}, learning fine-grained elements of data and their inferential associations. To fill the gap, this paper thoroughly investigates recent advances in element-wise ZSIR and provides a sound basis for its future development. Concretely, we first integrate three basic ZSIR tasks, \textit{i.e.}, object recognition, compositional recognition, and foundation model-based open-world recognition, into a unified element-wise paradigm and provide a detailed taxonomy and analysis of the main approaches. Next, we summarize the benchmarks, covering technical implementations, standardized datasets, and some more details as a library~$\href{https://github.com/eigenailab/Element-Wise-Zero-Shot-Image-Recognition}{\textcolor{blue}{[\mathrm{link}]}}$. Last, we sketch out related applications, discuss vital challenges, and suggest potential future directions.
\end{abstract}

}

\maketitle
\section{Introduction}

\begin{figure}[t]
\centering
\centerline{\includegraphics[width=0.47\textwidth]{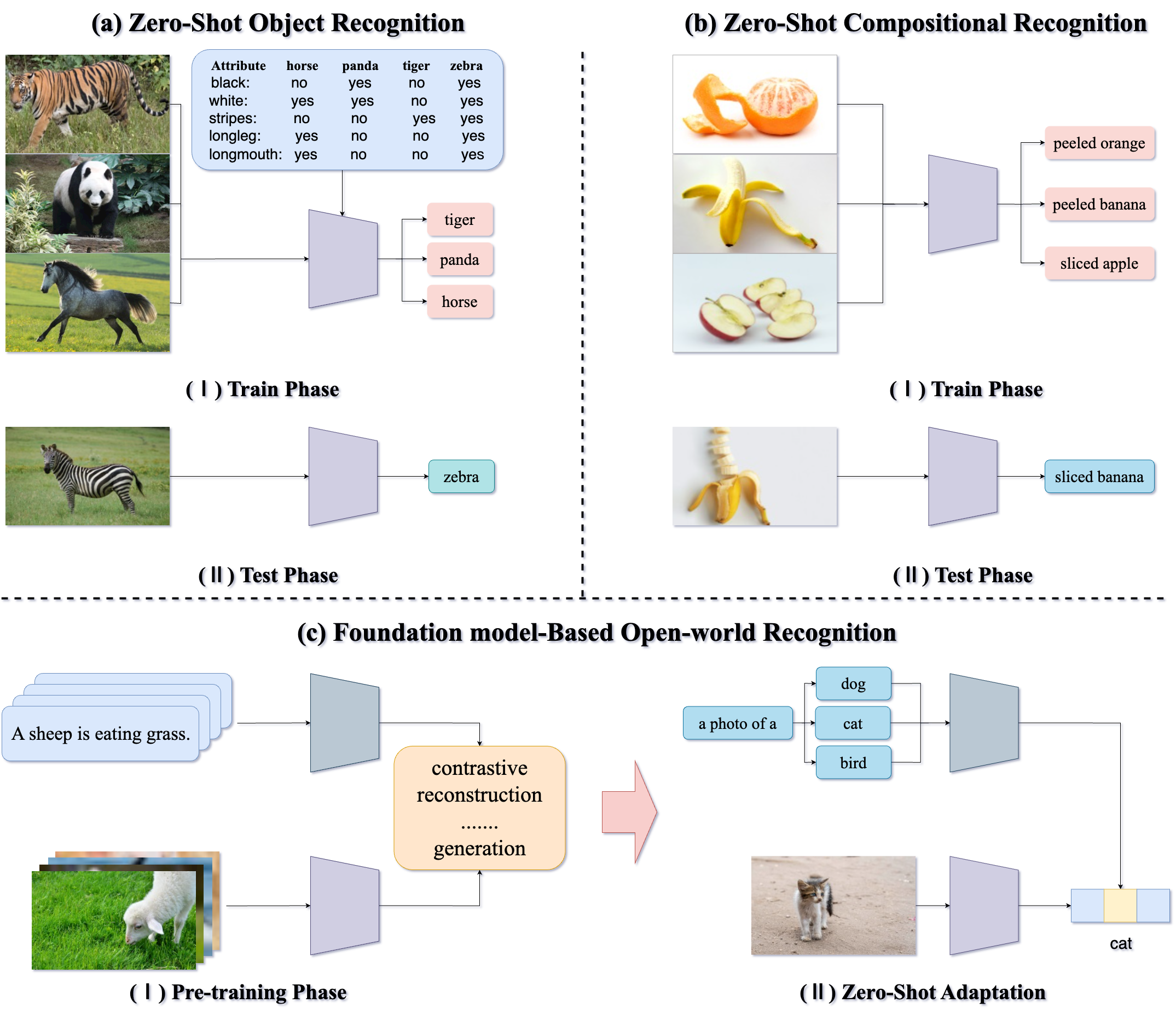}}
\vspace{-3mm}
\caption{Three tasks of ZSIR: (a) ZSOR utilizes shared attributes/texts to identify unseen categories; (b) ZSCR infers unseen compositions by learning from seen objects and states; and (c) FBOR exploits the broad fundamentals learned by pre-trained VLMs to implement zero-shot recognition directly in downstream tasks.}
\label{fig:threetasks_introduction}
\vspace{-6mm}
\end{figure}

\IEEEPARstart{W}{ith} the continuous advancement of deep learning techniques, automated image recognition models are gradually becoming comparable to or even better than humans. 
The success relies on ingenious network architectures and, more importantly, massive data along with accurate annotations. 
%
%two factors, \textit{i.e.}, massive data along with accurate annotations and a close-world environment. 
However, the deficiencies are also obvious. On the one hand, the cost of data collection can be huge. For example, despite being a representative large-scale dataset, ImageNet~\cite{deng2009imagenet} contains 14 million images with 21,814 categories, whose coverage is still far fewer than real-world scenarios, \textit{i.e.}, a conservative estimate suggests that there are at least 3 million species only for the board `\textit{Insect}' category. 
On the other hand, image samples of some categories, such as rare/extinct species, rare medical cases, and privacy-sensitive data, are usually limited or even unavailable. 
Moreover, the requirement for modern deep models has also become more complex, not only to know what an object is but also to describe its accurate states~\cite{chen2014inferring, misra2017red}, e.g. the `\textit{Sliced}' apple and an `\textit{Old}' dog, etc. The superposition of states and objects incurs a geometrical increase in the difficulty of data collection and annotation. 
%
%In recent years, more and more interests have moved from close-world recognition to open environments to endow models with the ability to discriminate novel/unseen categories that well-mimic the human recognition process~\cite{xian2018zero, purushwalkam2019task, pourpanah2022review}. 
In recent years, more and more interests have moved from coarse- to fine-grained scenarios to endow models with the ability to discriminate microscopic characteristics that mimic the human recognition process~\cite{yang2018learning, akata2015evaluation}. 
%
%The way humans think makes the aspiration possible. People can recognize things by extracting and reorganizing their shared characteristics for logical reasoning~\cite{battaglia2018relational}. 
%
% For example, one can identify `\textit{Zebra}' by combining the morphology of `\textit{Horse}', the stripes of `\textit{Tiger}', and the color of `\textit{Panda}', implying that human cognition of upper concepts can be converted into the dismantling and reconstruction of base \textit{elements} and further motivate the development of element-wise learning (EWL) in the scenario of open environments. 
For example, one can identify `\textit{Zebra}' by recognizing an animal with `\textit{black-and-white}' (color) and `\textit{stripes}' (physical trait), implying that human cognition of upper concepts can be converted into the dismantling and reconstruction of base \textit{elements} and further motivate the development of element-wise learning (EWL) in image recognition tasks. 
%and similarly identify a blue car and a red bicycle by combining a red car and a blue bicycle. 
%It implies that human cognition of upper things can be translated into the dismantling and reconstruction of base elements. 
%
%We have to ask: \textit{Can this paradigm of mind be transplanted to the modern automatic recognition system}?
%
%Naturally, this question motivates the development and application of element-wise learning in open-world recognition tasks. 
%
Concretely, the objective of EWL is to construct a bottom-up paradigm by executing element-wise representation and reasoning, \textit{i.e.}, a pyramidal knowledge architecture from micro to macro, attribute to object, and local to global. In particular, elements are carriers of properties~\cite{nagarajan2018attributes}, which describe the meta concepts of samples. 
%such as materials, colors, patterns, styles, expressions, parts, or functions. 
The concepts or descriptors conveyed by an element vary depending on the specific tasks, like concrete components, e.g., `\textit{Wings}' in recognizing bird species~\cite{wah2011caltech}, or abstract descriptions, e.g., `\textit{Studying} in recognizing scenes~\cite{patterson2012sun}. 

Employing element-wise representation to capture discriminative elements is already customary for fine-grained classification task~\cite{wei2021fine,yang2018learning, akata2015evaluation} and has also been proven effective in many other areas, such as face recognition~\cite{wang2016walk}, image retrieval~\cite{kovashka2012whittlesearch}, and visual question answering~\cite{wu2017image}. 
On top of that, element-wise reasoning is employed to assemble a complex set of elements to achieve the recognition. 
%Notably, the crux of its extension to the open environments relies on novel combinations of base elements. 
%
The very early stage began with systematic and structured combinational logic such as predicate logic, boolean logic, and so on~\cite{santa2018neural}, and then researchers explored the relationship between combinational elements and objects in visual recognition tasks~\cite{biederman1987recognition, hoffman1987parts}. Afterward, modern technologies are also designed to follow the idea of combinational reasoning, such as deformable network~\cite{felzenszwalb2008discriminatively} and graph~\cite{battaglia2018relational}.
%
%It is inspired by the hallmark of human intelligence~\cite{misra2017red}, which possesses systematic and structured combinational logic such as predicate logic, boolean logic, and so on~\cite{santa2018neural}. 
% In the early stage, researchers have explored the relationship between combinational elements and objects in visual recognition~\cite{biederman1987recognition, hoffman1987parts}. Afterward, modern technologies are also designed to follow the idea of combinational reasoning, such as deformable network~\cite{felzenszwalb2008discriminatively} and graph~\cite{battaglia2018relational}.

Zero-shot image recognition (ZSIR), as a well-defined visual-to-semantics mapping problem adapted for unseen domains, has attracted increasing attention in most recent years and made profound progress~\cite{lampert2009learning}. ZSIR employs texts/words as the medium to bridge the image visual disparity between seen and unseen categories, thus generalizing the models to recognize unseen domains. 
Simply put, the visual features are compressed into the shared semantics, and the model unifies the shared semantics and visual features of the seen categories to reason images from unseen categories. 
Compared to other low-shot schemes such as few-shot learning~\cite{wang2020generalizing} and one-shot learning~\cite{o2019one}, ZSIR reduces the data requirement to zero, thus completely releasing the data collection and annotation pressure. Moreover, compared to another similar open-world problem, \textit{i.e.}, out-of-distribution (OOD) detection~\cite{yang2024generalized}, ZSIR is designed to assign precise labels for unseen data instead of simply categorizing them as OOD samples, which is usually more favorable in real-world applications. 

% In practice, ZSIR and EWL are highly compatible in the recognition process that well-mimics the way humans observe the world, \textit{i.e.}, learning the basics through fine-grained visual-to-semantics interaction to identify novel concepts. Their integration can provide a powerful and promising solution for real open-environment deployment. 
In practice, ZSIR and EWL are highly compatible in the recognition process that well-mimics the way humans observe the world, \textit{i.e.}, learning the basics through fine-grained visual-to-semantics interaction to identify concepts. 
Moreover, ZSIR can also be inherently considered as a natural extension of EWL towards the open-environment scenarios, \textit{i.e.}, infer unseen categories with novel combinations of learned base elements. 
Therefore, their integration can provide a powerful and promising solution for real open-environment deployment. 
%In recent years, ZSIR techniques based on element-wise representation and reasoning have been rapidly developed and widely applied. 
However, existing studies on ZSIR and EWL are usually fragmented and independent of each other, lacking a systematic investigation that summarizes their integration with a unified paradigm. Although some related reviews~\cite{xian2018zero, wang2019survey, pourpanah2022review} have been presented for ZSIR in recent years, they do not delve in-depth into element-wise techniques. 
Worse still, existing reviews focus only on zero-shot object recognition, which is only one subtask of current ZSIR, and thus are constrained from sufficiently presenting the holistic portrait of ZSIR. 
To fill the gap, we conduct an extensive survey to thoroughly investigate recent advances in the development of element-wise ZSIR approaches and attempt to integrate the three mainstream tasks of ZSIR, including object recognition, compositional recognition, and foundation model-based open-world recognition, to provide a unified and more comprehensive review. 

In summary, our contributions are four-fold:

\begin{itemize}
    \item We, for the first time, integrate three mainstream ZSIR tasks, \textit{i.e.}, object recognition, compositional recognition, and foundation model-based open-world recognition, to provide a comprehensive review from an element-wise perspective.
    
    \item We discuss the differences and synergies of different ZSIR tasks and analyze the key challenges. We summarize related cutting-edge techniques and provide a new taxonomy of element-wise ZSIR.
    
    \item We collect and summarize some benchmarks, including implementations, datasets, and some more details, as a library for follow-up research.

    \item We present the widespread applications, discuss vital challenges and suggest potential future directions.     
\end{itemize}

%\section{Description, Comparison, Challenge, and Organization}
\section{Overview}

\subsection{Mainstream Tasks}

% \begin{figure}[t]
% \centering
% \centerline{\includegraphics[width=0.5\textwidth]{fig/ThreeTasks.pdf}}
% \caption{Three major masks of zero-shot image recognition. Zero-Shot Object Recognition utilizes attributes shared between classes to identify unseen samples. Zero-Shot Compositional Recognition reasons about unseen compositions by learning seen compositions. Foundation model-Based Open-world Recognition exploits the broad fundamentals learned by pre-trained VLMs to implement zero-shot recognition directly in downstream tasks.}
% \label{fig:threetasks_introduction}
% \end{figure}

%Three mainstream ZSIR tasks are integrated into this survey to provide a more comprehensive investigation, including object recognition, compositional recognition, and foundation model-based open-world recognition. 
%Firstly, we give a detailed introduction to the three tasks.

\subsubsection{Zero-Shot Object Recognition}
%\textit{Zero-Shot Object Recognition (ZSOR)}. 
Classic zero-shot learning (ZSL) is the most common setting for ZSIR~\cite{lampert2009learning}, addressing the challenge of classifying images that come from novel unseen categories without observing any prior visual samples before~\cite{lampert2009learning}. In ZSL, the annotations such as attributes, words/sentences, or semantic embeddings (\textit{i.e.}, we use \textit{semantics} uniformly later) are shared to describe categories at a high level. During training, only image samples of seen categories are available, while the model is trained to recognize unseen categories. Hence, the shared semantics serve as the crucial link between seen and unseen categories. To fully adapt to real-world scenarios, generalized ZSL (GZSL)~\cite{xian2018zero} further extends the above setting to a more realistic one by requiring both seen and unseen categories to be jointly recognized during inference. Most recent studies encompass both ZSL and GZSL settings and are collectively referred to as zero-shot object recognition (ZSOR) in this survey (\textit{i.e.}, top-left of Fig.~\ref{fig:threetasks_introduction}).
% Humans can recognize up to 30,000 categories, even those they have never encountered before, by leveraging both verbal and visual cues~\cite{biederman1987recognition}. Inspired by this, Zero-Shot Learning (ZSL)~\cite{lampert2009learning} seeks to replicate this ability in automated models. In ZSL, attribute annotations—such as words, sentences, or semantic embeddings—are provided to describe categories. During training, only images of some categories and their attribute annotations are available, while the model must later recognize unseen categories during inference. The attribute annotations of these unseen categories serve as the crucial link for conveying their semantics. However, ZSL is limited by its focus on unseen categories alone, which does not fully reflect real-world scenarios. Generalized Zero-Shot Learning (GZSL)~\cite{xian2018zero} further extends the setting by requiring both unseen and seen categories to be recognized during the testing period. Most research encompasses both ZSL and GZSL settings, collectively referred to as Zero-Shot Object Recognition. While several surveys have thoroughly reviewed ZSOR techniques, our focus is specifically on element-wise learning approaches within this domain.

\subsubsection{Zero-Shot Compositional Recognition}
%\textit{Zero-Shot Compositional Recognition (ZSCR)}. 
Another critical capability in the recent ZSL research is to describe the sample's relevant states, such as recognizing whether an apple is sliced or raw instead of simply classifying it as an object. Such capability is essential for developing models with a deeper understanding of the world and is also a key component in building human-like intelligence~\cite{lake2017building}. 
Inspired by it, compositional learning has emerged in ZSL to explore combinational reasoning capabilities~\cite{chen2014inferring,misra2017red,purushwalkam2019task}. Concretely, the model is trained from image samples and their corresponding compositional states and objects, and is expected to recognize both seen and novel unseen compositions during inference. 
In particular, each state and object appears at least once during training. The classic setting assumes that the labels' prior is available, \textit{i.e.}, some state-object compositions are filtered out, speeding up the inference. In contrast, the open-world setting~\cite{mancini2021open} extends compositional ZSL by expanding the inference scope to the full set, \textit{i.e.}, every state-object composition counts. Therefore, many irrational combinations can occur, and the model is expected to cull out on its own. We collectively refer to them as zero-shot compositional recognition (ZSCR) in this survey (\textit{i.e.}, top-right of Fig.~\ref{fig:threetasks_introduction}).

\subsubsection{Foundation Model-Based Open-World Recognition}
%\textit{Foundation model-Based Open-world Recognition (FBOR)}. 
%Despite having made significant strides in open-world recognition, 
Both ZSOR and ZSCR fall short of achieving complete openness when faced with the open-world scenario, as their objectives and predictions remain fixed and well-defined. A critical question arises: can models learn broad and foundational knowledge and adapt seamlessly to unknown tasks? The advent of large vision-language pre-trained models (VLMs)~\cite{radford2021learning, jia2021scaling} offers a promising solution. These models focus on broad vision-language interactions, learning vast amounts of elemental knowledge without a specific goal, \textit{i.e.}, much like humans learn about the world through observation. Remarkably, they exhibit excellent zero-shot generalization ability, performing well on various unseen domain image recognition tasks without further training. This capability has found wide applications, particularly in low-shot scenarios~\cite{zhou2022learning, zhou2022conditional}. However, recent studies~\cite{menon2022visual, guo2023calip} suggest that the zero-shot potential of VLMs is not fully realized. In response, foundation model-based open-world recognition (FBOR) has been developed to push such a limit in foundation models (\textit{i.e.}, bottom of Fig.~\ref{fig:threetasks_introduction}).
% As shown in Fig. \ref{fig:threetasks_introduction} (bottom). While ZSOR and ZSCR have made significant strides in open-world recognition, they fall short of achieving complete openness, as their objectives and class predictions remain fixed and well-defined. A critical question arises: Can models learn broad, foundational knowledge like humans and adapt seamlessly to unknown tasks? The advent of large vision-language pretraining models (VLMs)~\cite{radford2021learning, jia2021scaling} offers a promising solution. These models focus on vision-language interactions during training, learning vast amounts of elemental knowledge without a specific goal—much like humans learn about the world through observation. Remarkably, they exhibit excellent zero-shot generalization, performing well on various unfamiliar image recognition tasks without further training. This capability has found wide applications, particularly in low-shot scenarios~\cite{zhou2022learning, zhou2022conditional}. However, recent studies~\cite{menon2022visual, guo2023calip} suggest that the zero-shot potential of VLMs is not fully realized. In response, Foundation model-Based Open-world Recognition has been developed to push the limits of zero-shot performance in foundation models. 

% \begin{figure*}[htbp]
%     \centering
%     \begin{minipage}{\linewidth}
%     \centerline{\includegraphics[width=1.0\textwidth]{fig/flowchart_survey.pdf}}
%     \end{minipage}
%     \caption{An overview of organization.}
%     \label{fig:overview}
% \end{figure*}

\subsection{Comparison}

\noindent \textit{i)~Differences.}  
ZSOR and ZSCR both run in relatively narrow environments, \textit{i.e.}, the categories to be predicted are known. Despite some studies proposing open-world ZSCR, the essence is only to extend the label set from a subset to the full set. FBOR, on the other hand, is task-agnostic, with the need to learn a wider range of knowledge ahead of time to cope with unknown challenges. 
In general, the element annotations of ZSOR are usually larger and more detailed, while ZSCR contains only individual state and object (category) annotations. On the other hand, the element annotations of ZSOR are category-level and thus may shift on individual instances, while the annotations of ZSCR are instance-level. Moreover, the goal of ZSOR is to recognize objects composed of different attributes, whereas the goal of ZSCR is to recognize combinations composed of different states and objects.
\noindent \textit{ii)~Commonalities.} 
As mentioned earlier, element-wise representation and reasoning are motivation shared by several tasks, \textit{i.e.}, learning the visual meta-concepts corresponding to text annotations, and then conducting combinational reasoning to recognize novel upper-level concepts. In addition, many techniques are common across several tasks, such as attention mechanism, graph modeling, etc., which we will discuss further in Sect.~\ref{shared technologies}.

\subsection{Challenge}
\subsubsection{Fine-Grained Visual Analysis}
Element-wise ZSIR requires detailed visual anatomy to echo the semantics and capture underlying meta concepts. ZSIR has fine-grained semantic information to describe the element-wise characteristics. However, due to huge costs, images in hand typically lack enough element-wise annotations. Therefore, semantics and visual local clues are usually learned in a weakly supervised manner, and their matching accuracy cannot be fully guaranteed. Worse still, irrelevant clues, such as the backgrounds, can inevitably interfere with the learning quality. 
To address these issues, recent studies incorporated fine-grained visual analysis into ZSIR and have made certain progress, particularly in accuracy improvement. Studies of such kind mainly involve visual localization~\cite{zhu2019semantic, xie2019attentive}, visual representation~\cite{xie2020region, guo2023graph}, and visual-semantics correlation modeling~\cite{huynh2020fine, chen2022transzero}. 
However, based on our observation, three challenges still hinder the fine-grained visual analysis in ZSIR, including 
\textit{i)~Elements Complexity,}
\textit{i.e.}, the number of element-wise annotations tends to be larger compared to category-wise annotations, and the corresponding visual features are much more subtle~\cite{wah2011caltech, liu2016deepfashion}. In addition, the processing of abstract attributes is also a challenge due to the difficulty of finding the corresponding explicit visual clues; 
\textit{ii)~Visual Occlusion and Overlap,}
\textit{i.e.}, visual occlusion occurs between objects and backgrounds or between elements and elements, which may result in the absence of critical visual clues. Besides, visual overlap refers to multiple semantic expressiveness of the same visual clues~\cite{atzmon2020causal}. For example, the model can hardly completely decouple the visual clues of `\textit{red}' and `\textit{car}' when recognizing a red car; and 
\textit{ii)~Visual Interaction,} 
\textit{i.e.}, the independent semantics of visual clues can be vague or even ambiguous. For example, if a piece of skin is cut off from a cow's trunk, it is hard to recognize which part or even which species the skin comes from. Thus, the semantic meaning of visual clues can also depend on the whole to varying degrees. 

\subsubsection{Domain Sift}
Domain shift is a long-standing challenge in ZSIR, which refers to the distribution bias (\textit{i.e.}, towards seen categories) of visual features when mapped into semantic space~\cite{fu2015transductive}. 
The inducement of domain shift is the absence of visual prior in unseen categories, which makes the models favor too much on seen categories and thus learn biased mapping functions~\cite{jiang2019transferable}. It further misleads the model in recognizing the embeddings of samples of unseen categories as somewhat seen categories that hinder the model's generalization ability. 
Meanwhile, some other factors can also aggravate the domain shift problem. 
First, different imaging mechanisms cause variational views, lightings, backgrounds, etc.~\cite{zhou2022domain}, and thus, the visual characteristics can be diverse even for samples of the same category. Domain adaptation and generalization~\cite{ganin2015unsupervised, zhou2022domain, rao2023srcd} partially mitigates this issue in general recognition tasks, while ZSIR is much more challenging since domain shift and unknown prediction need to be accommodated at the same time. 
Second and more troublesomely, context dependency, \textit{i.e.}, objects or states expressing the same semantic information may present very different visual characterizations in different contexts~\cite{ge2022dual, wei2019adversarial}. For example, the object `\textit{tail}' appears hairy and straight on a \textit{horse} but smooth and curly on a \textit{pig}, and the state `\textit{old}' presents itself as the metallic luster of rust on a \textit{car}, while as the wrinkled skin on a \textit{person}. 

\begin{figure*}[t]
    \centering
    \begin{minipage}{\linewidth}
    \centerline{\includegraphics[width=1.0\textwidth]{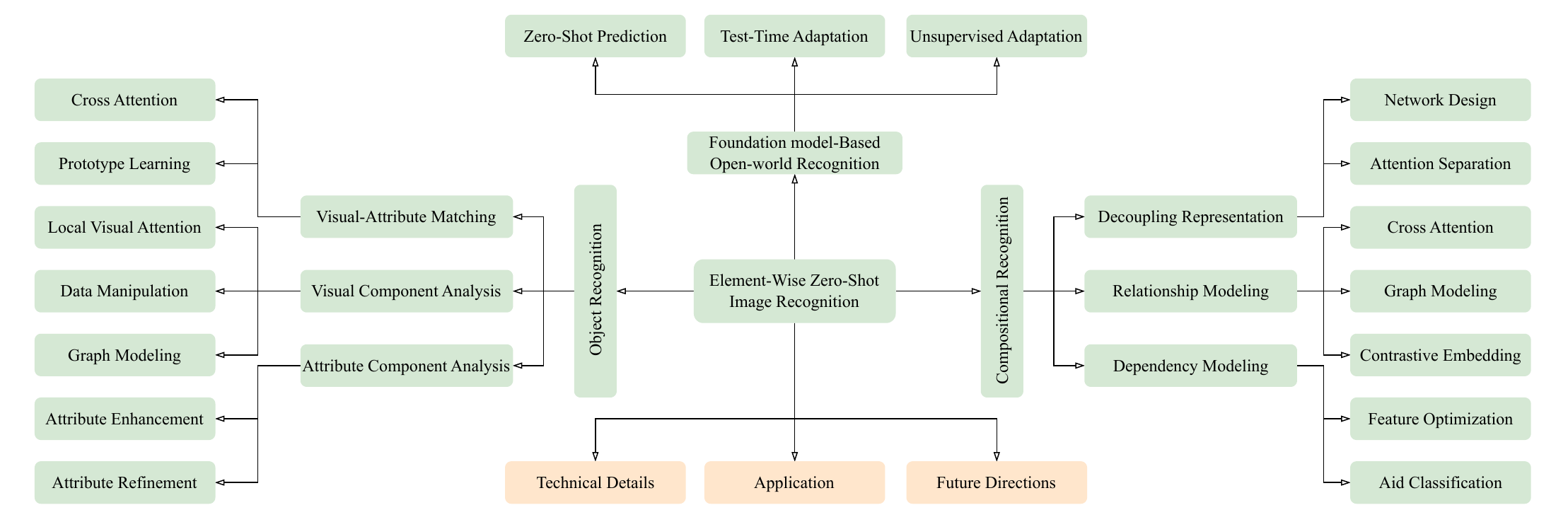}}
    \end{minipage}
    \caption{An overview of the organization and taxonomy of this survey.}
    \label{fig:overview}
\end{figure*}

\subsection{Organization}
To sum up, Fig.~\ref{fig:overview} outlines the overall organization as well as the taxonomy of this survey. 
Concretely, Sect.~\ref{section:zs-or} first covers ZSOR techniques, including problem formulation and key algorithms. Similarly, Sect.~\ref{section:zs-cr} reviews ZSCR, and Sect.~\ref{section:fbor} introduces FBOR. Then, Sect.~\ref{section:technical details} details technical implementations, standardized datasets, and shared techniques. Sect.~\ref{section:application} explores widespread applications, and Sect.~\ref{section:future directions} suggests potential future directions.

\section{Zero-Shot Object Recognition}
\label{section:zs-or}

\subsection{Problem Formulation}

Given the seen domain $\mathcal{D}^s=\{(x^s, y^s, a^s)|x^s\in\mathcal{X}^s, y^s\in\mathcal{Y}^s, a^s\in\mathcal{A}^s\}$, where $\mathcal{X}^s$, $\mathcal{Y}^s$, and $\mathcal{A}^s$ denote visual samples, category labels, and semantics (e.g., a set of attributes), 
and similarly, let $\mathcal{D}^u=\{(x^u, y^u, a^u)|x^u\in\mathcal{X}^u, y^u\in\mathcal{Y}^u, a^u\in\mathcal{A}^u\}$ denote the unseen domain. 
Notably, $a^s\cup a^u \in \mathbb{R}^{m\times k}$ are class-level semantics annotations, where $m$ denotes the number of categories and $k$ is the number of attributes.
Without loss of generality, the task of ZSOR can be modeled as learning a mapping/relational function $\Psi: \mathcal{X} \to \mathcal{A}$, wherein $\mathcal{X}^u$ is strictly inaccessible for training. 
During inference, the learned function $\Psi$ is applied to recognize samples from the unseen domain only, \textit{i.e.}, ZSL, or from both seen and unseen domains, \textit{i.e.}, Generalized ZSL (GZSL). 

The success of ZSOR relies on the shared attribute between $\mathcal{A}^s$ and $\mathcal{A}^u$, which act as the bridge between seen and unseen domains. The intuitive idea is to learn the visual representations corresponding to attributes, which is the notion of element-wise learning. However, due to the single target and relatively closed environment of object recognition, learning for $\Psi$ can be transformed into a plain class-wise distribution matching problem. Class-wise distribution matching concerns neither the discrepancies of the bottom attributes nor the variations of fine-grained vision. It treats both visual and attribute features as a whole, and the objective is to reduce the distributional distance between the two to learn an unbiased mapping function. Mainstream techniques include semantic space design, generative approach, and regularization. Semantic space design is to search for a suitable low-dimensional space that perfectly matches visual and attribute embeddings~\cite{romera2015embarrassingly, jiang2019transferable}. The generative approach is to simulate the visual features of unseen classes~\cite{xian2018feature, verma2018generalized} by training a generator that produces visual features based on attributes with the help of networks such as GANs~\cite{goodfellow2020generative, chen2020canzsl, chen2020rethinking}, VAEs~\cite{kingma2013auto, chen2021semantics} and generative flows~\cite{chen2022gsmflow, chen2021mitigating}. Regularization is imposing constraints to improve the mapping performance or robustness of the model. For example, Liu et al.~\cite{liu2018generalized} use an information entropy loss to reduce uncertainty. Despite the impressive results achieved by these methods, they provide limited insights for micro-operations and bottom-up analyses and are, therefore, out of the scope of this survey. The readers can refer to other related surveys~\cite{wang2019survey, pourpanah2022review} for a deeper understanding.

Depending on the concrete elements the method operates on, we categorize the element-wise techniques into three broad categories, \textit{i.e.}, \textit{Visual Component Analysis}, \textit{Visual-Attribute Matching}, and \textit{Attribute Component Analysis}. Then, the methods are further categorized into different sub-categories based on the specific techniques they use.

\subsection{Visual Component Analysis}

Visual component analysis is the refined dissection of an image or its features, involving localization, modeling, and so on. Images often contain a large amount of redundant information, only a few of which are decisive for discrimination. At the same time, images lack fine-grained annotations. Therefore, the goal of visual component analysis is to reduce the interference of irrelevant information as well as noise, and prompt the model to focus on those critical visual parts. Commonly used methods can be classified into the following three sub-categories: \textit{Local Visual Attention}, \textit{Data Manipulation}, and \textit{Graph Modeling}.

\subsubsection{Local Visual Attention}

\begin{figure}[htbp]
    \centering
    \begin{minipage}{\linewidth}
    \centerline{\includegraphics[width=0.75\textwidth]{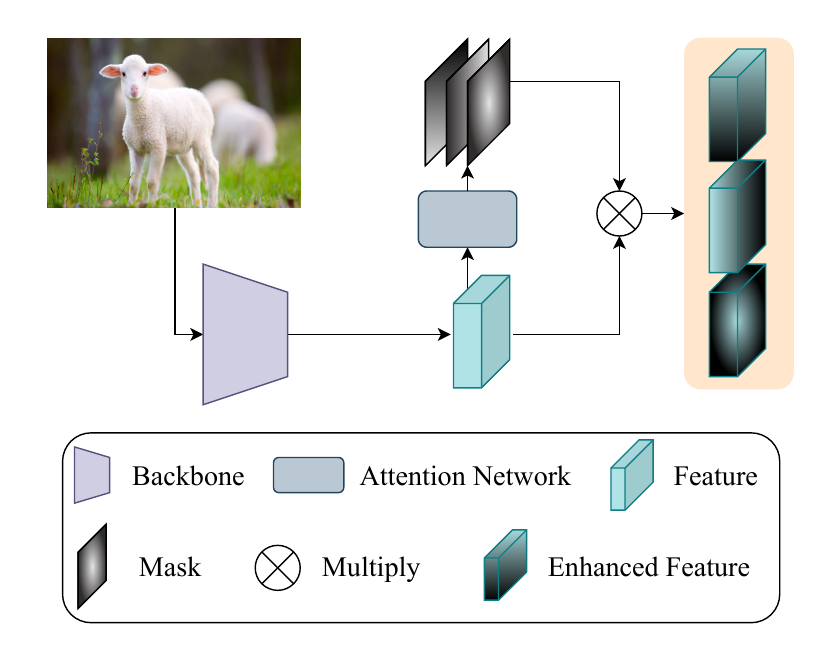}}
    \end{minipage}
    \vspace{-3mm}
    \caption{A schematic diagram of local visual attention. Visual features are passed through a sub-network to generate multiple masks, which are then multiplied to obtain enhanced features.}
    \label{fig:localattention}
\vspace{-4mm}
\end{figure}

Attention mechanisms are effective to improve efficiency and accuracy in many visual tasks by mimicking human focuses. Motivated by this, local visual attention is to utilize the attention mechanism to identify the most valuable visual regions and reduce the influence of background or irrelevant regions. Specifically, local visual attention generates multiple masks for visual features. Each mask is a normalized weight matrix that reveals a significant region. Enhanced features can be obtained by weighting the original features using mask, thus improving the discriminability (Fig. \ref{fig:localattention}). Suppose that $x\in \mathbb{R}^{C\times H\times W}$ denotes the visual feature, where $C$, $H$, and $W$ are the channel, height, and weight, respectively. Let $M=\mathcal{F}(x)$ represents the attention mask, where $\mathcal{F}$ denotes a small learnable network, e.g., $1\times 1$ convolutional network, and $M\in \mathbb{R}^{N \times H \times W}$ denotes $N$ attention masks. Afterward, multiplying the masks with the original features yields $N$ regional features, which can be expressed as:
\begin{equation}
    x_{region} = \{xm_1, xm_2, ..., xm_N\},
\end{equation}
where $x_{region} \in \mathbb{R}^{N\times C\times H\times W}$ and $[m_1, m_2, ..., m_N]=M, m_i \in \mathbb{R}^{H \times W}$. How to use these regional features to improve model's performance. AREN~\cite{xie2019attentive} considers certain mask learned regions to be unimportant and therefore adopts an adaptive threshold to further filter out unimportant information, \textit{i.e.}, region features smaller than the threshold are set to 0. RSAN~\cite{wang2021region} expects the regions learned by each mask to correspond to the attributes one by one. It uses max-pooling to get the representative value in each region feature and concatenates them into a one-dimensional vector, after which it uses Mean Squared Error Loss to minimize the distance between this vector and the attribute embedding. Considering the existence of topological relationships between regions, RGEN~\cite{xie2020region} introduces graph modeling to maintain their potential connections.

\subsubsection{Data Manipulation}

Data manipulation has a same objective as local visual attention, \textit{i.e.}, to search and localize the most critical regions. The difference is that the latter works at the feature space, whereas the former works at the image level. An intuitive approach is employing detection techniques to recognize parts of an object, such as a bird's head, body, and claws. The detection network can provide coordinates for multiple parts simultaneously, allowing image patches to be cropped accordingly or deep features to be localized. This method introduces fine-grained semantic information that enhances the model's classification ability~\cite{ji2018stacked, elhoseiny2017link}. Additionally, it boosts the performance of generative methods, which often struggle with capturing fine-grained visual differences. For example, AGAA~\cite{zhu2018generative} utilizes a detector to extract region information that is fed into a discriminator, along with class embeddings generated by a generator. However, this technique is limited in practice since the detector's performance heavily depends on part-level visual annotations, which are often impractical to obtain.

To alleviate this problem, a series of unsupervised detection methods have been proposed. LDF~\cite{li2018discriminative} employs a coordinate generating network that generates coordinates directly from image features. Specifically, suppose $x \in \mathbb{R}^{C\times H\times W}$ is the visual feature, and the coordinate can be expressed as:
\begin{equation}
    Z = [z_h, z_w, z_l] = \mathcal{F}(x),
\end{equation}
where $\mathcal{F}$ is a learnable network. $Z$ is the window, $z_h, z_w$ denote the coordinates, and $z_l$ denotes the length of the region. After that, the regions located at the coordinates are zoomed in and re-fed to the classification network. SGMA~\cite{zhu2019semantic} incorporates local visual attention techniques to coordinates generation. Furthermore, SR2E~\cite{ge2021semantic} and ERPCNet~\cite{li2022entropy} introduce the reinforcement learning paradigm, which induce the cropping behavior of the model by setting up special rewards and penalties to obtain the most discriminative sub-image. POPRNet~\cite{liu2024part} takes unsupervised detection technique. It generates multiple proposal regions for an image and then sorts and selects Top-K region features based on category confidence. ZSLViT~\cite{chen2024progressive} utilizes the patch split mechanism of ViT~\cite{dosovitskiy2020image} to get multiple region features and later filters the key visual regions by calculating the attention weights between the regional and class tokens.

Unlike the above approaches, to capture the key visual semantic embeddings, VGSE-SMO~\cite{xu2022vgse} takes a patch clustering approach. It utilizes an unsupervised segmentation algorithm to decompose images into multiple patches, after which individual patches are linked to class labels via a classification network. By doing so, the model can implicitly learn the relationship between classes and fine-grained visual patches that align better with each other.

\subsubsection{Graph Modeling}

Visual region localization helps models focus on important local features but often overlooks the potential relationships between regions. For example, a bird’s body parts are biologically connected, and treating them independently can cause semantic ambiguity due to a narrowed field of view. Sharing information between regions can mitigate this by preserving essential semantic context. Graph modeling techniques are well suited for handling such irregular non-geometric data with a structured message propagation mechanism~\cite{battaglia2018relational}. In recent years, with the birth and development of graph convolutional networks (GCNs)~\cite{kipf2016semi}, graph modeling techniques have been widely used in various computer vision tasks.

Regional visual features are natural graph nodes and thus can be well embedded in graph modeling techniques. For example, GKU~\cite{guo2023graph} constructs a graph for each sample to capture fine-grained local structural relationships. They redefine the visual-semantic mapping problem as a graph-semantic mapping problem, thus suppressing the influence of partial visual domain shift. However, GKU requires the dataset to be supported by fine-grained regional annotations. Without the supervision of manual annotations, RIAE~\cite{hu2022region} incorporates all regions into the graph node set. To mine the relationships between regions, it introduces cosine similarity to measure the weight of an edge between any two region nodes. GNDAN~\cite{chen2022gndan}, on the other hand, first locates the visual regions corresponding to the attributes with cross attention techniques, and then later uses the graph for modeling.

\subsection{Visual-Attribute Matching}

The goal of visual-attribute matching is to align visual features with their corresponding attributes by learning fine-grained, element-wise representations through the pairing of visual-attribute embeddings. Unlike visual component analysis, this approach explicitly focuses on interacting with attributes, allowing for a better perception of subtle attribute variations and improved generalization to unseen domains. This text-driven visual learning aligns more closely with human learning patterns and presents a greater challenge. Current methods in this area can be categorized into two sub-categories: \textit{Cross Attention} and \textit{Prototype Learning}.

\subsubsection{Cross Attention}
% \begin{figure}[htbp]
%     \centering
%     \begin{minipage}{\linewidth}
%     \centerline{\includegraphics[width=0.9\textwidth]{fig/crossattention.pdf}}
%     \end{minipage}
%     \vspace{-3mm}
%     \caption{A schematic diagram of cross attention. Attribute embeddings and region features compute similarity to obtain multiple attention maps.}
%     \label{fig:crossattention}
% \vspace{-4mm}
% \end{figure}
Cross attention allows for the simultaneous processing of different sources of information and efficient modeling of their dependencies. Among them, dot product attention, due to its high computational efficiency and low space cost, is widely applied in many advanced models such as Transformer~\cite{vaswani2017attention}. The core of dot product attention is to compute the similarity between query and key-value pairs to assign attention weights, and the similarity measure is the dot product of vectors. Given a query series $Q$, a key series $K$ and a value series $V$, the attention score can be formulated as:
\begin{equation}
Attention(Q, K, V) = softmax(\frac{QK^{\mathsf{T}}}{\sqrt{d_k}})V,
\end{equation}
where $d_k$ denotes the dimension of the key vector used to scale the dot product result. When $Q$, $K$, and $V$ all come from the same input series, the attention score learns the dependencies between the parts of the series, \textit{i.e.}, self-attention. When there are two different input series involved in the computation, the attention score learns the dependencies between the different series, \textit{i.e.}, cross attention.

\begin{figure}[t]
    \centering
    \begin{minipage}{\linewidth}
    \centerline{\includegraphics[width=0.85\textwidth]{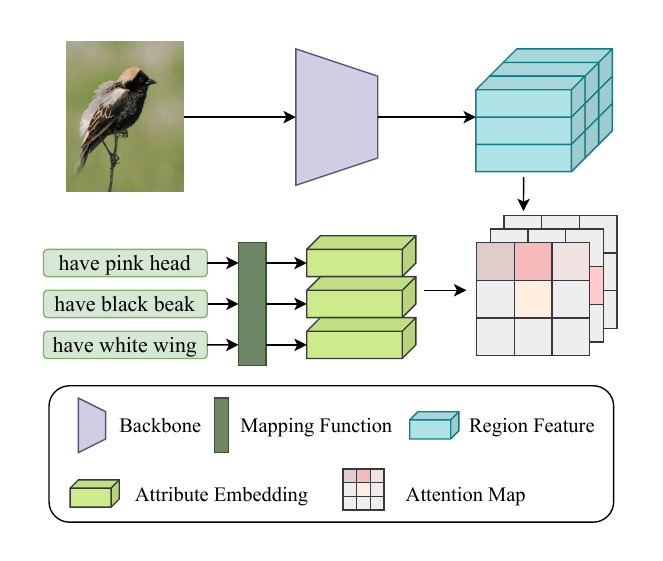}}
    \end{minipage}
    \vspace{-3mm}
    \caption{A schematic diagram of cross attention. Attribute embeddings and region features compute similarity to obtain multiple attention maps.}
    \label{fig:crossattention}
\vspace{-4mm}
\end{figure}

To uncover the interactions between attributes and visual regions, the cross attention technique is naturally introduced into ZSOR (Fig. \ref{fig:crossattention}). For example, DAZLE~\cite{huynh2020fine} proposes a dense attention technique that assigns weights to each region based on attributes. Specifically, it utilizes the features output from the last layer of the convolutional network and treats each point on the feature as a region. The similarity between attributes and regions is then computed with cross attention. Suppose $x\in \mathbb{R}^{C\times r}$ denotes the visual feature with $r=H\times W$ regions. Let $a\in \mathbb{R}^{d\times m}$ denote the attribute vector, where $m$ is the number of attributes, and $d$ is the vector dimension. The cross attention score is calculated as:
\begin{equation}
    Attention(a, x) = \frac{\exp(a^{\mathsf{T}}wx)}{\sum^r\exp(a^{\mathsf{T}}wx)}x, \label{eq:attribute-to-region}
\end{equation}
where $w\in \mathbb{R}^{d\times C}$ is a learnable matrix that ensures the visual and attribute vectors are in the same dimension space. The weighted visual features reflect attribute preferences. After that, the attention scores of all the attributes are used for classification. GEM~\cite{liu2021goal} employs the same approach, however it retains the maximum attention scores with max-pooling during classification. Composer~\cite{huynh2020compositional} synthesizes the features of unseen classes with critical regions identified by attention mechanism, while AREES~\cite{liu2022zero} exploits the regional features to improve the performance.

Some studies resort to ViT~\cite{dosovitskiy2020image} to model dependencies between vision and attributes~\cite{naeem2022i2dformer, chen2023duet, liu2023progressive}. The essential cross attention mechanism is identical. The difference is that ViT divides the image into patches, each of which is considered a visual region. 
In addition, bidirectional attention is explored in depth. It is important to know not only which region the attribute is attending to, but also which attribute the region is attending to. Based on this idea, several methods are developed. For example, MSDN~\cite{chen2022msdn} proposes a mutual distillation network to align attribute-region attention. Ge et al.~\cite{ge2022dual} also integrate bidirectional attention techniques to improve the visual-attribute matching ability of the model.

\subsubsection{Prototype Learning}

The gist of prototype learning is to perform tasks like classification and regression using a set of representative samples, known as prototypes~\cite{yang2018robust} to regularize the learning process. A prototype is a standardized example that captures the essence of a category or type, such as the mean vector of a set of vectors or the center of mass of a distribution. Prototype learning mimics human analogical reasoning by drawing conclusions based on typical examples. Its main advantage is its robustness in handling complex data distributions, particularly when there is overlap or imbalance between data categories. As a result, prototype learning is considered to be more effective for tasks like out-of-distribution detection~\cite{liu2023improving} and low-shot learning~\cite{wang2019panet}.

For visual-attribute matching, there is semantic overlap in the visual region. The same region may contribute to multiple attributes. Although cross attention is able to measure the contribution value by similarity, the lack of an explicit means of visual decoupling may lead to pseudo-correlation of visual-attributes. Therefore, learning a separate visual prototype for each attribute is a promising solution. When the prototype is acquired, the visual region can decide which attribute it belongs to by calculating the distance between the region vector and the prototype. It can be observed that cross attention discovers pairing relations through the similarity between the attribute and visual embeddings, while prototype learning discovers visual-attribute pairing relations indirectly through the similarity between the embeddings of samples and prototypes. 
%visual embeddings and visual prototype embeddings.

APN~\cite{xu2020attribute} first introduces prototype learning to visual-attribute matching. It initializes a learnable prototype vector for each attribute and then calculates the similarity between the visual region and the prototype. After that, the largest region vector is taken as the attribute vector, and then regresses with the attribute embedding. In addition, it incorporates a decorrelation loss~\cite{jayaraman2014decorrelating} to ensure the independence between individual attributes. DPPN~\cite{wang2021dual} considers that the prototype possesses a seen-domain preference, which cannot capture the delicate instance-level attribute shift. Therefore, it is necessary to equip each instance with exclusive attribute prototypes. It proposes a progressive attribute localization method, which dynamically adjusts the prototype by locating attributes through the prototype while updating the prototype in turn. In addition, some other methods~\cite{ge2022dual,guo2023group,cheng2023discriminative,du2023boosting,chen2023zero} also achieve favorable results by using prototype learning.

\subsection{Attribute Component Analysis}
Attributes serve as the crucial link between seen/unseen classes and provide essential supervisory signals for learning fine-grained visual representations. Consequently, the quality of the attribute space largely determines the upper limit of a recognition system's performance. However, most attributes are manually created or pre-trained without proper adjustment, resulting in redundancy, noise, and unstructured relationships. To address this, attribute component analysis aims to optimize the attribute space through methods such as \textit{Attribute Enhancement} and \textit{Refinement}.

\subsubsection{Attribute Enhancement}

Attribute enhancement aims to optimize attribute embeddings to improve their discriminability, using techniques like key attribute localization and attribute space optimization. Key attribute localization functions similarly to visual attention, identifying and focusing on the most critical parts. The cross attention mechanism mentioned in Equation \ref{eq:attribute-to-region} is one of the attribute localization techniques, just changing the input sequences. In addition, LFGAA~\cite{liu2019attribute} designs a latent guided attention module. It assumes that each image has different attention for attributes and, therefore, introduces multiple subnetworks to generate exclusive attribute weights for each image. Attribute space optimization adjusts pre-trained attribute embeddings to reduce semantic overlap that affects discrimination, especially in fine-grained datasets. For instance, many bird species share similar attributes, which makes distinguishing them challenging. For this reason, AREES~\cite{liu2022zero} delineates the attribute embeddings with similar semantic distances into a cluster by K-means clustering algorithm, and the attributes in each cluster are subtracted from the cluster's center of mass, \textit{i.e.}, the mean vector, to obtain the enhanced attribute vector. APNet~\cite{liu2020attribute} introduces graph modeling to discover the logical relationships between classes and updates the attributes by propagating the attribute features of neighboring nodes.

\subsubsection{Attribute Refinement}

Since attributes are typically annotated by humans, the value they add from a human perspective to deep learning is uncertain. Attribute refinement seeks to address this by examining the design of attribute annotations, though few studies exist on the topic. MCZSL~\cite{akata2016multi} points out that the quality of manual annotations is uncontrollable and time-consuming. Therefore, it attempts to design an automated attribute annotation scheme. Specifically, it obtains attribute descriptions from online media such as Wikipedia and designs a series of filtering strategies to filter out inappropriate attributes. DEDN~\cite{rao2024dual} proposes a divide-and-conquer attribute learning scheme. It argues that attribute annotations are confusing and lead to semantic asymmetry problems. For example, for the attributes describing scenes, \{\textit{studying, swimming}\} and \{\textit{stone, tree}\} are not in the same dimension, with the former denoting function and the latter denoting object. Therefore, it designs a dual expert network in which one expert network focuses on processing attribute information in different dimensions.

\section{Zero-Shot Compositional Recognition}
\label{section:zs-cr}

\subsection{Problem Formulation}

Suppose $\mathcal{S} = \{s_1, s_2, ..., s_n\}$ is a set containing $n$ state elements and $\mathcal{O} = \{o_1, o_2, ..., o_m\}$ is a set containing $m$ object elements. Then a compositional set can be obtained as $\mathcal{Y}=\{(s_1, o_1), (s_1, o_2), ..., (s_n, o_m) | s\in \mathcal{S}, o\in \mathcal{O}\}$, which containing $n\times m$ compositions. Now we have a seen domain $D^s=\{(x^s, y^s)|x^s\in \mathcal{X}^s, y^s\in \mathcal{Y}^s\}$, where $\mathcal{X}^s$ denotes the training image set and $\mathcal{Y}^s \subseteq \mathcal{Y}$ is the label set. Similarly, we have an unseen domain denoted by $D^u=\{(x^u, y^u)|x^u\in \mathcal{X}^u, y^u\in \mathcal{Y}^u\}$, where $\mathcal{X}^u$ denotes the test image set and $\mathcal{Y}^u \subseteq \mathcal{Y}$ is the label set. Data in the seen and unseen domains are disjoint, \textit{i.e.}, $X^s\cap X^u=\phi$ and $Y^s\cap Y^u=\phi$. The objective of ZSCR is to generalize to the unseen domain by learning the element-wise knowledge in the seen domain. Specifically, for closed-world compositional recognition, the goal is to learn a mapping function $\Psi: \mathcal{X} \to \mathcal{Y}^s\cup \mathcal{Y}^u$, wherein $\mathcal{X}^u$ is strictly inaccessible for training. Typically, the size of $\mathcal{Y}^s\cup \mathcal{Y}^u$ is smaller than that of $\mathcal{Y}$ since some compositions are not feasible. For open-world compositional recognition~\cite{mancini2021open}, the goal is to learn a mapping function $\Psi: \mathcal{X} \to \mathcal{Y}$, assuming that which compositions are infeasible is unknown. 

We divide the cutting-edge techniques of ZSCR into three categories based on their core components, including \textit{Decoupling Representation}, \textit{Dependency Modeling}, and \textit{Relationship Modeling}. We then further split the sub-categories based on specific technical details or targets of action.

\subsection{Decoupling Representation}

A common baseline approach involves mapping images to deep semantic space, computing similarity to text embeddings, and optimizing with cross-entropy loss. The goal is to learn the posterior probability $p(y|\psi(x))$, where $y=(s, o)$ and $\psi$ is a function mapping the image into the semantic space. This approach leads to model simplifying optimization by learning direct correspondences between visual features and labels, often missing independent effects of states and objects. For example, if the model has only seen \textit{sliced apple}, it may tightly associate \textit{sliced} with \textit{apple} and fail to recognize \textit{sliced orange}. An alternative is to separately predict the labels for states and objects to avoid semantic interference, known as decoupling representation. This approach aims to disentangle states and objects, encouraging the model to learn independent visual representations. By separately recognizing states and objects, the model can better generalize to new compositions. Decoupling representation focuses on learning posterior probabilities $p(s|\psi_s(x))$ and $p(o|\psi_o(x))$, where $\psi_s$ and $\psi_o$ are mapping functions. Mainstream approaches can be categorized as: \textit{Network Design} and \textit{Attention Separation}.

\subsubsection{Network Design}

Many studies resort to modifying the network structure to decompose visual features. Nan et al.~\cite{nan2019recognizing} add two sub-networks at the bottom of the backbone network to extract state and object features, respectively, and then each predicts the corresponding labels and is trained with cross-entropy loss. CASUAL~\cite{atzmon2020causal} employs a similar network structure. The difference is that it introduces a text-to-image mapping network that reverse maps text embeddings into visual space. Meanwhile, it uses the Hilbert-Schmidt Information Criterion, which is a non-parametric method for estimating the statistical dependence between samples of two random variables, to optimize training for ensuring state and object independence. ProtoProp~\cite{ruis2021independent} uses a prototypical network to learn independent representations. Specifically, regional features of images and prototypes of objects and states are computed for similarity and later optimized with cross-entropy loss. Hu et al.~\cite{hu2023leveraging} propose a two-layer decoupling network. First, three sub-networks are used to extract state, object, and composition features, respectively. In the second layer, the state and object features are fed into a reorganization network to synthesize into a composition, while the composition feature is fed into two other sub-networks for further decoupling. CAILA~\cite{zheng2024caila} uses CLIP as a visual coder, which inserts three adapters in the middle of the network to capture the state, object, and composition semantic information, respectively. It is worth mentioning that the three-branch network has become a common infrastructure for most approaches, \textit{i.e.}, three sub-networks extract state, object, and composition visual features, respectively~\cite{learning2022decomposable}.

\subsubsection{Attention Separation}

The motivation for attention separation is to dissect semantic features by capturing the parts of objects and states attended to separately. OADis~\cite{saini2022disentangling} introduces a cross-sample attention mechanism to refine features. It equips each sample with two companion samples, one with the same object and different state, and the other with the same state and different object. Afterwards, cross attention is performed with the corresponding samples separately to refine the semantic features. ADE~\cite{hao2023learning} adopts a similar training paradigm. Unlike pre-matched pairs, Jing et al.~\cite{jing2024retrieval} utilize image retrieval techniques to find suitable matching samples for object and state branches, after which they are used to enhance the representations. Zhang et al.~\cite{zhang2022learning} propose a gradient-based attention decomposition method. They view compositional recognition as a domain generalization problem, and one of the solutions is to learn domain-invariant features. To this end, they measure the variation of features by evaluating the difference in the back-propagation gradients of sample pairs to recognize the domain-invariant part. DFSP~\cite{lu2023decomposed} firstly decouples text embeddings by prototype computation and, after that, uses the text embeddings and image features for cross-attention to decompose the visual features. Similarly, Troika~\cite{huang2024troika} fuses multi-path prompt and cross attention to optimize text embeddings.

\subsection{Dependency Modeling}

Decoupling techniques pursue the pure semantic representations while mitigating the bias toward seen compositional concepts, introducing the bias toward the single concept, \textit{i.e.}, state or object. This bias comes from the fact that the same concept is represented differently in diverse contexts. Due to the existence of complex acting mechanisms in the real world, such as physical systems and biochemical reactions, the visual presentation of the same state on various objects varies greatly. For example, \textit{old} is represented as metal rust spots on \textit{car}, while it is shown as skin wrinkles on \textit{person}. Obviously, concise text annotations cannot capture such subtle differences, and the independent visual representations learned by the model from the seen domain are biased, hindering the recognition accuracy of the unseen domain. To address this problem, many studies are devoted to modeling the dependency relationship between states and objects. By introducing a priori information about states or objects to correct each other's semantic representations or classification confidence, they can be better generalized to the unseen domain. In short, the goal of dependency modeling is to learn the conditional probabilities $p(s|\psi_s(x), o)$ and $p(o|\psi_o(x), s)$. The relevant technologies, according to the implementation objectives, can be categorized into \textit{Feature Optimization} and \textit{Aid Classification}.

\subsubsection{Feature Optimization}

\begin{figure}[t]
    \centering
    \begin{minipage}{\linewidth}
    \centerline{\includegraphics[width=1.0\textwidth]{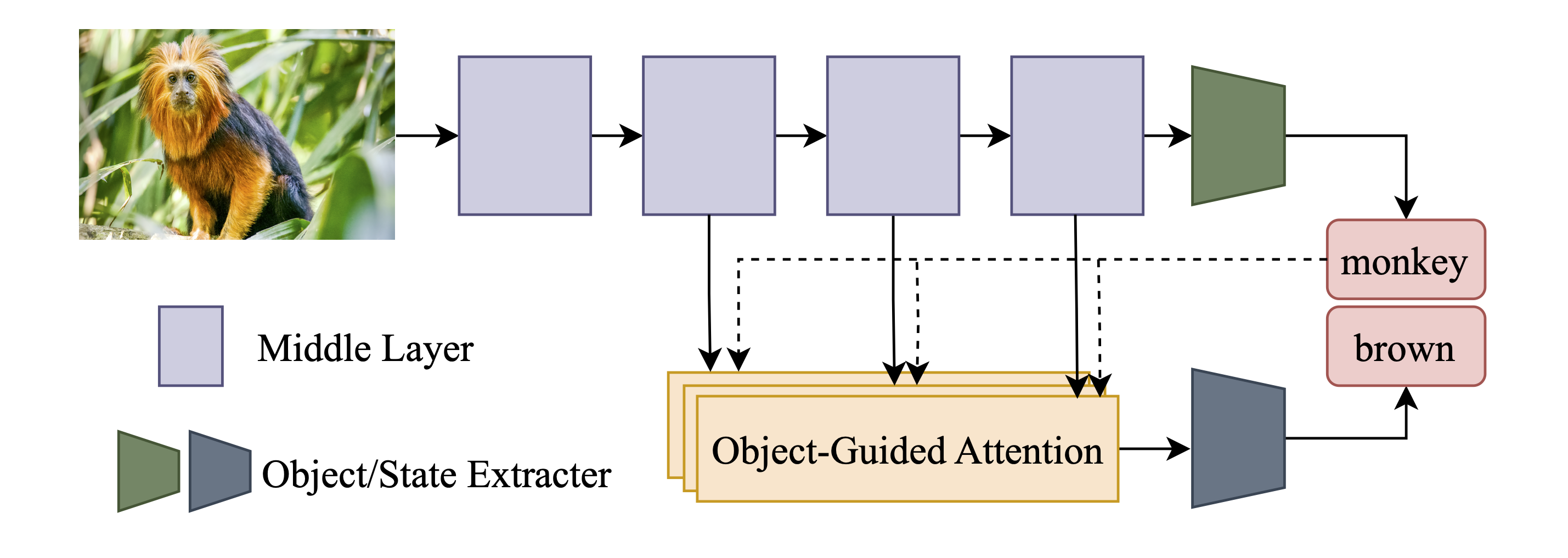}}
    \end{minipage}
\vspace{-3mm}
    \caption{A schematic diagram of dependency modeling~\cite{kim2023hierarchical}. Object information is used as a signal to guide the extraction process of state features.}
    \label{fig:dependency}
\vspace{-4mm}
\end{figure}

The goal of feature optimization is to adjust each other's feature representations by leveraging the semantic relevance of states and objects. Decoupled state/object features are expected to capture dependent semantic details while maintaining their discriminative semantic information. Visual feature optimization corrects the visual embedding distribution to avoid overfitting the seen domain caused by the model focusing on a single visual feature. For example, Panda et al.~\cite{panda2024compositional} set up multiple encoders between the state and object branching networks to transmit object features to the state network to model the dependency of the state on the object. The same idea is taken by CoT~\cite{kim2023hierarchical}, which uses object information as cueing signals while introducing an attention-based method to capture state-related visual features (Fig. \ref{fig:dependency}). Another approach is to optimize text features. Since text annotations are usually pre-extracted, they can hardly perceive dependency relationships. To this end, Nagarajan et al.~\cite{nagarajan2018attributes} model states as operators, \textit{i.e.}, encode the state as a mapping function rather than an embedding, and thus explicitly learn the semantic embedding of an object "after it has been operated on". SymNet~\cite{li2020symmetry} introduces two sub-networks, the decoupling network and the coupling network, which are used to learn independent and compositional representations of objects and states. CANet~\cite{wang2023learning} recognizes the object labels of an image first and then corrects the text embedding of the state by using the object text embedding as a priori condition.

\subsubsection{Aid Classification}

Compared to adjusting the feature distribution, aid classification aims to directly introduce semantic information of objects/states to make the classifier more discriminative. In earlier work, Chen et al.~\cite{chen2014inferring} utilize support vector machines (SVMs) for compositional recognition. They learn the weights of SVMs by capturing state dependencies on objects and generalize to unseen compositions. SAD-SP~\cite{liu2023simple} takes an intuitive approach by introducing an attention module in addition to the conventional decoupling network. State features are used to predict object soft labels and vice versa. The predicted soft labels are then used to weigh the predicted output of the conventional network. ProCC~\cite{huo2024procc} performs information exchange at each layer of the state/object classifier. Specifically, it passes the output of one layer of the state classifier through a one-dimensional convolution and softmax, and then the output of the corresponding layer of the object classifier is multiplied with it and used as the input of the next layer, and vice versa.

\subsection{Relationship Modeling}

Dependency modeling promotes the generalizability of models by exploring the intrinsic one-to-one state-object connections. Naturally, we wonder whether many-to-many complex mesh relationships, \textit{i.e.}, interactions between multiple objects or states, contribute to model performance. Such relationships describe the latent logical relations of elements in the real world due to various factors, such as subject affiliation, visual similarity, co-occurrence, etc. To name a few, we would spontaneously categorize \textit{broccoli} and \textit{spinach}, separating \textit{beef} since they belong to different subjects; a \textit{fork} and a \textit{spoon} are more visually similar than an \textit{umbrella}, and also they have a higher probability of co-occurrence. Intuitively, an understanding of these relations facilitates logical inference. It is because humans have a remarkable ability to draw analogies, make connections, and imagine, thus reasoning and recognizing by tapping into the relationships between different elements. When we see a \textit{furry black bear} and know the similarities between \textit{polar bears} and \textit{black bears}, it is easy to migrate the state of \textit{furry} to \textit{polar bears}. Driven by this idea, relationship modeling aims to enhance the inference and representation capabilities of models by uncovering and maintaining potential relationships between states and objects. The dominant techniques can be summarized as \textit{Cross Attention}, \textit{Graph modeling}, and \textit{Contrastive Embedding}.

\subsubsection{Cross Attention}

To model the relationship, cross attention is used to capture similarities between text embeddings and thus correct text representations. BMPNet~\cite{xu2021relation} introduces the cross attention mechanism to improve independent object/state representations. Specifically, for a state, the attention weights of the state and other states are first computed, and then the attention weights of an object are computed, resulting in an enhanced state representation. The object ditto. Unlike object/state relational modeling, CAPE~\cite{khan2023learning} aims to reveal co-occurrence relations between compositions. It adapts the compositional features by calculating the attention weights between the text embeddings of the compositions.

\subsubsection{Graph Modeling}

As mentioned in the previous section, graphs are excellent tools for structured data and thus can be well adapted to the relationship modeling task. In earlier work, Cruz et al.~\cite{santa2018neural} develop a logic tree to model relations between states and objects. They want the model to learn the notion of connectives such as \{\textit{and, or, not}\} to understand the potential connections between nouns. CGE~\cite{naeem2021learning} models the entire set of objects and the set of states. It treats each object and state as a graph node and then identifies edges between nodes by the composition labels. Graph convolutional network is later introduced to propagate the information. CVGAE~\cite{anwaar2022leveraging} expects the model to be able to determine on its own whether there is a connection between objects and states and thus be adaptable to the open-world compositional recognition task. It designs a reconstruction task where the reconstruction network recovers the adjacency matrix based on latent variables. GIPCOL~\cite{xu2024gipcol}, on the other hand, fuses prompt learning and graph modeling. It employs CLIP as the backbone and takes the graph-encoded text embeddings as prompts to train. The graph neural network is involved in the learning process, thereby optimizing the text embeddings to fit the task better.

\subsubsection{Contrastive Embedding}

%\begin{figure}[t]
%    \centering
%    \begin{minipage}{\linewidth}
%    \centerline{\includegraphics[width=0.9\textwidth]{fig/contrastiveCZSL.pdf}}
%    \end{minipage}
%    \caption{A schematic diagram of contrastive %embedding~\cite{wei2019adversarial}.}
%    \label{fig:contrastiveCZSL}
%\end{figure}

Contrastive embedding utilizes contrastive learning to optimize the semantic space and implicitly learn the relationships between samples. Contrastive learning~\cite{jaiswal2020survey}, as a representative unsupervised learning method, demonstrates powerful semantic representational ability. It aims at clustering similar samples together and separating different samples. The learning mode is to constrain the semantic distance by comparing positive and negative examples, and the learned semantic distance reflects the affinity between the samples, \textit{i.e.}, the semantic relationship. Therefore, many areas exploit contrast learning to optimize semantic embeddings. Unlike the attention mechanism and graph modeling described above, contrast learning does not pre-define the relationship between samples explicitly but rather iteratively adjusts the semantic distances between samples to place them in appropriate positions. The performance of contrast learning largely depends on the number and quality of negative examples, and therefore, negative sampling is another critical aspect of contrastive learning.

Wei et al.~\cite{wei2019adversarial} propose a quintuplet loss to optimize the text embedding space. Specifically, they use a sub-network to synthesize state and object texts into compositions. After that, the label of an image is used as a positive example and supplemented with three negative examples for contrastive learning. The negative examples are text compositions with the same state but different objects, the same object but different states, and different objects and states with the target image. SCEN~\cite{li2022siamese} performs contrastive learning in the visual semantic space. It defines two databases to sample the negative examples, one with the same state as the target image but with different objects, and the other with the same object but with different states. After that, contrastive learning is implemented in the decoupled state/object embedding space. Yang et al.~\cite{yang2023dual} adopt a similar approach to explore the relations among states or objects. Meanwhile, they introduce instance-level contrast learning to optimize the compositional embedding space.

\subsection{Other Categories}

In addition to the above mainstream techniques, many other studies provide unique perspectives to address the problem of compositional recognition. For example, TMN~\cite{purushwalkam2019task} presents a gated network that recognizes different compositions by injecting state and object information into the network to determine the switching of pathways. Similarly, TAFE-Net~\cite{wang2019tafe} proposes a task-aware learning paradigm that guides the studying process by using task information as cues. Some methods are devoted to analyzing the imbalance issue. MUST~\cite{jiang2024mutual} argues that the difficulty of recognizing elements varies, e.g., \textit{mashed banana} is hard to be identified as \textit{banana} while relatively easy to be identified as \textit{mashed}. For this reason, it introduces a weighted loss to balance the effects of different elements. ProLT~\cite{jiang2024revealing}, on the other hand, points out that the data imbalance problem caused by different sample sizes of elements affects the recognition accuracy. To this end, it similarly proposes a novel weighted loss to solve this problem. CDS-CZSL~\cite{li2024context} argues that for an object, the importance of different states is different. For example, \textit{sliced} is more important than \textit{red} for \textit{strawberry}, cause \textit{red} is a more general state that can be reflected in a wide range of categories. In this regard, it first highlights those important states by clustering and contrasting, and then a specificity-refined primitive loss is implemented. Moreover, several studies develop new tasks to address larger challenges~\cite{karthik2022kg, hu2024dynamic}.

\section{Foundation Model-Based Open-World Recognition}
\label{section:fbor}

\subsection{Problem Description}

Large Vision-Language Models (VLMs) are advanced deep learning models that seamlessly integrate visual and language processing capabilities~\cite{zhang2024vision}. By analyzing extensive datasets of image-text pairs, these models learn the intricate relationships between vision and language, allowing them to understand and generate detailed language descriptions of image content~\cite{radford2021learning}. VLMs are renowned for their robust cross-modal understanding and nuanced element perception, which make them particularly effective for zero-shot adaptation tasks. It is only necessary to associate the text description with the class name of the downstream recognition task, e.g., \textit{a photo of a [class]}, to achieve the purpose of classification and yield excellent results. However, recent studies suggest the capabilities of VLMs have not been fully exploited. Thus the foundation model-based open-world recognition receives more attention. 

To provide a deep understanding of this topic, we first compare several prominent VLM-related research tasks. 
First, \textit{Better VLMs} explores to improve the pre-training process of VLMs. The motivation for most studies stems from two aspects, \textit{i.e.}, data efficiency and learning efficiency. Data-efficient pre-training aims to reduce the data required while preserving strong representations~\cite{jia2021scaling, cui2022contrastive}, since large datasets, like CLIP's 400 million image-text pairs, are resource-heavy. Learning-efficient pre-training is to explore better alternative training methods like reconstruction~\cite{singh2022flava}, generation~\cite{huang2023nlip}. 
% Upgrading for VLMs has largely been locked into the exploration of pre-training approaches. 
% The motivation for most studies stems from two aspects, data-efficient pre-training and learning-efficient pre-training. Data-efficient pre-training aims to reduce the data required while preserving strong representations~\cite{jia2021scaling, cui2022contrastive}, since large datasets, like CLIP’s 400 million image-text pairs, are resource-heavy. Learning-efficient pre-training is to explore better alternative training methods like reconstruction~\cite{singh2022flava}, generation~\cite{huang2023nlip}.
Next, \textit{Few-Shot Adaptation and OOD Detection}, in which the former involves fine-tuning pre-trained VLMs with limited labeled data to create specialized models for specific tasks~\cite{zhou2022learning, zhou2022conditional}, which the latter aims to identify samples not belonging to the target categories~\cite{esmaeilpour2022zero, wang2023clipn}. Note that the OOD does not need to recognize the specific category of the foreign samples. 
% Few-shot adaptation involves fine-tuning pre-trained VLMs with limited labeled data to create specialized models for specific tasks~\cite{zhou2022learning, zhou2022conditional}. Out-of-distribution (OOD) detection aims to identify samples not belonging to the target categories~\cite{esmaeilpour2022zero, wang2023clipn}. Note that the OOD does not need to recognize the specific category of the foreign samples.
Notably, both few-shot adaptation and OOD detection require labeled data to support fine-tuning VLMs. In contrast, \textit{Zero-Shot Adaptation} releases this limitation. It consists of three basic tasks that apply VLMs to open environment recognition tasks under different constraints, including \textit{i) Zero-shot Prediction}: with only the category names of the task known and without any image data applying VLMs to the downstream task; \textit{ii) Test-time Adaptation}: know the category names of the task and allow adjustments at the online testing phase; and \textit{iii) Unsupervised Adaptation}: Know the category names of the task and have part or entire of the unlabeled test data in advance. 
% \textit{Zero-Shot Adaptation.} Both few-shot adaptation and OOD detection require labeled data as support to fine-tune VLMs, and zero-shot adaptation releases this limitation. Zero-shot adaptation consists of three basic tasks that apply VLMs to open environment recognition tasks under different constraints. \textbf{Zero-shot prediction}: With only the category names of the task known and without any image data applying VLMs to the downstream task. \textbf{Test-Time Adaptation}: Know the category names of the task and allow adjustments at the online testing phase. \textbf{Unsupervised Adaptation}: Know the category names of the task and have part or entire of the unlabeled test data in advance.
In this survey, we refer to the three tasks of zero-shot adaptation, i.e., \textit{i)}, \textit{ii)} and \textit{iii)}, as the category of foundation model-based open recognition.

\begin{figure}[t]
    \centering
    \begin{minipage}{\linewidth}
    \centerline{\includegraphics[width=0.8\textwidth]{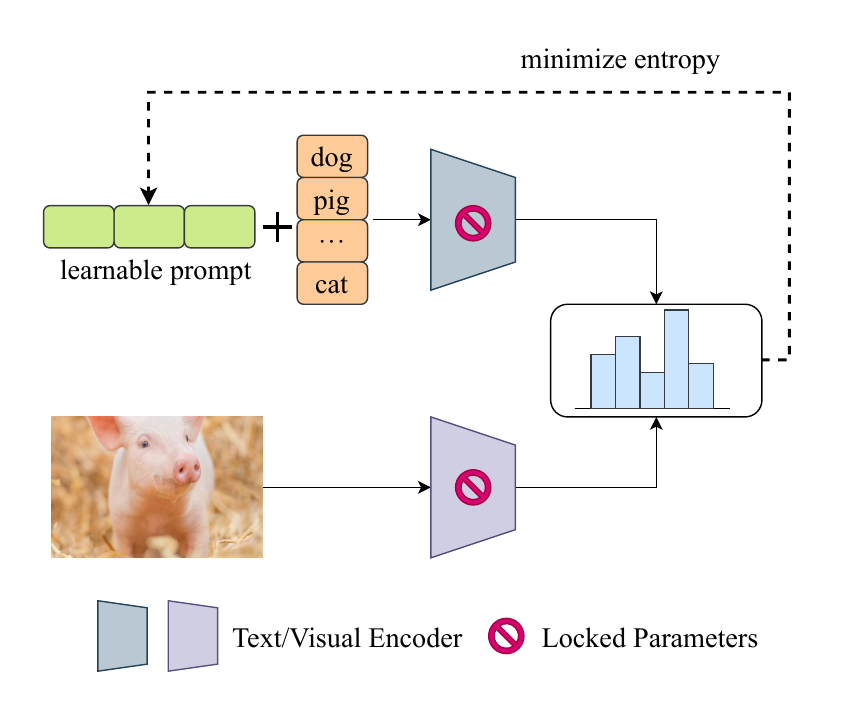}}
    \end{minipage}
\vspace{-4mm}
    \caption{A schematic diagram of test-time prompt learning~\cite{shu2022test}. Adjust the prompt by minimizing the entropy of the output.}
    \label{fig:TTP}
\vspace{-5mm}
\end{figure}

\subsection{Zero-Shot Prediction}

VLMs have powerful zero-shot prediction capabilities. For a specific task, simply editing the category names into the prompt, e.g., \textit{a photo of [class]}, enables direct classification without fine-tuning. However, there is a performance gap between such a deployment and a model that is fully supervised and trained on the downstream task. To address this issue, zero-shot prediction presents promising solutions. A prevalent solution is to design better prompts. The commonly used \textit{a photo of [class]}, while intuitive, is missing a lot of detailed information. For this reason, Menon et al.~\cite{menon2022visual} propose to use a large language model such as GPT-3~\cite{brown2020language} to generate multiple word descriptions for classes. In the testing time, the similarity between the image sample and each word description is calculated, and the mean value is taken as the discriminative basis. CuPL~\cite{pratt2023does} has a similar idea, which uses this description directly as the prompt for testing. CHiLS~\cite{novack2023chils} also adopts a large language model, which generates multiple subclasses for each category and uses their combined similarity as the discriminative basis. APPLe~\cite{chendon} uses the GPT-3 descriptions to initialize prompt prototypes as the category classifiers. Unlike the design of prompt, SuS-X~\cite{udandarao2023sus} uses Diffusion~\cite{rombach2022high} to pre-generate multiple pseudo-visual samples for each class, which serves as a support set to assist the judgment at the testing time. CALIP~\cite{guo2023calip} takes the perspective of embedding space, and it introduces the attention mechanism to capture the correlation between text embeddings and visual embeddings, which effectively improves the zero-shot generalization capability.

\subsection{Test-Time Adaptation}

The purpose of Test-Time Adaptation (TTA) is to dynamically adjust the model to progressively adapt to the downstream task during the testing process. The mainstream schemes are categorized into prompt adaptation and cache bank. Prompt adaptation is a technique that combines prompt learning and TTA. During testing, the prompt is set as learnable parameters and the parameters of the model are fixed. The model continuously optimizes the prompt representation and thus adapts to the target task. TPT~\cite{shu2022test} first presents this concept by augmenting the test image into multiple views and then filtering out those with low confidence. The confidence outputs of the remaining augmented samples are averaged as the final prediction, and at the same time, the information entropy is minimized to train the prompt (Fig. \ref{fig:TTP}). This paradigm provides a basic template for subsequent research on prompt adaptation. DiffTPT~\cite{feng2023diverse} argues that the augmentation approach of TPT lacks diversity and, therefore, introduces the diffusion model to generate richer samples for the test image. PromptAlign~\cite{abdul2024align} argues that adjusting the prompt should be done while aligning the distribution of the target and pre-training data. Therefore it introduces additional loss to align the statistics of target and pre-training data, \textit{i.e.}, mean and variance. SwapPrompt~\cite{ma2024swapprompt} and DART~\cite{liu2024dart} use the exponential moving average algorithm to update the prompt.

The goal of the Cache bank is to store the tested samples to improve the performance of subsequent tests. An intuitive approach is to cache the previous high-confidence samples as the support set to be queried by subsequent samples. DMN-ZS~\cite{zhang2024dual} introduces an attention mechanism to compute the similarity between the new test samples and the support set to obtain the weighted features. TDA~\cite{karmanov2024efficient} sets up two cache banks to store the one-hot labels and the negative soft labels of the reliable samples, respectively. The new test samples interact with them to generate predictions. Zhang et al.~\cite{zhang2024robust} employ a cache bank to save high-confidence samples from learning high-quality prompts. 
In addition, MTA~\cite{zanella2024test} proposes a new loss function to optimize visual features in the semantic space.

\subsection{Unsupervised Adaptation}

Unsupervised adaptation is suitable for offline testing environments where some or all test data is available. Although there are no labels available, image samples can provide a priori information to help better prediction. Existing research combines different algorithms to address this problem. UPL~\cite{huang2022unsupervised} fuses prompt learning and self-supervised learning to learn prompt representations through a small number of high-quality pseudo-labels. MUST~\cite{li2022masked} incorporates image reconstruction techniques to facilitate the model's understanding of visual information. InMaP~\cite{qian2024intra} introduces prototype learning, which mines a set of visual prototypes from the visual space to replace the text embeddings for prediction. LaFTer~\cite{mirza2024lafter} trains a text classifier and then uses it to classify images and trains the image encoder with pseudo-labels. ZLaP~\cite{kalantidis2024label} presents a label propagation algorithm to assign labels to images. In addition, CLIPPR~\cite{kahana2022improving} introduces a label distribution prior to predicting the data of the target domain.

\section{Technical Details}
\label{section:technical details}

\begin{table}[t]
    \renewcommand{\arraystretch}{0.8}
    \centering
    \caption{A list of commonly used benchmark datasets for zero-shot image recognition.}
    \centering
    \resizebox{\linewidth}{!}{
    \begin{tabular}{l c c c c }
        \toprule
        \textbf{Dataset} & \textbf{Year} & \textbf{Images} & \textbf{Class/Attr.} & \textbf{Download}\\
        \midrule
        \multicolumn{5}{c}{Zero-Shot Object Recognition}\\
        \midrule
         CUB~\cite{wah2011caltech} & 2011 & 11,788  & 200/312 & \href{https://www.vision.caltech.edu/datasets/cub_200_2011/}{\textcolor{blue}{[Link]}} \\
         Flowers102~\cite{nilsback2008automated} & 2008 & 8,189  & 102/- & \href{https://www.robots.ox.ac.uk/~vgg/data/flowers/102/}{\textcolor{blue}{[Link]}} \\
         SUN~\cite{patterson2012sun}  & 2012 & 14,340 & 717/102 & \href{https://cs.brown.edu/~gmpatter/sunattributes.html}{\textcolor{blue}{[Link]}}\\
        NABirds~\cite{van2015building}  & 2015 & 48,562 & 1011/- & \href{https://dl.allaboutbirds.org/nabirds}{\textcolor{blue}{[Link]}}\\
        DeepFashion~\cite{liu2016deepfashion} & 2016 & 289,222  & 46/1000 & \href{https://mmlab.ie.cuhk.edu.hk/projects/DeepFashion.html}{\textcolor{blue}{[Link]}}\\
        AWA~\cite{lampert2013attribute}  & 2013 & 30,475  & 50/85 & \href{https://cvml.ista.ac.at/AwA/}{\textcolor{blue}{[Link]}}\\
        AWA2~\cite{xian2018zero} &  2018 & 37,322  & 50/85 & \href{https://cvml.ista.ac.at/AwA2/}{\textcolor{blue}{[Link]}}\\
        APY~\cite{farhadi2009describing} &  2009 & 15,339  & 32/64 & \href{https://vision.cs.uiuc.edu/attributes/}{\textcolor{blue}{[Link]}}\\
        \midrule
        \multicolumn{5}{c}{Zero-Shot Compositional Recognition}\\
        \midrule
        UT-Zappos~\cite{yu2014fine} & 2014 & 50,025  & 12/16 & \href{https://vision.cs.utexas.edu/projects/finegrained/}{\textcolor{blue}{[Link]}} \\
        MIT-States~\cite{isola2015discovering} & 2015 & 53,753  & 245/115 & \href{https://web.mit.edu/phillipi/Public/states_and_transformations/index.html}{\textcolor{blue}{[Link]}} \\
        C-GQA~\cite{naeem2021learning} & 2021 & 39,298  & 674/413 & \href{https://s3.mlcloud.uni-tuebingen.de/czsl/cgqa-updated.zip}{\textcolor{blue}{[Link]}} \\
        %StanfordVRD~\cite{lu2016visual} & 2016 & 11,788  & 200/312 & \href{}{\textcolor{blue}{[Link]}} \\
        AO-CLEVr~\cite{atzmon2020causal} & 2020 & 180,000  & 3/8 & \href{https://drive.google.com/drive/folders/1BBwW9VqzROgJXmvnfXcOxbLob8FB_jLf}{\textcolor{blue}{[Link]}} \\
        VAW-CZSL~\cite{Saini_2022_CVPR} & 2022 & 92,583  & 541/440 & \href{https://drive.google.com/drive/folders/1CalwDXkkGALxz0e-aCFg9xBmf7Pu4eXL}{\textcolor{blue}{[Link]}} \\
        Clothing16K & 2020 & 16,170  & 8/9 & \href{https://www.kaggle.com/datasets/kaiska/apparel-dataset}{\textcolor{blue}{[Link]}} \\
        \midrule
        \multicolumn{5}{c}{Foundation model-Based Open-world Recognition}\\
        \midrule
        ImageNet~\cite{deng2009imagenet} & 2009 & 1,431,167 & 1000 & \href{https://image-net.org/download.php}{\textcolor{blue}{[Link]}} \\
        Caltech101~\cite{fei2004learning} & 2004 & 9,145  & 101 & \href{https://www.vision.caltech.edu/datasets/}{\textcolor{blue}{[Link]}} \\
        Caltech256~\cite{griffin2007caltech} & 2007 & 20,607  & 256 & \href{https://www.vision.caltech.edu/datasets/}{\textcolor{blue}{[Link]}} \\
        SUN397~\cite{xiao2010sun} & 2010 & 108,753  & 397 & \href{https://3dvision.princeton.edu/projects/2010/SUN/}{\textcolor{blue}{[Link]}} \\
        Food101~\cite{bossard2014food} & 2014 & 101,000  & 101 & \href{https://s3.amazonaws.com/fast-ai-imageclas/food-101.tgz}{\textcolor{blue}{[Link]}} \\
        Flowers102~\cite{nilsback2008automated} & 2008 & 8,189  & 102 & \href{https://www.robots.ox.ac.uk/~vgg/data/flowers/102/}{\textcolor{blue}{[Link]}} \\
        StanfordCars~\cite{krause2013collecting} & 2013 & 16,185  & 197 & \href{https://s3.amazonaws.com/fast-ai-imageclas/stanford-cars.tgz}{\textcolor{blue}{[Link]}} \\
        FGVCAircraft~\cite{maji2013fine} & 2013 & 10,000  & 100 & \href{https://www.robots.ox.ac.uk/~vgg/data/fgvc-aircraft/}{\textcolor{blue}{[Link]}} \\
        OxfordPets~\cite{parkhi2012cats} & 2012 & 7,349  & 37 & \href{https://www.robots.ox.ac.uk/~vgg/data/pets/}{\textcolor{blue}{[Link]}} \\
        DTD~\cite{cimpoi2014describing} & 2014 & 5,640  & 47 & \href{https://www.robots.ox.ac.uk/~vgg/data/dtd/}{\textcolor{blue}{[Link]}} \\
        EuroSAT~\cite{helber2019eurosat} & 2019 & 27,000  & 10 & \href{https://zenodo.org/records/7711810\#.ZAm3k-zMKEA}{\textcolor{blue}{[Link]}} \\
        Country211~\cite{radford2021learning} & 2021 & 64,300  & 211 & \href{https://openaipublic.azureedge.net/clip/data/country211.tgz}{\textcolor{blue}{[Link]}} \\
        CIFAR-10~\cite{krizhevsky2009learning} & 2009 & 60,000  & 10 & \href{https://www.cs.toronto.edu/~kriz/cifar.html}{\textcolor{blue}{[Link]}} \\
        CIFAR-100~\cite{krizhevsky2009learning} & 2009 & 60,000  & 100 & \href{https://www.cs.toronto.edu/~kriz/cifar.html}{\textcolor{blue}{[Link]}} \\
        Birdsnap~\cite{berg2014birdsnap} & 2014 & 49,829  & 500 & \href{https://thomasberg.org/}{\textcolor{blue}{[Link]}} \\
        \bottomrule

         \multicolumn{5}{p{9cm}}{\footnotesize{\textit{[Class/Attr.] means the number of classes and attributes for ZSOR, objects and states for ZSCR, classes for FBOR. Click the \textcolor{blue}{[link]} to check more information in the corresponding repository.}}} \\
         \end{tabular}}
    \label{tab:datasets}
\vspace{-4mm}
\end{table}

\begin{table*}[htbp]
    \newcolumntype{M}[1]{>{\centering\arraybackslash}m{#1}}
    \renewcommand{\arraystretch}{0.8}
    \centering
    \caption{More details on selected advances for zero-shot object recognition.}
    \centering
    %\resizebox{\textwidth}{!}{
    \resizebox{0.88\textwidth}{40mm}{ 
    \begin{tabular}{c| c| l| c| c| c| c| l| l| c }
        \toprule
        \textbf{Topic} & \textbf{Tax.} & \textbf{Name} & \textbf{Venue} & \textbf{Year} & \textbf{Backbone} & \textbf{Aux.} & \textbf{KeyWords} & \textbf{Dataset} & \textbf{Code}\\
        \midrule
         \multirow{41}{*}{\rotatebox[origin=c]{90}{Zero-Shot Object Recognition}} & \multirow{18}{*}{\rotatebox[origin=c]{90}{Visual}} & LH2B~\cite{elhoseiny2017link} & CVPR & 2017  & VGG16 & Region & Detection & CUB, NAB& \href{https://github.com/EthanZhu90/ZSL_PP_CVPR17}{\textcolor{blue}{[Link]}} \\
         & & S2GA~\cite{ji2018stacked} & NeurIPS & 2018  & VGG16 & Region & Detection & CUB, NAB& \href{https://github.com/ylytju/sga/tree/master}{\textcolor{blue}{[Link]}} \\
         & & AGAA~\cite{zhu2018generative} & CVPR & 2018  & VGG16 & Region & Detection & CUB, NAB& \href{https://github.com/EthanZhu90/ZSL_GAN}{\textcolor{blue}{[Link]}} \\
         & & LDF~\cite{li2018discriminative} & CVPR & 2018  & GNet, VGG19 & -& Crop & CUB, AWA& \href{https://github.com/zbxzc35/Zero_shot_learning_using_LDF_tensorflow}{\textcolor{blue}{[Link]}}\\
         & & AREN~\cite{xie2019attentive} & CVPR & 2019  & ResNet101 &- & Attention & CUB, SUN,AWA2, APY& \href{https://github.com/gsx0/Attentive-Region-Embedding-Network-for-Zero-shot-Learning}{\textcolor{blue}{[Link]}} \\
         & & SGMA~\cite{zhu2019semantic} & NeurIPS & 2019  & VGG19&-  & Crop & CUB, FLO, AWA& \href{https://github.com/wuhuicumt/LearningWhereToLook/tree/master}{\textcolor{blue}{[Link]}}\\
         & & CANZSL~\cite{chen2020canzsl} & WACV & 2020 & VGG16 & Region & Consistency & CUB, NAB & 
          \href{https://github.com/uqzhichen/CANZSL}{\textcolor{blue}{[Link]}} \\
         & & RGEN~\cite{xie2020region}  & ECCV & 2020 & ResNet101 &- & Attention, Graph & CUB, SUN, AwA2, APY& -\\
         & & SDGZSL~\cite{chen2021semantics} & ICCV & 2021 & ResNet101 & - & Disentangling & CUB, AWA2, APY, FLO & 
         \href{https://github.com/uqzhichen/SDGZSL}{\textcolor{blue}{[Link]}} \\
        & & RSAN~\cite{wang2021region}  & CIKM & 2021 & ResNet101 &- & Attention & CUB, SUN, AWA2& -\\
        & & SR2E~\cite{ge2021semantic} & AAAI & 2021  & ResNet101 &- & Crop & CUB, SUN, AWA2, APY& - \\
        & & VGSE-SMO~\cite{xu2022vgse} & CVPR & 2022 & ResNet50 &- & Crop, Clustering & CUB, SUN, AWA2& \href{https://github.com/wenjiaXu/VGSE}{\textcolor{blue}{[Link]}} \\
        & & GSMFlow~\cite{chen2022gsmflow} & TMM & 2022 & ResNet101 & - & Generation Shifts & CUB, AWA2, APY, FLO & \href{https://github.com/uqzhichen/GSMFlow_TMM}{\textcolor{blue}{[Link]}} \\
        & & ERPCNet~\cite{li2022entropy} & TCSVT & 2022  & ResNet101 &- & Crop & CUB, SUN, AWA2, APY& - \\
        & & GKU~\cite{guo2023graph} & AAAI & 2023  & ResNet34 & Region & Graph & CUB, NAB& - \\
        & & EOPA~\cite{chen2023explanatory} & TPAMI & 2023  & ANet, ResNet50 &- & Graph & CUB, SUN, FLO, AWA2& - \\
        & & POPRNet~\cite{liu2024part} & TIP & 2024  & ResNet101 & -& Crop & CUB, SUN, AWA2& \href{https://github.com/ManLiuCoder/POPRNet}{\textcolor{blue}{[Link]}} \\
        & & ZSLViT~\cite{chen2024progressive} & CVPR & 2024  & ViT-B & - & Attention & CUB, SUN, AWA2& - \\
        \cmidrule{2-10}
        % zsor ==================== visual-attribute===============
        & \multirow{19}{*}{\rotatebox[origin=c]{90}{Visual-Attribute}} & DAZLE~\cite{huynh2020fine}  & CVPR & 2020  & ResNet101 &- & Attention & CUB, SUN, AWA2, DEP& \href{https://github.com/hbdat/cvpr20_DAZLE}{\textcolor{blue}{[Link]}}\\
        & & Composer~\cite{huynh2020compositional} & NeurIPS & 2020  & ResNet101 &- & Attention, Generation & CUB, SUN, AWA2, DEP& \href{https://github.com/hbdat/neurIPS20_CompositionZSL}{\textcolor{blue}{[Link]}} \\
        & & APN~\cite{xu2020attribute} & NeurIPS & 2020  & ResNet101 &- & Prototype & CUB, SUN, AWA2& \href{https://github.com/wenjiaXu/APN-ZSL/tree/master}{\textcolor{blue}{[Link]}} \\
        & & GEM~\cite{liu2021goal} &  CVPR & 2021  & ResNet101 & Region & Attention & CUB, SUN, AWA2& \href{https://github.com/osierboy/GEM-ZSL}{\textcolor{blue}{[Link]}}\\
        & & DPPN~\cite{wang2021dual} & NeurIPS & 2021  & ResNet101 &- & Prototype & CUB, SUN, AWA2, APY& \href{https://github.com/Roxanne-Wang/DPPN-GZSL}{\textcolor{blue}{[Link]}} \\
        & & LAPE~\cite{wang2022language} & TCSVT & 2022  & ResNet101 & - & Attention & CUB, SUN, AWA2, APY & - \\
        & & RIAE~\cite{hu2022region} & IS & 2022 & ResNet101 &- & Attention, Graph & CUB, SUN, AWA2& - \\
        & & DPDN~\cite{ge2022dual} & MM & 2022  & ResNet101 & -& Attention, Prototype & CUB, SUN, AWA2& - \\
        & & GNDAN~\cite{chen2022gndan} & TNNLS & 2022  & ResNet101 & -& Attention, Graph & CUB, SUN, AWA2& \href{https://github.com/shiming-chen/GNDAN}{\textcolor{blue}{[Link]}} \\
        & & AREES~\cite{liu2022zero} & TNNLS & 2022  & ResNet101 &- & Attention, Generation & CUB, SUN, AWA, AWA2, APY& - \\
        & & MSDN~\cite{chen2022msdn} &  CVPR & 2022  & ResNet101 &- & Attention & CUB, SUN, AWA2& \href{https://github.com/shiming-chen/MSDN}{\textcolor{blue}{[Link]}}\\
        & & TransZero~\cite{chen2022transzero} & AAAI & 2022  & ResNet101 &- & Attention & CUB, SUN, AWA2& \href{https://github.com/shiming-chen/TransZero}{\textcolor{blue}{[Link]}} \\
        & & I2DFormer~\cite{naeem2022i2dformer} & NeurIPS & 2022  & ViT-B &- & Attention & CUB, FLO, AWA2& \href{https://github.com/ferjad/I2DFormer}{\textcolor{blue}{[Link]}} \\
        & & DUET~\cite{chen2023duet} & AAAI & 2023  & ViT-B &- & Attention, Contrast & CUB, SUN, AWA2& \href{https://github.com/zjukg/DUET}{\textcolor{blue}{[Link]}} \\
        & & PSVMA~\cite{liu2023progressive} & CVPR & 2023  & ViT-B &- & Attention & CUB, SUN, AWA2& \href{https://github.com/ManLiuCoder/PSVMA}{\textcolor{blue}{[Link]}} \\
        & & HRT~\cite{cheng2023hybrid} & PR & 2023  & ResNet101 &- & Attention & CUB, SUN, AWA2& - \\
        & & CC-ZSL~\cite{cheng2023discriminative} & TCSVT & 2023  & ResNet101 &- & Prototype, Contrast & CUB, SUN, AWA2& \href{https://github.com/KORIYN/CC-ZSL}{\textcolor{blue}{[Link]}} \\
        & & CoAR-ZSL~\cite{du2023boosting} & TNNLS & 2023  & R101, ViT-L &- & Attention, Prototype & CUB, SUN, AWA2& \href{https://github.com/dyabel/CoAR-ZSL.git}{\textcolor{blue}{[Link]}} \\
        & & GIRL~\cite{guo2023group} & IS & 2023  & ResNet101 & -& Prototype & CUB, SUN, AWA2, APY& \href{https://github.com/TingML/GIRL}{\textcolor{blue}{[Link]}} \\
        \cmidrule{2-10}
        %zsor ==========================attribute=======================================
        & \multirow{4}{*}{\rotatebox[origin=c]{90}{Attribute}} & MCZSL~\cite{akata2016multi} & CVPR & 2016  & VGG16 & Region &  Crop & CUB & - \\
        & & LFGAA~\cite{liu2019attribute} & ICCV & 2019  & GNet, R101, V19 &- & Attention & CUB, SUN, AWA2& \href{https://github.com/ZJULearning/AttentionZSL}{\textcolor{blue}{[Link]}} \\
        & & APNet~\cite{liu2020attribute} & AAAI & 2020  & ResNet101 &- & Graph & CUB, SUN, AWA, AWA2, APY& - \\
        & & DEDN~\cite{rao2024dual} & IJCAI & 2024  & ResNet101 &- & Attention, Distillation & CUB, SUN, AWA2& \href{https://github.com/zjrao/DEDN}{\textcolor{blue}{[Link]}} \\
        \bottomrule

         \multicolumn{10}{p{18cm}}{\footnotesize{\textit{[Tax.] means taxonomy. [Aux.] means auxiliary. [GNet] means GoogleNet, [ANet] means AlexNet, [R101] means ResNet101, [V19] means VGG19.}}} \\
         \multicolumn{10}{p{20cm}}{\footnotesize{\textit{[Region] means region-level annotations. [FLO] means Flowers102, [NAB] means NABirds, [DEP] means DeepFashion. Click the \textcolor{blue}{[link]} to check more information in the corresponding repository.}}} \\
    \end{tabular}}
    \label{tab:oject recognition of implement details}
\vspace{-4mm}
\end{table*}

\begin{table*}[htbp]
    \newcolumntype{M}[1]{>{\centering\arraybackslash}m{#1}}
    \renewcommand{\arraystretch}{0.8}
    \centering
    \caption{More details on selected advances for zero-shot compositional recognition.}
    \centering
    %\resizebox{\textwidth}{!}{
    \resizebox{0.85\textwidth}{38mm}{ 
    %\scalebox{0.8}{
    \begin{tabular}{c| c| l| c| c| c| l| l| c }
        \toprule
        \textbf{Topic} & \textbf{Tax.} & \textbf{Name} & \textbf{Venue} & \textbf{Year} & \textbf{Backbone} & \textbf{KeyWords} & \textbf{Dataset} & \textbf{Code}\\
        \midrule
         \multirow{38}{*}{\rotatebox[origin=c]{90}{Zero-Shot Compositional Recognition}} & \multirow{13}{*}{\rotatebox[origin=c]{90}{Decoupling}} & Nan et al.~\cite{nan2019recognizing} & AAAI & 2019  & ResNet18 & Contrast & UTZ, MIT & - \\
         & & CAUSAL~\cite{atzmon2020causal} & NeurIPS & 2020 & ResNet18 & Mapping Function & UTZ, AOC & \href{https://github.com/NVlabs/causal_comp}{\textcolor{blue}{[Link]}} \\
         & & ProtoProp~\cite{ruis2021independent} & NeurIPS & 2021 & ResNet18 &  Prototype, Graph & UTZ, AOC, CGQ & \href{https://github.com/FrankRuis/protoprop}{\textcolor{blue}{[Link]}} \\
         & & OADis~\cite{saini2022disentangling} & CVPR & 2022 & ResNet18  & Attention & UTZ, MIT, VAW & \href{https://github.com/nirat1606/OADis}{\textcolor{blue}{[Link]}} \\
         & & Zhang et al.~\cite{zhang2022learning} & ECCV & 2022 & ResNet18  & Attention & UTZ, AOC, CLO & \href{https://github.com/PRIS-CV/IVR}{\textcolor{blue}{[Link]}} \\
         & & DECA~\cite{learning2022decomposable} & TMM & 2022 & ResNet18  & Mapping & UTZ, MIT & \href{https://github.com/muliyangm/DeCa}{\textcolor{blue}{[Link]}} \\
         & & DRANet~\cite{li2023distilled} & ICCV & 2023 & ResNet18  & Attention & UTZ, MIT, CGQ & - \\
         & & ADE~\cite{hao2023learning} & CVPR & 2023 & ViT-B  & Attention & UTZ, CGQ, CLO & \href{https://haoosz.github.io/ade-czsl/}{\textcolor{blue}{[Link]}} \\
         & & DFSP~\cite{lu2023decomposed} & CVPR & 2023 & CLIP  & Attention, Prototype & UTZ, MIT, CGQ & \href{https://github.com/Forest-art/DFSP.git}{\textcolor{blue}{[Link]}} \\
         & & Hu et al.~\cite{hu2023leveraging} & AAAI & 2023 & ResNet18  & Contrast & UTZ, CGQ & \href{https://github.com/hxm97/SCD-CZSL}{\textcolor{blue}{[Link]}} \\
         & & CAILA~\cite{zheng2024caila} & WACV & 2024 & CLIP  & Mapping & UTZ, MIT, CGQ, VAW & \href{https:// github.com/zhaohengz/CAILA}{\textcolor{blue}{[Link]}} \\
         & & Jing et al.~\cite{jing2024retrieval} & AAAI & 2024 & CLIP  & Attention & UTZ, MIT, CGQ & - \\
         & & Troika~\cite{huang2024troika} & CVPR & 2024 & CLIP  & Attention & UTZ, MIT, CGQ & \href{https://github.com/bighuang624/Troika}{\textcolor{blue}{[Link]}} \\
        %======================= dependency===============
        \cmidrule{2-9}
         & \multirow{7}{*}{\rotatebox[origin=c]{90}{Dependency}} & Nagarajan et al.~\cite{nagarajan2018attributes} & ECCV & 2018  & ResNet18  & Contrast & UTZ, MIT & \href{https://github.com/attributes-as-operators}{\textcolor{blue}{[Link]}} \\
         & & SymNet~\cite{li2020symmetry} & CVPR & 2020  & ResNet18  & Mapping Function & UTZ, MIT & \href{https://github.com/DirtyHarryLYL/SymNet}{\textcolor{blue}{[Link]}} \\
         & & SAD-SP~\cite{liu2023simple} & TPAMI & 2023  & ResNet18 & Mapping Function & UTZ, MIT, CGQ & - \\
         & & CoT~\cite{kim2023hierarchical} & ICCV &  2023 & ResNet18, ViT-B &  Attention & MIT, CGQ, VAW & \href{https://github.com/HanjaeKim98/CoT}{\textcolor{blue}{[Link]}} \\
         & & CANet~\cite{wang2023learning} & CVPR &  2023 & ResNet18  & Mapping Function & UTZ, MIT, CGQ & \href{https://github.com/wqshmzh/CANet-CZSL}{\textcolor{blue}{[Link]}} \\
         & & Panda et al.~\cite{panda2024compositional} & PR & 2024  & ResNet18  & Graph & UTZ, MIT, CGQ & - \\
         & & ProCC~\cite{huo2024procc} & AAAI &  2024 & ResNet18  & Mapping Function & UTZ, MIT, CGQ & \href{https://github.com/huofushuo/procc}{\textcolor{blue}{[Link]}} \\
         %======================= relation ===============
        \cmidrule{2-9}
        & \multirow{8}{*}{\rotatebox[origin=c]{90}{Relationship}} & Wei et al.~\cite{wei2019adversarial} & ICCV & 2019  & ResNet18  & Contrast & UTZ, MIT & - \\
        & & BMPNet~\cite{xu2021relation} & TMM & 2021  & ResNet18  & Attention & UTZ, MIT & \href{https://github.com/daoyuan98/Relation-CZSL}{\textcolor{blue}{[Link]}} \\
        & & CGE~\cite{naeem2021learning} & CVPR & 2021  & ResNet18  & Graph & UTZ, MIT, CGQ & \href{https://github.com/ExplainableML/czsl}{\textcolor{blue}{[Link]}} \\
        & & SCEN~\cite{li2022siamese} & CVPR & 2022  & ResNet18  & Contrast & UTZ, MIT, CGQ & \href{https://github.com/XDUxyLi/SCEN-master}{\textcolor{blue}{[Link]}} \\
        & & CVGAE~\cite{anwaar2022leveraging} & MM & 2022  & ResNet18  & Graph & UTZ, MIT, CGQ & \href{https://github.com/blindsubmission1/CVGAE}{\textcolor{blue}{[Link]}} \\
        & & CAPE~\cite{khan2023learning} & WACV & 2023  & ResNet18  & Attention & UTZ, MIT, CGQ & - \\
        & & Yang et al.~\cite{yang2023dual} & TMM & 2023  & ResNet18  & contrast & UTZ, MIT & \\
        & & GIPCOL~\cite{xu2024gipcol} & WACV & 2024  & CLIP  & Graph & UTZ, MIT, CGQ & \href{https://github.com/HLR/GIPCOL}{\textcolor{blue}{[Link]}} \\
        %======================= others ===============
        \cmidrule{2-9}
        & \multirow{4}{*}{\rotatebox[origin=c]{90}{Others}} & TMN~\cite{purushwalkam2019task} & ICCV & 2019  & ResNet18  & Gated Network & UTZ, MIT & \href{http://www.cs.cmu.edu/~spurushw/projects/compositional.html}{\textcolor{blue}{[Link]}} \\
        & & MUST~\cite{jiang2024mutual} & PR & 2024  & ResNet18  & Weighted Loss & UTZ, MIT, CGQ & \href{https://github.com/LanchJL/MUST}{\textcolor{blue}{[Link]}} \\
        & & ProLT~\cite{jiang2024revealing} & AAAI & 2024  & ResNet18, CLIP  & Weighted Loss & UTZ, MIT, CGQ & - \\
        & & CDS-CZSL~\cite{li2024context} & CVPR & 2024  & CLIP & Weighted Loss & UTZ, MIT, CGQ & - \\
        \bottomrule

         \multicolumn{9}{p{17cm}}{\footnotesize{\textit{[UTZ] means UT-Zsppos, [MIT] means MIT-States, [CGQ] means C-GQA, [AOC] means AO-CLEVr, [VAW] means VAW-CZSL, [CLO] means Clothing16K. Click the \textcolor{blue}{[link]} to check more information in the corresponding repository.}}} \\
    \end{tabular}
    }
    \label{tab:comp reco implement details}
\vspace{-4mm}
\end{table*}

\subsection{Standardized Datasets}
In Table~\ref{tab:datasets}, we list the commonly used standardized datasets for three tasks ZSOR, ZSCR, and FBOR, respectively. 
Specifically, there are 8 commonly used datasets used for ZSOR, including fine-grained datasets such as CUB~\cite{wah2011caltech}, which is a bird dataset; SUN~\cite {patterson2012sun}, which is a scene dataset, and coarse-grained datasets such as AWA~\cite{lampert2013attribute}, which is a dataset of various types of animals. Most of the datasets contain detailed attribute annotations. Note that Flowers102~\cite{nilsback2008automated} and NABirds~\cite{van2015building} do not have attribute annotations. Flowers102 contains textual descriptions, \textit{i.e.}, each image is described by several sentences. NABirds has region annotations, which mark the parts of the bird and are typically used in visual component analysis methods.
For ZSCR, there are 6 commonly used datasets. Different from ZSOR, the column \textit{[Class/Attr.]} indicates the number of objects and the number of states they contain. It can be seen that their compositional sizes cover different orders of magnitude, ranging from $3/8$ (AO-CLEVr~\cite{atzmon2020causal}), which means up to 24 compositions, to $674/413$ (C-GQA~\cite{naeem2021learning}), which means 278,362 compositions.
Last, we show 15 commonly used datasets for FBOR that cover different categories and different data sizes. ImageNet~\cite{deng2009imagenet} includes data with up to 1000 categories, while EuroSAT~\cite{helber2019eurosat} and CIFAR-10~\cite{krizhevsky2009learning} have only 10 categories. The data types are also diverse, including food~\cite{bossard2014food}, flowers~\cite{nilsback2008automated}, pets~\cite{parkhi2012cats}, transportation~\cite{krause2013collecting, maji2013fine}, etc. They are also fine-grained datasets. In addition, DTD~\cite{cimpoi2014describing} is a texture dataset, and EuroSAT~\cite{helber2019eurosat} is a remote sensing dataset.

\subsection{Implementation}

To facilitate the deep understanding and fast reusing of ZSIR approaches, this survey further provides the implementation details of most representative methods in Tables~\ref{tab:oject recognition of implement details}, \ref{tab:comp reco implement details}, and \ref{tab:implement details fbor}. 
Specifically, Table~\ref{tab:oject recognition of implement details} and \ref{tab:comp reco implement details} show the implementation details of the related studies of ZSOR and ZSCR, including the backbone, auxiliary, dataset, and code addresses. It can be seen that commonly used backbones include ResNet~\cite{he2016deep}, VGG~\cite{simonyan2014very}, GoogleNet~\cite{szegedy2015going}, AlexNet~\cite{krizhevsky2012imagenet}, ViT~\cite{dosovitskiy2020image}, and CLIP~\cite{radford2021learning}. \textit{[Auxiliary]} refers to additional information such as region-level annotations. We provide the dataset and code address to facilitate access to the performance of the method as well as its implementation. Finally, we label the study with up to two keywords based on the specific technique used.

In Table~\ref{tab:implement details fbor}, we show more details of the related work in foundation model-based open-world recognition. We list their visual and text encoders, as well as auxiliary information. GPT-3~\cite{brown2020language} and Diffusion~\cite{rombach2022high} are the exotic models. \textit{[Statistic]} denotes the mean and variance of the pre-training data. \textit{[Distribution]} denotes the prior distribution of labels. We also provide their code repositories for quick reference. 
%Since FBOR is at its early stage and has less related work, it is not easy to categorize. To better understand the focus of each work, we summarize their contributions in a single sentence.

\begin{table*}[htbp]
    \newcolumntype{M}[1]{>{\centering\arraybackslash}m{#1}}
    \renewcommand{\arraystretch}{0.8}
    \centering
    \caption{More details on selected advances for fundation model-based open-world recognition.}
    \centering
    \resizebox{\textwidth}{!}{
    %\resizebox{\textwidth}{33mm}{ 
    \begin{tabular}{c| l| c| c| c| c | c| l| c }
        \toprule
        \textbf{Topic} & \textbf{Name} & \textbf{Venue} & \textbf{Year} & \textbf{VEncoder} & \textbf{TEncoder} & \textbf{Auxiliary} & \textbf{Contribution} & \textbf{Code}\\
        \midrule
          \multirow{6}{*}{\rotatebox[origin=c]{90}{ZSP}} & Menon et al.~\cite{menon2022visual} & arXiv & 2022 & ViT-B$^\dagger$, ViT-L$^\dagger$ & Transformer & GPT-3 & Generate description words according to category names. & - \\
          & APPLe~\cite{chendon} & arXiv & 2023  & ViT-L/14 & Transformer & GPT-3 & Propose prompt prototype learning by initializing from description sentences.  & - \\
          & CuPL~\cite{pratt2023does} & ICCV & 2023  & ViT-L/14 & Transformer & GPT-3 & Generate description sentences according to category names. & \href{https://github.com/sarahpratt/CuPL}{\textcolor{blue}{[Link]}} \\
          & SuS-X~\cite{udandarao2023sus} & ICCV & 2023  & ResNet50 & Transformer & Diffusion & Generate visual support set with Diffusion. & \href{https://github.com/vishaal27/SuS-X}{\textcolor{blue}{[Link]}} \\
          & CHiLS~\cite{novack2023chils} & ICML & 2023  & ViTL/14@336px & Transformer  & GPT-3 & Generate sub-class names according to category names. & \href{https://github.com/acmi-lab/CHILS}{\textcolor{blue}{[Link]}}\\
          & CALIP~\cite{guo2023calip} & AAAI & 2023  & ResNet50 & Transformer & - & Propose attention interaction of image and text embeddings. & \href{https://github.com/ZiyuGuo99/CALIP}{\textcolor{blue}{[Link]}}\\
        \cmidrule{1-9}
        % zsor ==================== tta ===============
        \multirow{9}{*}{\rotatebox[origin=c]{90}{TTA}} & TPT~\cite{shu2022test}  & NeurIPS & 2022  & ResNet50, ViT-B/16 & Transformer & - & Propose test-time prompt adaptation. & \href{https://azshue.github.io/TPT/}{\textcolor{blue}{[Link]}}\\
        & DiffTPT~\cite{feng2023diverse} & ICCV & 2023  & ResNet50, ViT-B/16 & Transformer & Diffusion & Propose to augment image with Diffusion.& \href{https://github.com/chunmeifeng/DiffTPT}{\textcolor{blue}{[Link]}} \\
        & PromptAlign~\cite{abdul2024align} & NeurIPS & 2023  & ViT-B/16, ViT-B/32 & Transformer  & Statistic & Align the distributions of test and pre-training set. & \href{https://jameelhassan.github.io/promptalign/}{\textcolor{blue}{[Link]}} \\
        & SwapPrompt~\cite{ma2024swapprompt} & NeurIPS & 2023  & ResNet50 & Transformer  &- & Update the prompt using exponential moving average. & - \\
        & Zhang et al.~\cite{zhang2024robust} & AAAI & 2024  & ViT-B/16 & Transformer &- & Introduce multiple manual prompts. & - \\
        & DART~\cite{liu2024dart} & AAAI & 2024  & ViT-B/16 & Transformer  &- & Learn a separate prompt for each class. & - \\
        & TDA~\cite{karmanov2024efficient} & CVPR & 2024  & ResNet50, ViT-B/16 & Transformer &- & Two cache banks are set up to store hard and soft labels. & \href{https://kdiaaa.github.io/tda/}{\textcolor{blue}{[Link]}} \\
        & MTA~\cite{zanella2024test} & CVPR & 2024  & ViT-B/16 & Transformer &- & Improve visual representation in the embedding space. & \href{https://github.com/MaxZanella/MTA}{\textcolor{blue}{[Link]}} \\
        & DMN-ZS~\cite{zhang2024dual} & CVPR & 2024  & ResNet50, ViT-B/16 & Transformer & -& Cache high-confidence samples for subsequent predictions. & \href{https://github.com/YBZh/DMN}{\textcolor{blue}{[Link]}} \\
        \cmidrule{1-9}
        %zsor ==========================ua=======================================
        \multirow{6}{*}{\rotatebox[origin=c]{90}{UA}} & CLIPPR~\cite{kahana2022improving} & arXiv &  2022 & ViT-B/32 & Transformer & Distribution & Predicted label aligns a priori label distribution. &  \href{https://github.com/jonkahana/CLIPPR}{\textcolor{blue}{[Link]}} \\
         & UPL~\cite{huang2022unsupervised} & arXiv &  2022 & ResNet50 & Transformer &- & Select high-confidence samples for fine-tuning. & \href{https://github.com/tonyhuang2022/UPL}{\textcolor{blue}{[Link]}} \\
         & MUST~\cite{li2022masked} & ICLR & 2023  & ViT-B/16, ViT-L/14 & Transformer &- & Introduce visual reconstruction and teacher-student model. & \href{https://github.com/salesforce/MUST}{\textcolor{blue}{[Link]}} \\
         & InMaP~\cite{qian2024intra} & NeurIPS & 2023  & ResNet50, ViT-B/16 & Transformer & - & Learn visual prototypes in the embedding space. & - \\
         & LaFTer~\cite{mirza2024lafter} & NeurIPS & 2023  & ViT-B/32 & Transformer & GPT-3 & Train a text classifier to categorize images. & - \\
         & ZLaP~\cite{kalantidis2024label} & CVPR & 2024  & ResNet50, ViT-B/16 & Transformer &- & Introduce a label propagation algorithm. & \href{https://github.com/vladan-stojnic/ZLaP}{\textcolor{blue}{[Link]}} \\
        \bottomrule

         \multicolumn{9}{p{23cm}}{\footnotesize{\textit{[ZSP] means zero-shot prediction. [TTA] means test-time adaptation. [UA] means unsupervised adaptation. [VEncoder] means visual encoder. [TEncoder] means text encoder. [ViT-B$^\dagger$]: ViT-B/32, ViT-B/16. [ViT-L$^\dagger$]: ViT-L/14, ViT-L/14@336px. [Statistic] means the mean and variance of pretrained dataset. [Distribution] means the label distribution. Click the \textcolor{blue}{[link]} to check more information in the corresponding repository.}}} \\
    \end{tabular}}
    \label{tab:implement details fbor}
\vspace{-4mm}
\end{table*}

\subsection{Shared Technologies}
\label{shared technologies}
It can be seen from Tables~\ref{tab:oject recognition of implement details}, \ref{tab:comp reco implement details}, and \ref{tab:implement details fbor} that some techniques are common across the three tasks. We further discuss these shared techniques to reveal more commonalities between these tasks. 
In particular, attention is the most commonly used technique, distributed across almost every sub-direction in the three tasks. This suggests that attention techniques are extremely generalizable and can be applied to different task goals and with different research motivations. In ZSOR, attention is used to localize key visual regions~\cite{xie2019attentive, xie2020region}, model visual-text correlations~\cite{huynh2020fine, chen2022msdn, chen2022transzero}, and localize key attributes~\cite{liu2019attribute}. In ZSCR, attention is used to model object and state correlations~\cite{saini2022disentangling, zhang2022learning, kim2023hierarchical}, and enhance representations~\cite{xu2021relation, khan2023learning}. In FBOR, attention is used to capture correlations between images and text~\cite{guo2023calip}. Prototype learning is another more generalized technique. In ZSOR, prototypes are used to represent attributes~\cite{xu2020attribute, wang2021dual}. In ZSCR, prototypes are used to represent objects and states~\cite{ruis2021independent, lu2023decomposed}. In FBOR, prototypes are used to replace the embedding of prompts in deep space~\cite{qian2024intra}. Graph and contrast learning are the more commonly used techniques in ZSOR and ZSCR. The graph is used to model the topological relationships between visual regions in ZSOR~\cite{hu2022region, guo2023graph}, while in ZSCR, it is used to capture the logical relationships between individual objects and states~\cite{naeem2021learning, anwaar2022leveraging}. Contrast learning is used in ZSOR to optimize attribute embeddings~\cite{chen2023duet}, while in ZSCR, it is used to untangle and model semantic relations~\cite{wei2019adversarial, li2022siamese}. In FBOR, thanks to the robust zero-shot performance of VLMs, fine-tuning prompts become a popular approach~\cite{shu2022test, ma2024swapprompt}. Similarly, in ZSCR, some methods with CLIP as the backbone introduce prompt learning~\cite{lu2023decomposed, xu2024gipcol}, which effectively improves the performance of the model.

\section{Application}
\label{section:application}
%Zero-shot learning is of great practical significance due to its radical freeing up of the pressure of image data collection. Meanwhile, element-wise zero-shot learning provides an intuitive humanoid study pattern that is suitable for a variety of reasoning tasks. 
In recent years, besides image recognition, element-wise zero-shot techniques have been applied in a wide range of areas and developed synergistically. We ketch out some following related applications.

\subsection{Language and Text}
%\textit{Language and Text.} 
Language, as one of the main media used by humans, has always received attention from researchers. Due to label pressures, zero-shot generalization is also a long-standing direction in language studies~\cite{hou2024large}. In recent years, large language models have been developed rapidly~\cite{kojima2022large}, and they can respond to human conversations fluently, showing their powerful zero-shot reasoning ability. Meanwhile, some studies~\cite{wei2021finetuned} proposed fine-tuning schemes for language models to improve the zero-shot performance. In some specific tasks, zero-shot techniques have been widely applied to solve the problem of insufficient data samples, such as machine translation~\cite{johnson2017google}, question answering~\cite{baek2023knowledge}, etc. 
In addition, zero-shot techniques also play an important role in text analysis, such as sentiment classification~\cite{kumar2023zero}, topic classification~\cite{chen2021zero}, etc., which is extremely helpful for research in sociology, psychology, and other disciplines. 
%The content of texts is quite complex, and the text elaborating on the same topic may have multiple forms of expression. 
%
Meanwhile, the collection and labeling of text samples is a huge obstacle, and zero-shot generalization offers a promising solution~\cite{bari2020zero}. 
Moreover, recent studies, such as named entity recognition~\cite{bari2020zero} and information retrieval~\cite{thakur2021beir}, also extensively rely on zero-shot ability.

\subsection{Scene Understanding}
%\textit{Scene Understanding.} 
Scene understanding is a topic based on image recognition but is more challenging. Instead of assigning a label to an entire image, it usually identifies and localizes multiple elements in an image. For example, multi-label recognition task~\cite{huynh2020shared} requires simultaneous recognition of multiple objects in an image. In contrast, object detection~\cite{ren2015faster, liu2016ssd} and semantic segmentation~\cite{long2015fully, kirillov2023segment} are two more complex and practical tasks. Object detection requires identifying and localizing multiple objects in an image, while semantic segmentation is a pixel-level classification task that requires assigning labels to each pixel according to semantic information. Scene understanding is an important research topic, which is not only the basis for AI to understand the world but also a key part of deployment to industrial scenarios. Related technologies are needed in a variety of scenarios, such as autonomous driving, medical diagnostic assistance, security monitoring, retail, and so on. However, they also require more detailed annotation and larger data for training than image recognition. Therefore, incorporating zero-shot learning into scene understanding has become an attractive direction in recent years. For example, zero-shot object detection~\cite{bansal2018zero, rahman2018zero, li2019zero} and zero-shot semantic segmentation~\cite{gu2020context, huynh2022open, he2023primitive} are dedicated to the deployment of models in the open world.

\subsection{3D Vision}
%\textit{3D Vision.} 
With the increasing maturity of 3D sensors, the cost of acquiring 3D vision data has dropped significantly~\cite{guo2020deep}. Compared with traditional 2D images, 3D visual data contains important depth information and, therefore, can provide richer semantic information. 3D vision data has a wide range of applications, covering automatic driving, medical diagnosis, remote sensing, etc. Correspondingly, research on 3D vision data has also received more and more attention. However, compared with 2D data, the acquisition and labeling cost of 3D data is more expensive. Therefore, many researches integrate the concept of zero-shot into 3D tasks, such as point cloud classification~\cite{naeem20223d, cheraghian2022zero}, object detection~\cite{zhang2023sam3d}, and semantic segmentation~\cite{abdelreheem2023satr}. In addition to semantic recognition tasks, some studies are also exploring the application of zero-shot learning to tasks such as generation~\cite{liu2023zero}, style transfer~\cite{liu2023stylerf}, and synthesis~\cite{xu2023dream3d, li2024genzi}. There are also some studies that have explored in-depth on special tasks, such as human pose estimation~\cite{jiang2024back}, shape correspondence~\cite{abdelreheem2023zero}, and so on.

\subsection{Ohters}
%\textit{Other Tasks.} 
Zero-shot learning offers promising solutions for many specific application scenarios. For example, medical and remote sensing images are much more difficult to collect than natural images due to sample scarcity and data protection. Therefore, the research on zero-shot medical image~\cite{javed2024cplip} recognition and remote sensing image~\cite{li2023rs} recognition has been developed rapidly in recent years. Anomaly detection~\cite{li2024zero} plays a key role in securing many industrial fields. However, the number and type of anomaly samples are often insufficient. Some studies resort to zero-shot learning to solve this problem. In addition, zero-shot techniques have been applied in many special tasks to alleviate the low-data problem as well as to enhance the generalization ability of the model, such as action recognition~\cite{chen2021elaborative}, image retrieval~\cite{wang2024content}, and video classification~\cite{hong2023fine}.

\section{Future Directions}
\label{section:future directions}

Zero-shot learning contributes powerful technical solutions for automated recognition in the open environment. Meanwhile, research based on element-wise representation and reasoning has driven the rapid development and wide application of zero-shot technology. Despite the remarkable achievements, there are still some potential limitations that have not been deeply explored. In this section, we share our perspectives on future research directions.

\subsection{Continual Zero-Shot Learning}
Existing studies are based on closed systems to design solutions. However, in reality, there are a large number of growing systems with dynamically changing environments and goals. For example, for zero-shot object recognition, attributes and categories may be progressively inflated. Early in the observation phase, fewer attributes are accessible, followed by a steady stream of new attribute annotations added. Similarly, the detected categories may increase over time forward. It is difficult for existing schemes to handle such dynamic recognition tasks, and continual learning has come a long way in this regard~\cite{wang2024comprehensive}. However, simply transplanting techniques from continual learning to zero-shot recognition tasks suffers from deviation. Therefore, fusing continual learning and zero-shot learning will further increase the value of applications to real-world scenarios.

\subsection{Multi-label Compositional Recognition}
Zero-shot compositional recognition allows the model to not only learn to recognize objects but also to describe their state. However, current research is limited to a single object and single state. A robust compositional recognition system should have the ability to recognize multiple objects as well as multiple states. For a single object, the recognition system should describe its state from multiple perspectives to provide more detailed information~\cite{misra2017red}. For multiple objects, their individual states, as well as their interactions, relationships, etc., should be described~\cite{zhou2021multi}. Expanding from single labels to multiple labels is a tough choice, but it is an inescapable step toward building a smarter recognition system.

\subsection{VLMs with Downstream Fine-Grained Annotations}
The power of VLMs in the open world has been demonstrated by the fact that it requires only the names of the categories in the downstream task to accomplish zero-shot inference. Meanwhile, fine-grained annotations can improve the zero-shot inference performance of VLMs~\cite{menon2022visual, pratt2023does}. What happens if VLMs are combined with downstream exclusive attribute annotations, such as attribute texts or state descriptions? There are a number of issues that deserve deeper investigation, including domain shift between pre-trained text embeddings of VLMs and downstream text, fine-grained visual-text matching, catastrophic forgetting, and so on.

\subsection{Annotation Cost and Quality}
Element-wise schemes in zero-shot object recognition typically require extensive refined attribute annotations, especially for visual-attribute matching schemes. However, the attribute-level annotations are time- and labor-intensive compared to class-level labeling. Worse still, once they settle into concrete real-world scenarios, such as industrial inspection or medical pathology, the expert knowledge can be a bottleneck, which further raises the labor cost. In addition, attribute engineering is a complex crossover field. Even attributes annotated by experienced experts do not guarantee benefits for deep learning, which implies that high-quality attribute annotations require professionals with dual knowledge of both specific domains and deep learning. Despite some studies attempting to make breakthroughs in the field of automated annotation~\cite{akata2016multi}, it is clear that there is still a long way to go.

\subsection{Deployment Cost}
Compared to class-wise semantic modeling, many element-wise schemes have to process a higher density of information, which introduces a more luxurious deployment cost. Such cost is reflected in bloated network structures and high computational complexity \textit{in the deployment phase}. As a result, they are unfriendly to edge tasks and mini-endpoints, which have to trade off performance and memory~\cite{guo2023parsnets}. However, they can be more favorable to a scenario associated with resource-constrained devices due to the low or even zero data requirements. Such a scenario can also well align with ubiquitous devices and data in real-world applications. Therefore, it is promising to investigate on-device-friendly element-wise algorithms.

\subsection{Unified Framework}
Zero-shot studies share many common techniques, such as attention, graph modeling, and so on. However, barriers between them still exist due to the different goals and contexts of the tasks. A technique and framework that is versatile across all fields is lacking. However, the significance of such a unified framework for both academia and application is immense~\cite{rahman2018unified}. The quest for a unified framework means exploring more commonalities across zero-shot tasks and thinking about the operational mechanisms of the entire realm from the ground-up principles. From a realistic point of view, a mature intelligence should be able to handle multiple tasks at the same time.

\subsection{Robust Representation Learning}
Despite some solutions have been proposed~\cite{chen2023zero} to tackle with domain shift, there is a lack of quantitative assessments regarding their resistance to interference. Consequently, the robustness of existing representation methods in real-world environments remains uncertain. Recently, Shafiee et al.~\cite{shafiee2022zero} introduce adversarial attack technique into zero-shot models. The interplay between model attacks and defenses has been a long-standing topic. Attackers aim to weaken models at minimal cost, while defenders strive to develop more robust representations to resist these perturbations. This dynamic process of learning from each other leads to continuous progress, offering a new perspective on combating domain shift.

%While Chen et al.~\cite{chen2023zero} proposed an approach that can improve the zero-shot models with adversarial samples.

\subsection{Poor Theoretical Foundation}
The development of zero-shot learning is established on the beautiful hypothesis that deep neural networks can reason logically like humans~\cite{lampert2009learning, xian2018zero}. As mentioned, we can learn to identify basic elements and recognize new concepts by combining and reasoning about them. Nevertheless, there are not many solid theories on the compatibility between human reasoning and machine inductive ability, leading to a lack of explainability. Meanwhile, some flaws also challenge the plausibility of the hypothesis, such as the correspondence between abstract attributes and vision. Rigorous theoretical guidance is at the helm of a field moving forward, and it is of great prospective to dive into the mysterious black box in the future.

\section{Conclusion}
We provide a comprehensive exposition of three major tasks in the field of zero-shot image recognition that share the notion of element-wise representation and reasoning. Specifically, we show the background, commonalities, and challenges faced by several tasks. Then related recent techniques are meticulously categorized and explained. At the same time, we provide specific technical details, including datasets, training details, code repository addresses, and more. Finally, we show the wide range of applications of element-wise zero-shot learning and look at future directions.

\bibliographystyle{IEEEtran}
\bibliography{reference}

% Generated by IEEEtran.bst, version: 1.14 (2015/08/26)
\begin{thebibliography}{100}
\providecommand{\url}[1]{#1}
\csname url@samestyle\endcsname
\providecommand{\newblock}{\relax}
\providecommand{\bibinfo}[2]{#2}
\providecommand{\BIBentrySTDinterwordspacing}{\spaceskip=0pt\relax}
\providecommand{\BIBentryALTinterwordstretchfactor}{4}
\providecommand{\BIBentryALTinterwordspacing}{\spaceskip=\fontdimen2\font plus
\BIBentryALTinterwordstretchfactor\fontdimen3\font minus \fontdimen4\font\relax}
\providecommand{\BIBforeignlanguage}[2]{{%
\expandafter\ifx\csname l@#1\endcsname\relax
\typeout{** WARNING: IEEEtran.bst: No hyphenation pattern has been}%
\typeout{** loaded for the language `#1'. Using the pattern for}%
\typeout{** the default language instead.}%
\else
\language=\csname l@#1\endcsname
\fi
#2}}
\providecommand{\BIBdecl}{\relax}
\BIBdecl

\bibitem{deng2009imagenet}
J.~Deng, W.~Dong, R.~Socher, L.-J. Li, K.~Li, and L.~Fei-Fei, ``Imagenet: A large-scale hierarchical image database,'' in \emph{CVPR}.\hskip 1em plus 0.5em minus 0.4em\relax Ieee, 2009, pp. 248--255.

\bibitem{chen2014inferring}
C.-Y. Chen and K.~Grauman, ``Inferring analogous attributes,'' in \emph{CVPR}, 2014, pp. 200--207.

\bibitem{misra2017red}
I.~Misra, A.~Gupta, and M.~Hebert, ``From red wine to red tomato: Composition with context,'' in \emph{CVPR}, 2017, pp. 1792--1801.

\bibitem{yang2018learning}
Z.~Yang, T.~Luo, D.~Wang, Z.~Hu, J.~Gao, and L.~Wang, ``Learning to navigate for fine-grained classification,'' in \emph{ECCV}, 2018, pp. 420--435.

\bibitem{akata2015evaluation}
Z.~Akata, S.~Reed, D.~Walter, H.~Lee, and B.~Schiele, ``Evaluation of output embeddings for fine-grained image classification,'' in \emph{CVPR}, 2015, pp. 2927--2936.

\bibitem{nagarajan2018attributes}
T.~Nagarajan and K.~Grauman, ``Attributes as operators: factorizing unseen attribute-object compositions,'' in \emph{ECCV}, 2018, pp. 169--185.

\bibitem{wah2011caltech}
C.~Wah, S.~Branson, P.~Welinder, P.~Perona, and S.~Belongie, ``The caltech-ucsd birds-200-2011 dataset,'' 2011.

\bibitem{patterson2012sun}
G.~Patterson and J.~Hays, ``Sun attribute database: Discovering, annotating, and recognizing scene attributes,'' in \emph{CVPR}.\hskip 1em plus 0.5em minus 0.4em\relax IEEE, 2012, pp. 2751--2758.

\bibitem{wei2021fine}
X.-S. Wei, Y.-Z. Song, O.~Mac~Aodha, J.~Wu, Y.~Peng, J.~Tang, J.~Yang, and S.~Belongie, ``Fine-grained image analysis with deep learning: A survey,'' \emph{IEEE TPAMI}, vol.~44, no.~12, pp. 8927--8948, 2021.

\bibitem{wang2016walk}
J.~Wang, Y.~Cheng, and R.~S. Feris, ``Walk and learn: Facial attribute representation learning from egocentric video and contextual data,'' in \emph{CVPR}, 2016, pp. 2295--2304.

\bibitem{kovashka2012whittlesearch}
A.~Kovashka, D.~Parikh, and K.~Grauman, ``Whittlesearch: Image search with relative attribute feedback,'' in \emph{CVPR}.\hskip 1em plus 0.5em minus 0.4em\relax IEEE, 2012, pp. 2973--2980.

\bibitem{wu2017image}
Q.~Wu, C.~Shen, P.~Wang, A.~Dick, and A.~Van Den~Hengel, ``Image captioning and visual question answering based on attributes and external knowledge,'' \emph{IEEE TPAMI}, vol.~40, no.~6, pp. 1367--1381, 2017.

\bibitem{santa2018neural}
R.~Santa~Cruz, B.~Fernando, A.~Cherian, and S.~Gould, ``Neural algebra of classifiers,'' in \emph{WACV}.\hskip 1em plus 0.5em minus 0.4em\relax IEEE, 2018, pp. 729--737.

\bibitem{biederman1987recognition}
I.~Biederman, ``Recognition-by-components: a theory of human image understanding.'' \emph{Psychological review}, vol.~94, no.~2, p. 115, 1987.

\bibitem{hoffman1987parts}
D.~D. Hoffman and W.~A. Richards, ``Parts of recognition,'' in \emph{Readings in Computer Vision}.\hskip 1em plus 0.5em minus 0.4em\relax Elsevier, 1987, pp. 227--242.

\bibitem{felzenszwalb2008discriminatively}
P.~Felzenszwalb, D.~McAllester, and D.~Ramanan, ``A discriminatively trained, multiscale, deformable part model,'' in \emph{CVPR}.\hskip 1em plus 0.5em minus 0.4em\relax Ieee, 2008, pp. 1--8.

\bibitem{battaglia2018relational}
P.~W. Battaglia, J.~B. Hamrick, V.~Bapst, A.~Sanchez-Gonzalez, V.~Zambaldi, M.~Malinowski, A.~Tacchetti, D.~Raposo, A.~Santoro, R.~Faulkner \emph{et~al.}, ``Relational inductive biases, deep learning, and graph networks,'' \emph{arXiv}, 2018.

\bibitem{lampert2009learning}
C.~H. Lampert, H.~Nickisch, and S.~Harmeling, ``Learning to detect unseen object classes by between-class attribute transfer,'' in \emph{CVPR}.\hskip 1em plus 0.5em minus 0.4em\relax IEEE, 2009, pp. 951--958.

\bibitem{wang2020generalizing}
Y.~Wang, Q.~Yao, J.~T. Kwok, and L.~M. Ni, ``Generalizing from a few examples: A survey on few-shot learning,'' \emph{ACM computing surveys (csur)}, vol.~53, no.~3, pp. 1--34, 2020.

\bibitem{o2019one}
N.~O’Mahony, S.~Campbell, A.~Carvalho, L.~Krpalkova, G.~V. Hernandez, S.~Harapanahalli, D.~Riordan, and J.~Walsh, ``One-shot learning for custom identification tasks; a review,'' \emph{Procedia Manufacturing}, vol.~38, pp. 186--193, 2019.

\bibitem{yang2024generalized}
J.~Yang, K.~Zhou, Y.~Li, and Z.~Liu, ``Generalized out-of-distribution detection: A survey,'' \emph{IJCV}, pp. 1--28, 2024.

\bibitem{xian2018zero}
Y.~Xian, C.~H. Lampert, B.~Schiele, and Z.~Akata, ``Zero-shot learning—a comprehensive evaluation of the good, the bad and the ugly,'' \emph{IEEE TPAMI}, vol.~41, no.~9, pp. 2251--2265, 2018.

\bibitem{wang2019survey}
W.~Wang, V.~W. Zheng, H.~Yu, and C.~Miao, ``A survey of zero-shot learning: Settings, methods, and applications,'' \emph{ACM TIST}, vol.~10, no.~2, pp. 1--37, 2019.

\bibitem{pourpanah2022review}
F.~Pourpanah, M.~Abdar, Y.~Luo, X.~Zhou, R.~Wang, C.~P. Lim, X.-Z. Wang, and Q.~J. Wu, ``A review of generalized zero-shot learning methods,'' \emph{IEEE TPAMI}, vol.~45, no.~4, pp. 4051--4070, 2022.

\bibitem{lake2017building}
B.~M. Lake, T.~D. Ullman, J.~B. Tenenbaum, and S.~J. Gershman, ``Building machines that learn and think like people,'' \emph{Behavioral and brain sciences}, vol.~40, p. e253, 2017.

\bibitem{purushwalkam2019task}
S.~Purushwalkam, M.~Nickel, A.~Gupta, and M.~Ranzato, ``Task-driven modular networks for zero-shot compositional learning,'' in \emph{ICCV}, 2019, pp. 3593--3602.

\bibitem{mancini2021open}
M.~Mancini, M.~F. Naeem, Y.~Xian, and Z.~Akata, ``Open world compositional zero-shot learning,'' in \emph{CVPR}, 2021, pp. 5222--5230.

\bibitem{radford2021learning}
A.~Radford, J.~W. Kim, C.~Hallacy, A.~Ramesh, G.~Goh, S.~Agarwal, G.~Sastry, A.~Askell, P.~Mishkin, J.~Clark \emph{et~al.}, ``Learning transferable visual models from natural language supervision,'' in \emph{ICML}.\hskip 1em plus 0.5em minus 0.4em\relax PMLR, 2021, pp. 8748--8763.

\bibitem{jia2021scaling}
C.~Jia, Y.~Yang, Y.~Xia, Y.-T. Chen, Z.~Parekh, H.~Pham, Q.~Le, Y.-H. Sung, Z.~Li, and T.~Duerig, ``Scaling up visual and vision-language representation learning with noisy text supervision,'' in \emph{ICML}.\hskip 1em plus 0.5em minus 0.4em\relax PMLR, 2021, pp. 4904--4916.

\bibitem{zhou2022learning}
K.~Zhou, J.~Yang, C.~C. Loy, and Z.~Liu, ``Learning to prompt for vision-language models,'' \emph{IJCV}, vol. 130, no.~9, pp. 2337--2348, 2022.

\bibitem{zhou2022conditional}
------, ``Conditional prompt learning for vision-language models,'' in \emph{CVPR}, 2022, pp. 16\,816--16\,825.

\bibitem{menon2022visual}
S.~Menon and C.~Vondrick, ``Visual classification via description from large language models,'' \emph{arXiv}, 2022.

\bibitem{guo2023calip}
Z.~Guo, R.~Zhang, L.~Qiu, X.~Ma, X.~Miao, X.~He, and B.~Cui, ``Calip: Zero-shot enhancement of clip with parameter-free attention,'' in \emph{AAAI}, vol.~37, no.~1, 2023, pp. 746--754.

\bibitem{zhu2019semantic}
Y.~Zhu, J.~Xie, Z.~Tang, X.~Peng, and A.~Elgammal, ``Semantic-guided multi-attention localization for zero-shot learning,'' \emph{NeurIPS}, vol.~32, 2019.

\bibitem{xie2019attentive}
G.-S. Xie, L.~Liu, X.~Jin, F.~Zhu, Z.~Zhang, J.~Qin, Y.~Yao, and L.~Shao, ``Attentive region embedding network for zero-shot learning,'' in \emph{CVPR}, 2019, pp. 9384--9393.

\bibitem{xie2020region}
G.-S. Xie, L.~Liu, F.~Zhu, F.~Zhao, Z.~Zhang, Y.~Yao, J.~Qin, and L.~Shao, ``Region graph embedding network for zero-shot learning,'' in \emph{ECCV}.\hskip 1em plus 0.5em minus 0.4em\relax Springer, 2020, pp. 562--580.

\bibitem{guo2023graph}
J.~Guo, S.~Guo, Q.~Zhou, Z.~Liu, X.~Lu, and F.~Huo, ``Graph knows unknowns: Reformulate zero-shot learning as sample-level graph recognition,'' in \emph{AAAI}, vol.~37, no.~6, 2023, pp. 7775--7783.

\bibitem{huynh2020fine}
D.~Huynh and E.~Elhamifar, ``Fine-grained generalized zero-shot learning via dense attribute-based attention,'' in \emph{CVPR}, 2020, pp. 4483--4493.

\bibitem{chen2022transzero}
S.~Chen, Z.~Hong, Y.~Liu, G.-S. Xie, B.~Sun, H.~Li, Q.~Peng, K.~Lu, and X.~You, ``Transzero: Attribute-guided transformer for zero-shot learning,'' in \emph{AAAI}, vol.~36, no.~1, 2022, pp. 330--338.

\bibitem{liu2016deepfashion}
Z.~Liu, P.~Luo, S.~Qiu, X.~Wang, and X.~Tang, ``Deepfashion: Powering robust clothes recognition and retrieval with rich annotations,'' in \emph{CVPR}, 2016, pp. 1096--1104.

\bibitem{atzmon2020causal}
Y.~Atzmon, F.~Kreuk, U.~Shalit, and G.~Chechik, ``A causal view of compositional zero-shot recognition,'' \emph{NeurIPS}, vol.~33, pp. 1462--1473, 2020.

\bibitem{fu2015transductive}
Y.~Fu, T.~M. Hospedales, T.~Xiang, and S.~Gong, ``Transductive multi-view zero-shot learning,'' \emph{IEEE TPAMI}, vol.~37, no.~11, pp. 2332--2345, 2015.

\bibitem{jiang2019transferable}
H.~Jiang, R.~Wang, S.~Shan, and X.~Chen, ``Transferable contrastive network for generalized zero-shot learning,'' in \emph{ICCV}, 2019, pp. 9765--9774.

\bibitem{zhou2022domain}
K.~Zhou, Z.~Liu, Y.~Qiao, T.~Xiang, and C.~C. Loy, ``Domain generalization: A survey,'' \emph{IEEE TPAMI}, vol.~45, no.~4, pp. 4396--4415, 2022.

\bibitem{ganin2015unsupervised}
Y.~Ganin and V.~Lempitsky, ``Unsupervised domain adaptation by backpropagation,'' in \emph{ICML}.\hskip 1em plus 0.5em minus 0.4em\relax PMLR, 2015, pp. 1180--1189.

\bibitem{rao2023srcd}
Z.~Rao, J.~Guo, L.~Tang, Y.~Huang, X.~Ding, and S.~Guo, ``Srcd: Semantic reasoning with compound domains for single-domain generalized object detection,'' \emph{arXiv}, 2023.

\bibitem{ge2022dual}
J.~Ge, H.~Xie, S.~Min, P.~Li, and Y.~Zhang, ``Dual part discovery network for zero-shot learning,'' in \emph{ACM MM}, 2022, pp. 3244--3252.

\bibitem{wei2019adversarial}
K.~Wei, M.~Yang, H.~Wang, C.~Deng, and X.~Liu, ``Adversarial fine-grained composition learning for unseen attribute-object recognition,'' in \emph{ICCV}, 2019, pp. 3741--3749.

\bibitem{romera2015embarrassingly}
B.~Romera-Paredes and P.~Torr, ``An embarrassingly simple approach to zero-shot learning,'' in \emph{ICML}.\hskip 1em plus 0.5em minus 0.4em\relax PMLR, 2015, pp. 2152--2161.

\bibitem{xian2018feature}
Y.~Xian, T.~Lorenz, B.~Schiele, and Z.~Akata, ``Feature generating networks for zero-shot learning,'' in \emph{CVPR}, 2018, pp. 5542--5551.

\bibitem{verma2018generalized}
V.~K. Verma, G.~Arora, A.~Mishra, and P.~Rai, ``Generalized zero-shot learning via synthesized examples,'' in \emph{CVPR}, 2018, pp. 4281--4289.

\bibitem{goodfellow2020generative}
I.~Goodfellow, J.~Pouget-Abadie, M.~Mirza, B.~Xu, D.~Warde-Farley, S.~Ozair, A.~Courville, and Y.~Bengio, ``Generative adversarial networks,'' \emph{Communications of the ACM}, vol.~63, no.~11, pp. 139--144, 2020.

\bibitem{chen2020canzsl}
Z.~Chen, J.~Li, Y.~Luo, Z.~Huang, and Y.~Yang, ``Canzsl: Cycle-consistent adversarial networks for zero-shot learning from natural language,'' in \emph{WCACV}, 2020, pp. 874--883.

\bibitem{chen2020rethinking}
Z.~Chen, S.~Wang, J.~Li, and Z.~Huang, ``Rethinking generative zero-shot learning: An ensemble learning perspective for recognising visual patches,'' in \emph{ACM MM}, 2020, pp. 3413--3421.

\bibitem{kingma2013auto}
D.~P. Kingma and M.~Welling, ``Auto-encoding variational bayes,'' \emph{arXiv}, 2013.

\bibitem{chen2021semantics}
Z.~Chen, Y.~Luo, R.~Qiu, S.~Wang, Z.~Huang, J.~Li, and Z.~Zhang, ``Semantics disentangling for generalized zero-shot learning,'' in \emph{ICCV}, 2021, pp. 8712--8720.

\bibitem{chen2022gsmflow}
Z.~Chen, Y.~Luo, S.~Wang, J.~Li, and Z.~Huang, ``Gsmflow: Generation shifts mitigating flow for generalized zero-shot learning,'' \emph{IEEE TMM}, vol.~25, pp. 5374--5385, 2022.

\bibitem{chen2021mitigating}
Z.~Chen, Y.~Luo, S.~Wang, R.~Qiu, J.~Li, and Z.~Huang, ``Mitigating generation shifts for generalized zero-shot learning,'' in \emph{ACM MM}, 2021, pp. 844--852.

\bibitem{liu2018generalized}
S.~Liu, M.~Long, J.~Wang, and M.~I. Jordan, ``Generalized zero-shot learning with deep calibration network,'' \emph{NeurIPS}, vol.~31, 2018.

\bibitem{wang2021region}
Z.~Wang, Y.~Gou, J.~Li, Y.~Zhang, and Y.~Yang, ``Region semantically aligned network for zero-shot learning,'' in \emph{CIKM}, 2021, pp. 2080--2090.

\bibitem{ji2018stacked}
Z.~Ji, Y.~Fu, J.~Guo, Y.~Pang, Z.~M. Zhang \emph{et~al.}, ``Stacked semantics-guided attention model for fine-grained zero-shot learning,'' \emph{NeurIPS}, vol.~31, 2018.

\bibitem{elhoseiny2017link}
M.~Elhoseiny, Y.~Zhu, H.~Zhang, and A.~Elgammal, ``Link the head to the" beak": Zero shot learning from noisy text description at part precision,'' in \emph{CVPR}, 2017, pp. 5640--5649.

\bibitem{zhu2018generative}
Y.~Zhu, M.~Elhoseiny, B.~Liu, X.~Peng, and A.~Elgammal, ``A generative adversarial approach for zero-shot learning from noisy texts,'' in \emph{CVPR}, 2018, pp. 1004--1013.

\bibitem{li2018discriminative}
Y.~Li, J.~Zhang, J.~Zhang, and K.~Huang, ``Discriminative learning of latent features for zero-shot recognition,'' in \emph{CVPR}, 2018, pp. 7463--7471.

\bibitem{ge2021semantic}
J.~Ge, H.~Xie, S.~Min, and Y.~Zhang, ``Semantic-guided reinforced region embedding for generalized zero-shot learning,'' in \emph{AAAI}, vol.~35, no.~2, 2021, pp. 1406--1414.

\bibitem{li2022entropy}
Y.~Li, Z.~Liu, L.~Yao, X.~Wang, J.~McAuley, and X.~Chang, ``An entropy-guided reinforced partial convolutional network for zero-shot learning,'' \emph{IEEE TCSVT}, vol.~32, no.~8, pp. 5175--5186, 2022.

\bibitem{liu2024part}
M.~Liu, C.~Zhang, H.~Bai, and Y.~Zhao, ``Part-object progressive refinement network for zero-shot learning,'' \emph{IEEE TIP}, 2024.

\bibitem{chen2024progressive}
S.~Chen, W.~Hou, S.~Khan, and F.~S. Khan, ``Progressive semantic-guided vision transformer for zero-shot learning,'' in \emph{CVPR}, 2024, pp. 23\,964--23\,974.

\bibitem{dosovitskiy2020image}
A.~Dosovitskiy, L.~Beyer, A.~Kolesnikov, D.~Weissenborn, X.~Zhai, T.~Unterthiner, M.~Dehghani, M.~Minderer, G.~Heigold, S.~Gelly \emph{et~al.}, ``An image is worth 16x16 words: Transformers for image recognition at scale,'' \emph{arXiv}, 2020.

\bibitem{xu2022vgse}
W.~Xu, Y.~Xian, J.~Wang, B.~Schiele, and Z.~Akata, ``Vgse: Visually-grounded semantic embeddings for zero-shot learning,'' in \emph{CVPR}, 2022, pp. 9316--9325.

\bibitem{kipf2016semi}
T.~N. Kipf and M.~Welling, ``Semi-supervised classification with graph convolutional networks,'' \emph{arXiv}, 2016.

\bibitem{hu2022region}
Z.~Hu, H.~Zhao, J.~Peng, and X.~Gu, ``Region interaction and attribute embedding for zero-shot learning,'' \emph{IS}, vol. 609, pp. 984--995, 2022.

\bibitem{chen2022gndan}
S.~Chen, Z.~Hong, G.~Xie, Q.~Peng, X.~You, W.~Ding, and L.~Shao, ``Gndan: Graph navigated dual attention network for zero-shot learning,'' \emph{IEEE TNNLS}, vol.~35, no.~4, pp. 4516--4529, 2022.

\bibitem{vaswani2017attention}
A.~Vaswani, N.~Shazeer, N.~Parmar, J.~Uszkoreit, L.~Jones, A.~N. Gomez, {\L}.~Kaiser, and I.~Polosukhin, ``Attention is all you need,'' \emph{NeurIPS}, vol.~30, 2017.

\bibitem{liu2021goal}
Y.~Liu, L.~Zhou, X.~Bai, Y.~Huang, L.~Gu, J.~Zhou, and T.~Harada, ``Goal-oriented gaze estimation for zero-shot learning,'' in \emph{CVPR}, 2021, pp. 3794--3803.

\bibitem{huynh2020compositional}
D.~Huynh and E.~Elhamifar, ``Compositional zero-shot learning via fine-grained dense feature composition,'' \emph{NeurIPS}, vol.~33, pp. 19\,849--19\,860, 2020.

\bibitem{liu2022zero}
Y.~Liu, Y.~Dang, X.~Gao, J.~Han, and L.~Shao, ``Zero-shot learning with attentive region embedding and enhanced semantics,'' \emph{IEEE TNNLS}, vol.~35, no.~3, pp. 4220--4231, 2022.

\bibitem{naeem2022i2dformer}
M.~F. Naeem, Y.~Xian, L.~V. Gool, and F.~Tombari, ``I2dformer: Learning image to document attention for zero-shot image classification,'' \emph{NeurIPS}, vol.~35, pp. 12\,283--12\,294, 2022.

\bibitem{chen2023duet}
Z.~Chen, Y.~Huang, J.~Chen, Y.~Geng, W.~Zhang, Y.~Fang, J.~Z. Pan, and H.~Chen, ``Duet: Cross-modal semantic grounding for contrastive zero-shot learning,'' in \emph{AAAI}, vol.~37, no.~1, 2023, pp. 405--413.

\bibitem{liu2023progressive}
M.~Liu, F.~Li, C.~Zhang, Y.~Wei, H.~Bai, and Y.~Zhao, ``Progressive semantic-visual mutual adaption for generalized zero-shot learning,'' in \emph{CVPR}, 2023, pp. 15\,337--15\,346.

\bibitem{chen2022msdn}
S.~Chen, Z.~Hong, G.-S. Xie, W.~Yang, Q.~Peng, K.~Wang, J.~Zhao, and X.~You, ``Msdn: Mutually semantic distillation network for zero-shot learning,'' in \emph{CVPR}, 2022, pp. 7612--7621.

\bibitem{yang2018robust}
H.-M. Yang, X.-Y. Zhang, F.~Yin, and C.-L. Liu, ``Robust classification with convolutional prototype learning,'' in \emph{CVPR}, 2018, pp. 3474--3482.

\bibitem{liu2023improving}
J.~Liu, Y.~Ren, W.~Li, Y.~Zheng, and C.~Wang, ``Improving out-of-distribution detection with margin-based prototype learning,'' in \emph{ICONIP}.\hskip 1em plus 0.5em minus 0.4em\relax Springer, 2023, pp. 149--160.

\bibitem{wang2019panet}
K.~Wang, J.~H. Liew, Y.~Zou, D.~Zhou, and J.~Feng, ``Panet: Few-shot image semantic segmentation with prototype alignment,'' in \emph{ICCV}, 2019, pp. 9197--9206.

\bibitem{xu2020attribute}
W.~Xu, Y.~Xian, J.~Wang, B.~Schiele, and Z.~Akata, ``Attribute prototype network for zero-shot learning,'' \emph{NeurIPS}, vol.~33, pp. 21\,969--21\,980, 2020.

\bibitem{jayaraman2014decorrelating}
D.~Jayaraman, F.~Sha, and K.~Grauman, ``Decorrelating semantic visual attributes by resisting the urge to share,'' in \emph{CVPR}, 2014, pp. 1629--1636.

\bibitem{wang2021dual}
C.~Wang, S.~Min, X.~Chen, X.~Sun, and H.~Li, ``Dual progressive prototype network for generalized zero-shot learning,'' \emph{NeurIPS}, vol.~34, pp. 2936--2948, 2021.

\bibitem{guo2023group}
T.~Guo, J.~Liang, and G.-S. Xie, ``Group-wise interactive region learning for zero-shot recognition,'' \emph{IS}, vol. 642, p. 119135, 2023.

\bibitem{cheng2023discriminative}
D.~Cheng, G.~Wang, N.~Wang, D.~Zhang, Q.~Zhang, and X.~Gao, ``Discriminative and robust attribute alignment for zero-shot learning,'' \emph{IEEE TCSVT}, vol.~33, no.~8, pp. 4244--4256, 2023.

\bibitem{du2023boosting}
Y.~Du, M.~Shi, F.~Wei, and G.~Li, ``Boosting zero-shot learning via contrastive optimization of attribute representations,'' \emph{IEEE TNNLS}, 2023.

\bibitem{chen2023zero}
Z.~Chen, P.~Zhang, J.~Li, S.~Wang, and Z.~Huang, ``Zero-shot learning by harnessing adversarial samples,'' in \emph{ACM MM}, 2023, pp. 4138--4146.

\bibitem{liu2019attribute}
Y.~Liu, J.~Guo, D.~Cai, and X.~He, ``Attribute attention for semantic disambiguation in zero-shot learning,'' in \emph{ICCV}, 2019, pp. 6698--6707.

\bibitem{liu2020attribute}
L.~Liu, T.~Zhou, G.~Long, J.~Jiang, and C.~Zhang, ``Attribute propagation network for graph zero-shot learning,'' in \emph{AAAI}, vol.~34, no.~04, 2020, pp. 4868--4875.

\bibitem{akata2016multi}
Z.~Akata, M.~Malinowski, M.~Fritz, and B.~Schiele, ``Multi-cue zero-shot learning with strong supervision,'' in \emph{CVPR}, 2016, pp. 59--68.

\bibitem{rao2024dual}
Z.~Rao, J.~Guo, X.~Lu, J.~Liang, J.~Zhang, H.~Wang, K.~Wei, and X.~Cao, ``Dual expert distillation network for generalized zero-shot learning,'' in \emph{IJCAI}, 2024.

\bibitem{nan2019recognizing}
Z.~Nan, Y.~Liu, N.~Zheng, and S.-C. Zhu, ``Recognizing unseen attribute-object pair with generative model,'' in \emph{AAAI}, vol.~33, no.~01, 2019, pp. 8811--8818.

\bibitem{ruis2021independent}
F.~Ruis, G.~Burghouts, and D.~Bucur, ``Independent prototype propagation for zero-shot compositionality,'' \emph{NeurIPS}, vol.~34, pp. 10\,641--10\,653, 2021.

\bibitem{hu2023leveraging}
X.~Hu and Z.~Wang, ``Leveraging sub-class discimination for compositional zero-shot learning,'' in \emph{AAAI}, vol.~37, no.~1, 2023, pp. 890--898.

\bibitem{zheng2024caila}
Z.~Zheng, H.~Zhu, and R.~Nevatia, ``Caila: Concept-aware intra-layer adapters for compositional zero-shot learning,'' in \emph{WCACV}, 2024, pp. 1721--1731.

\bibitem{learning2022decomposable}
C.~Z.-S. Learning, ``A decomposable causal view of compositional zero-shot learning,'' 2022.

\bibitem{saini2022disentangling}
N.~Saini, K.~Pham, and A.~Shrivastava, ``Disentangling visual embeddings for attributes and objects,'' in \emph{CVPR}, 2022, pp. 13\,658--13\,667.

\bibitem{hao2023learning}
S.~Hao, K.~Han, and K.-Y.~K. Wong, ``Learning attention as disentangler for compositional zero-shot learning,'' in \emph{CVPR}, 2023, pp. 15\,315--15\,324.

\bibitem{jing2024retrieval}
C.~Jing, Y.~Li, H.~Chen, and C.~Shen, ``Retrieval-augmented primitive representations for compositional zero-shot learning,'' in \emph{AAAI}, vol.~38, no.~3, 2024, pp. 2652--2660.

\bibitem{zhang2022learning}
T.~Zhang, K.~Liang, R.~Du, X.~Sun, Z.~Ma, and J.~Guo, ``Learning invariant visual representations for compositional zero-shot learning,'' in \emph{ECCV}.\hskip 1em plus 0.5em minus 0.4em\relax Springer, 2022, pp. 339--355.

\bibitem{lu2023decomposed}
X.~Lu, S.~Guo, Z.~Liu, and J.~Guo, ``Decomposed soft prompt guided fusion enhancing for compositional zero-shot learning,'' in \emph{CVPR}, 2023, pp. 23\,560--23\,569.

\bibitem{huang2024troika}
S.~Huang, B.~Gong, Y.~Feng, M.~Zhang, Y.~Lv, and D.~Wang, ``Troika: Multi-path cross-modal traction for compositional zero-shot learning,'' in \emph{CVPR}, 2024, pp. 24\,005--24\,014.

\bibitem{kim2023hierarchical}
H.~Kim, J.~Lee, S.~Park, and K.~Sohn, ``Hierarchical visual primitive experts for compositional zero-shot learning,'' in \emph{ICCV}, 2023, pp. 5675--5685.

\bibitem{panda2024compositional}
A.~Panda and D.~P. Mukherjee, ``Compositional zero-shot learning using multi-branch graph convolution and cross-layer knowledge sharing,'' \emph{PR}, vol. 145, p. 109916, 2024.

\bibitem{li2020symmetry}
Y.-L. Li, Y.~Xu, X.~Mao, and C.~Lu, ``Symmetry and group in attribute-object compositions,'' in \emph{CVPR}, 2020, pp. 11\,316--11\,325.

\bibitem{wang2023learning}
Q.~Wang, L.~Liu, C.~Jing, H.~Chen, G.~Liang, P.~Wang, and C.~Shen, ``Learning conditional attributes for compositional zero-shot learning,'' in \emph{CVPR}, 2023, pp. 11\,197--11\,206.

\bibitem{liu2023simple}
Z.~Liu, Y.~Li, L.~Yao, X.~Chang, W.~Fang, X.~Wu, and A.~El~Saddik, ``Simple primitives with feasibility-and contextuality-dependence for open-world compositional zero-shot learning,'' \emph{IEEE TPAMI}, 2023.

\bibitem{huo2024procc}
F.~Huo, W.~Xu, S.~Guo, J.~Guo, H.~Wang, Z.~Liu, and X.~Lu, ``Procc: Progressive cross-primitive compatibility for open-world compositional zero-shot learning,'' in \emph{AAAI}, vol.~38, no.~11, 2024, pp. 12\,689--12\,697.

\bibitem{xu2021relation}
Z.~Xu, G.~Wang, Y.~Wong, and M.~S. Kankanhalli, ``Relation-aware compositional zero-shot learning for attribute-object pair recognition,'' \emph{IEEE TMM}, vol.~24, pp. 3652--3664, 2021.

\bibitem{khan2023learning}
M.~G. Z.~A. Khan, M.~F. Naeem, L.~Van~Gool, A.~Pagani, D.~Stricker, and M.~Z. Afzal, ``Learning attention propagation for compositional zero-shot learning,'' in \emph{WCACV}, 2023, pp. 3828--3837.

\bibitem{naeem2021learning}
M.~F. Naeem, Y.~Xian, F.~Tombari, and Z.~Akata, ``Learning graph embeddings for compositional zero-shot learning,'' in \emph{CVPR}, 2021, pp. 953--962.

\bibitem{anwaar2022leveraging}
M.~U. Anwaar, Z.~Pan, and M.~Kleinsteuber, ``On leveraging variational graph embeddings for open world compositional zero-shot learning,'' in \emph{ACM MM}, 2022, pp. 4645--4654.

\bibitem{xu2024gipcol}
G.~Xu, J.~Chai, and P.~Kordjamshidi, ``Gipcol: Graph-injected soft prompting for compositional zero-shot learning,'' in \emph{WCACV}, 2024, pp. 5774--5783.

\bibitem{jaiswal2020survey}
A.~Jaiswal, A.~R. Babu, M.~Z. Zadeh, D.~Banerjee, and F.~Makedon, ``A survey on contrastive self-supervised learning,'' \emph{Technologies}, vol.~9, no.~1, p.~2, 2020.

\bibitem{li2022siamese}
X.~Li, X.~Yang, K.~Wei, C.~Deng, and M.~Yang, ``Siamese contrastive embedding network for compositional zero-shot learning,'' in \emph{CVPR}, 2022, pp. 9326--9335.

\bibitem{yang2023dual}
Y.~Yang, R.~Pan, X.~Li, X.~Yang, and C.~Deng, ``Dual-stream contrastive learning for compositional zero-shot recognition,'' \emph{IEEE TMM}, vol.~26, pp. 1909--1919, 2023.

\bibitem{wang2019tafe}
X.~Wang, F.~Yu, R.~Wang, T.~Darrell, and J.~E. Gonzalez, ``Tafe-net: Task-aware feature embeddings for low shot learning,'' in \emph{CVPR}, 2019, pp. 1831--1840.

\bibitem{jiang2024mutual}
C.~Jiang, Q.~Ye, S.~Wang, Y.~Shen, Z.~Zhang, and H.~Zhang, ``Mutual balancing in state-object components for compositional zero-shot learning,'' \emph{PR}, vol. 152, p. 110451, 2024.

\bibitem{jiang2024revealing}
C.~Jiang and H.~Zhang, ``Revealing the proximate long-tail distribution in compositional zero-shot learning,'' in \emph{AAAI}, vol.~38, no.~3, 2024, pp. 2498--2506.

\bibitem{li2024context}
Y.~Li, Z.~Liu, H.~Chen, and L.~Yao, ``Context-based and diversity-driven specificity in compositional zero-shot learning,'' in \emph{CVPR}, 2024, pp. 17\,037--17\,046.

\bibitem{karthik2022kg}
S.~Karthik, M.~Mancini, and Z.~Akata, ``Kg-sp: Knowledge guided simple primitives for open world compositional zero-shot learning,'' in \emph{CVPR}, 2022, pp. 9336--9345.

\bibitem{hu2024dynamic}
X.~Hu and Z.~Wang, ``A dynamic learning method towards realistic compositional zero-shot learning,'' in \emph{AAAI}, vol.~38, no.~3, 2024, pp. 2265--2273.

\bibitem{zhang2024vision}
J.~Zhang, J.~Huang, S.~Jin, and S.~Lu, ``Vision-language models for vision tasks: A survey,'' \emph{IEEE TPAMI}, 2024.

\bibitem{cui2022contrastive}
Q.~Cui, B.~Zhou, Y.~Guo, W.~Yin, H.~Wu, O.~Yoshie, and Y.~Chen, ``Contrastive vision-language pre-training with limited resources,'' in \emph{ECCV}.\hskip 1em plus 0.5em minus 0.4em\relax Springer, 2022, pp. 236--253.

\bibitem{singh2022flava}
A.~Singh, R.~Hu, V.~Goswami, G.~Couairon, W.~Galuba, M.~Rohrbach, and D.~Kiela, ``Flava: A foundational language and vision alignment model,'' in \emph{CVPR}, 2022, pp. 15\,638--15\,650.

\bibitem{huang2023nlip}
R.~Huang, Y.~Long, J.~Han, H.~Xu, X.~Liang, C.~Xu, and X.~Liang, ``Nlip: Noise-robust language-image pre-training,'' in \emph{AAAI}, vol.~37, no.~1, 2023, pp. 926--934.

\bibitem{esmaeilpour2022zero}
S.~Esmaeilpour, B.~Liu, E.~Robertson, and L.~Shu, ``Zero-shot out-of-distribution detection based on the pre-trained model clip,'' in \emph{AAAI}, vol.~36, no.~6, 2022, pp. 6568--6576.

\bibitem{wang2023clipn}
H.~Wang, Y.~Li, H.~Yao, and X.~Li, ``Clipn for zero-shot ood detection: Teaching clip to say no,'' in \emph{ICCV}, 2023, pp. 1802--1812.

\bibitem{shu2022test}
M.~Shu, W.~Nie, D.-A. Huang, Z.~Yu, T.~Goldstein, A.~Anandkumar, and C.~Xiao, ``Test-time prompt tuning for zero-shot generalization in vision-language models,'' \emph{NeurIPS}, vol.~35, pp. 14\,274--14\,289, 2022.

\bibitem{brown2020language}
T.~Brown, B.~Mann, N.~Ryder, M.~Subbiah, J.~D. Kaplan, P.~Dhariwal, A.~Neelakantan, P.~Shyam, G.~Sastry, A.~Askell \emph{et~al.}, ``Language models are few-shot learners,'' \emph{NeurIPS}, vol.~33, pp. 1877--1901, 2020.

\bibitem{pratt2023does}
S.~Pratt, I.~Covert, R.~Liu, and A.~Farhadi, ``What does a platypus look like? generating customized prompts for zero-shot image classification,'' in \emph{ICCV}, 2023, pp. 15\,691--15\,701.

\bibitem{novack2023chils}
Z.~Novack, J.~McAuley, Z.~C. Lipton, and S.~Garg, ``Chils: Zero-shot image classification with hierarchical label sets,'' in \emph{ICML}.\hskip 1em plus 0.5em minus 0.4em\relax PMLR, 2023, pp. 26\,342--26\,362.

\bibitem{chendon}
Z.~Chen, ``Don't paint everyone with the same brush: Adaptive prompt prototype learning for vision-language models.''

\bibitem{udandarao2023sus}
V.~Udandarao, A.~Gupta, and S.~Albanie, ``Sus-x: Training-free name-only transfer of vision-language models,'' in \emph{ICCV}, 2023, pp. 2725--2736.

\bibitem{rombach2022high}
R.~Rombach, A.~Blattmann, D.~Lorenz, P.~Esser, and B.~Ommer, ``High-resolution image synthesis with latent diffusion models,'' in \emph{CVPR}, 2022, pp. 10\,684--10\,695.

\bibitem{feng2023diverse}
C.-M. Feng, K.~Yu, Y.~Liu, S.~Khan, and W.~Zuo, ``Diverse data augmentation with diffusions for effective test-time prompt tuning,'' in \emph{ICCV}, 2023, pp. 2704--2714.

\bibitem{abdul2024align}
J.~Abdul~Samadh, M.~H. Gani, N.~Hussein, M.~U. Khattak, M.~M. Naseer, F.~Shahbaz~Khan, and S.~H. Khan, ``Align your prompts: Test-time prompting with distribution alignment for zero-shot generalization,'' \emph{NeurIPS}, vol.~36, 2024.

\bibitem{ma2024swapprompt}
X.~Ma, J.~Zhang, S.~Guo, and W.~Xu, ``Swapprompt: Test-time prompt adaptation for vision-language models,'' \emph{NeurIPS}, vol.~36, 2024.

\bibitem{liu2024dart}
Z.~Liu, H.~Sun, Y.~Peng, and J.~Zhou, ``Dart: Dual-modal adaptive online prompting and knowledge retention for test-time adaptation,'' in \emph{AAAI}, vol.~38, no.~13, 2024, pp. 14\,106--14\,114.

\bibitem{zhang2024dual}
Y.~Zhang, W.~Zhu, H.~Tang, Z.~Ma, K.~Zhou, and L.~Zhang, ``Dual memory networks: A versatile adaptation approach for vision-language models,'' in \emph{CVPR}, 2024, pp. 28\,718--28\,728.

\bibitem{karmanov2024efficient}
A.~Karmanov, D.~Guan, S.~Lu, A.~El~Saddik, and E.~Xing, ``Efficient test-time adaptation of vision-language models,'' in \emph{CVPR}, 2024, pp. 14\,162--14\,171.

\bibitem{zhang2024robust}
D.-C. Zhang, Z.~Zhou, and Y.-F. Li, ``Robust test-time adaptation for zero-shot prompt tuning,'' in \emph{AAAI}, vol.~38, no.~15, 2024, pp. 16\,714--16\,722.

\bibitem{zanella2024test}
M.~Zanella and I.~Ben~Ayed, ``On the test-time zero-shot generalization of vision-language models: Do we really need prompt learning?'' in \emph{CVPR}, 2024, pp. 23\,783--23\,793.

\bibitem{huang2022unsupervised}
T.~Huang, J.~Chu, and F.~Wei, ``Unsupervised prompt learning for vision-language models,'' \emph{arXiv}, 2022.

\bibitem{li2022masked}
J.~Li, S.~Savarese, and S.~C. Hoi, ``Masked unsupervised self-training for label-free image classification,'' \emph{arXiv}, 2022.

\bibitem{qian2024intra}
Q.~Qian, Y.~Xu, and J.~Hu, ``Intra-modal proxy learning for zero-shot visual categorization with clip,'' \emph{NeurIPS}, vol.~36, 2024.

\bibitem{mirza2024lafter}
M.~J. Mirza, L.~Karlinsky, W.~Lin, H.~Possegger, M.~Kozinski, R.~Feris, and H.~Bischof, ``Lafter: Label-free tuning of zero-shot classifier using language and unlabeled image collections,'' \emph{NeurIPS}, vol.~36, 2024.

\bibitem{kalantidis2024label}
Y.~Kalantidis, G.~Tolias \emph{et~al.}, ``Label propagation for zero-shot classification with vision-language models,'' in \emph{CVPR}, 2024, pp. 23\,209--23\,218.

\bibitem{kahana2022improving}
J.~Kahana, N.~Cohen, and Y.~Hoshen, ``Improving zero-shot models with label distribution priors,'' \emph{arXiv}, 2022.

\bibitem{nilsback2008automated}
M.-E. Nilsback and A.~Zisserman, ``Automated flower classification over a large number of classes,'' in \emph{ICVGIP}.\hskip 1em plus 0.5em minus 0.4em\relax IEEE, 2008, pp. 722--729.

\bibitem{van2015building}
G.~Van~Horn, S.~Branson, R.~Farrell, S.~Haber, J.~Barry, P.~Ipeirotis, P.~Perona, and S.~Belongie, ``Building a bird recognition app and large scale dataset with citizen scientists: The fine print in fine-grained dataset collection,'' in \emph{CVPR}, 2015, pp. 595--604.

\bibitem{lampert2013attribute}
C.~H. Lampert, H.~Nickisch, and S.~Harmeling, ``Attribute-based classification for zero-shot visual object categorization,'' \emph{IEEE TPAMI}, vol.~36, no.~3, pp. 453--465, 2013.

\bibitem{farhadi2009describing}
A.~Farhadi, I.~Endres, D.~Hoiem, and D.~Forsyth, ``Describing objects by their attributes,'' in \emph{CVPR}.\hskip 1em plus 0.5em minus 0.4em\relax IEEE, 2009, pp. 1778--1785.

\bibitem{yu2014fine}
A.~Yu and K.~Grauman, ``Fine-grained visual comparisons with local learning,'' in \emph{CVPR}, 2014, pp. 192--199.

\bibitem{isola2015discovering}
P.~Isola, J.~J. Lim, and E.~H. Adelson, ``Discovering states and transformations in image collections,'' in \emph{CVPR}, 2015, pp. 1383--1391.

\bibitem{Saini_2022_CVPR}
N.~Saini, K.~Pham, and A.~Shrivastava, ``Disentangling visual embeddings for attributes and objects,'' in \emph{CVPR}, June 2022, pp. 13\,658--13\,667.

\bibitem{fei2004learning}
L.~Fei-Fei, R.~Fergus, and P.~Perona, ``Learning generative visual models from few training examples: An incremental bayesian approach tested on 101 object categories,'' in \emph{CVPRW}.\hskip 1em plus 0.5em minus 0.4em\relax IEEE, 2004, pp. 178--178.

\bibitem{griffin2007caltech}
G.~Griffin, A.~Holub, P.~Perona \emph{et~al.}, ``Caltech-256 object category dataset,'' Technical Report 7694, California Institute of Technology Pasadena, Tech. Rep., 2007.

\bibitem{xiao2010sun}
J.~Xiao, J.~Hays, K.~A. Ehinger, A.~Oliva, and A.~Torralba, ``Sun database: Large-scale scene recognition from abbey to zoo,'' in \emph{CVPR}.\hskip 1em plus 0.5em minus 0.4em\relax IEEE, 2010, pp. 3485--3492.

\bibitem{bossard2014food}
L.~Bossard, M.~Guillaumin, and L.~Van~Gool, ``Food-101--mining discriminative components with random forests,'' in \emph{ECCV}.\hskip 1em plus 0.5em minus 0.4em\relax Springer, 2014, pp. 446--461.

\bibitem{krause2013collecting}
J.~Krause, J.~Deng, M.~Stark, and L.~Fei-Fei, ``Collecting a large-scale dataset of fine-grained cars,'' 2013.

\bibitem{maji2013fine}
S.~Maji, E.~Rahtu, J.~Kannala, M.~Blaschko, and A.~Vedaldi, ``Fine-grained visual classification of aircraft,'' \emph{arXiv}, 2013.

\bibitem{parkhi2012cats}
O.~M. Parkhi, A.~Vedaldi, A.~Zisserman, and C.~Jawahar, ``Cats and dogs,'' in \emph{CVPR}.\hskip 1em plus 0.5em minus 0.4em\relax IEEE, 2012, pp. 3498--3505.

\bibitem{cimpoi2014describing}
M.~Cimpoi, S.~Maji, I.~Kokkinos, S.~Mohamed, and A.~Vedaldi, ``Describing textures in the wild,'' in \emph{CVPR}, 2014, pp. 3606--3613.

\bibitem{helber2019eurosat}
P.~Helber, B.~Bischke, A.~Dengel, and D.~Borth, ``Eurosat: A novel dataset and deep learning benchmark for land use and land cover classification,'' \emph{IEEE JSTAEORS}, vol.~12, no.~7, pp. 2217--2226, 2019.

\bibitem{krizhevsky2009learning}
A.~Krizhevsky, G.~Hinton \emph{et~al.}, ``Learning multiple layers of features from tiny images,'' 2009.

\bibitem{berg2014birdsnap}
T.~Berg, J.~Liu, S.~Woo~Lee, M.~L. Alexander, D.~W. Jacobs, and P.~N. Belhumeur, ``Birdsnap: Large-scale fine-grained visual categorization of birds,'' in \emph{CVPR}, 2014, pp. 2011--2018.

\bibitem{chen2023explanatory}
X.~Chen, X.~Deng, Y.~Lan, Y.~Long, J.~Weng, Z.~Liu, and Q.~Tian, ``Explanatory object part aggregation for zero-shot learning,'' \emph{IEEE TPAMI}, 2023.

\bibitem{wang2022language}
Z.~Wang, Y.~Gou, J.~Li, L.~Zhu, and H.~T. Shen, ``Language-augmented pixel embedding for generalized zero-shot learning,'' \emph{IEEE TCSVT}, vol.~33, no.~3, pp. 1019--1030, 2022.

\bibitem{cheng2023hybrid}
D.~Cheng, G.~Wang, B.~Wang, Q.~Zhang, J.~Han, and D.~Zhang, ``Hybrid routing transformer for zero-shot learning,'' \emph{PR}, vol. 137, p. 109270, 2023.

\bibitem{li2023distilled}
Y.~Li, Z.~Liu, S.~Jha, and L.~Yao, ``Distilled reverse attention network for open-world compositional zero-shot learning,'' in \emph{ICCV}, 2023, pp. 1782--1791.

\bibitem{he2016deep}
K.~He, X.~Zhang, S.~Ren, and J.~Sun, ``Deep residual learning for image recognition,'' in \emph{CVPR}, 2016, pp. 770--778.

\bibitem{simonyan2014very}
K.~Simonyan and A.~Zisserman, ``Very deep convolutional networks for large-scale image recognition,'' \emph{arXiv}, 2014.

\bibitem{szegedy2015going}
C.~Szegedy, W.~Liu, Y.~Jia, P.~Sermanet, S.~Reed, D.~Anguelov, D.~Erhan, V.~Vanhoucke, and A.~Rabinovich, ``Going deeper with convolutions,'' in \emph{CVPR}, 2015, pp. 1--9.

\bibitem{krizhevsky2012imagenet}
A.~Krizhevsky, I.~Sutskever, and G.~E. Hinton, ``Imagenet classification with deep convolutional neural networks,'' \emph{NeurIPS}, vol.~25, 2012.

\bibitem{hou2024large}
Y.~Hou, J.~Zhang, Z.~Lin, H.~Lu, R.~Xie, J.~McAuley, and W.~X. Zhao, ``Large language models are zero-shot rankers for recommender systems,'' in \emph{ECIR}.\hskip 1em plus 0.5em minus 0.4em\relax Springer, 2024, pp. 364--381.

\bibitem{kojima2022large}
T.~Kojima, S.~S. Gu, M.~Reid, Y.~Matsuo, and Y.~Iwasawa, ``Large language models are zero-shot reasoners,'' \emph{NeurIPS}, vol.~35, pp. 22\,199--22\,213, 2022.

\bibitem{wei2021finetuned}
J.~Wei, M.~Bosma, V.~Y. Zhao, K.~Guu, A.~W. Yu, B.~Lester, N.~Du, A.~M. Dai, and Q.~V. Le, ``Finetuned language models are zero-shot learners,'' \emph{arXiv}, 2021.

\bibitem{johnson2017google}
M.~Johnson, M.~Schuster, Q.~V. Le, M.~Krikun, Y.~Wu, Z.~Chen, N.~Thorat, F.~Vi{\'e}gas, M.~Wattenberg, G.~Corrado \emph{et~al.}, ``Google’s multilingual neural machine translation system: Enabling zero-shot translation,'' \emph{TACL}, vol.~5, pp. 339--351, 2017.

\bibitem{baek2023knowledge}
J.~Baek, A.~F. Aji, and A.~Saffari, ``Knowledge-augmented language model prompting for zero-shot knowledge graph question answering,'' \emph{arXiv}, 2023.

\bibitem{kumar2023zero}
P.~Kumar, K.~Pathania, and B.~Raman, ``Zero-shot learning based cross-lingual sentiment analysis for sanskrit text with insufficient labeled data,'' \emph{Applied Intelligence}, vol.~53, no.~9, pp. 10\,096--10\,113, 2023.

\bibitem{chen2021zero}
Q.~Chen, W.~Wang, K.~Huang, and F.~Coenen, ``Zero-shot text classification via knowledge graph embedding for social media data,'' \emph{IEEE IoT-J}, vol.~9, no.~12, pp. 9205--9213, 2021.

\bibitem{bari2020zero}
M.~S. Bari, S.~Joty, and P.~Jwalapuram, ``Zero-resource cross-lingual named entity recognition,'' in \emph{AAAI}, vol.~34, no.~05, 2020, pp. 7415--7423.

\bibitem{thakur2021beir}
N.~Thakur, N.~Reimers, A.~R{\"u}ckl{\'e}, A.~Srivastava, and I.~Gurevych, ``Beir: A heterogenous benchmark for zero-shot evaluation of information retrieval models,'' \emph{arXiv}, 2021.

\bibitem{huynh2020shared}
D.~Huynh and E.~Elhamifar, ``A shared multi-attention framework for multi-label zero-shot learning,'' in \emph{CVPR}, 2020, pp. 8776--8786.

\bibitem{ren2015faster}
S.~Ren, K.~He, R.~Girshick, and J.~Sun, ``Faster r-cnn: Towards real-time object detection with region proposal networks,'' \emph{NeurIPS}, vol.~28, 2015.

\bibitem{liu2016ssd}
W.~Liu, D.~Anguelov, D.~Erhan, C.~Szegedy, S.~Reed, C.-Y. Fu, and A.~C. Berg, ``Ssd: Single shot multibox detector,'' in \emph{ECCV}.\hskip 1em plus 0.5em minus 0.4em\relax Springer, 2016, pp. 21--37.

\bibitem{long2015fully}
J.~Long, E.~Shelhamer, and T.~Darrell, ``Fully convolutional networks for semantic segmentation,'' in \emph{CVPR}, 2015, pp. 3431--3440.

\bibitem{kirillov2023segment}
A.~Kirillov, E.~Mintun, N.~Ravi, H.~Mao, C.~Rolland, L.~Gustafson, T.~Xiao, S.~Whitehead, A.~C. Berg, W.-Y. Lo \emph{et~al.}, ``Segment anything,'' in \emph{ICCV}, 2023, pp. 4015--4026.

\bibitem{bansal2018zero}
A.~Bansal, K.~Sikka, G.~Sharma, R.~Chellappa, and A.~Divakaran, ``Zero-shot object detection,'' in \emph{ECCV}, 2018, pp. 384--400.

\bibitem{rahman2018zero}
S.~Rahman, S.~Khan, and F.~Porikli, ``Zero-shot object detection: Learning to simultaneously recognize and localize novel concepts,'' in \emph{ACCV}.\hskip 1em plus 0.5em minus 0.4em\relax Springer, 2018, pp. 547--563.

\bibitem{li2019zero}
Z.~Li, L.~Yao, X.~Zhang, X.~Wang, S.~Kanhere, and H.~Zhang, ``Zero-shot object detection with textual descriptions,'' in \emph{AAAI}, vol.~33, no.~01, 2019, pp. 8690--8697.

\bibitem{gu2020context}
Z.~Gu, S.~Zhou, L.~Niu, Z.~Zhao, and L.~Zhang, ``Context-aware feature generation for zero-shot semantic segmentation,'' in \emph{ACM MM}, 2020, pp. 1921--1929.

\bibitem{huynh2022open}
D.~Huynh, J.~Kuen, Z.~Lin, J.~Gu, and E.~Elhamifar, ``Open-vocabulary instance segmentation via robust cross-modal pseudo-labeling,'' in \emph{CVPR}, 2022, pp. 7020--7031.

\bibitem{he2023primitive}
S.~He, H.~Ding, and W.~Jiang, ``Primitive generation and semantic-related alignment for universal zero-shot segmentation,'' in \emph{CVPR}, 2023, pp. 11\,238--11\,247.

\bibitem{guo2020deep}
Y.~Guo, H.~Wang, Q.~Hu, H.~Liu, L.~Liu, and M.~Bennamoun, ``Deep learning for 3d point clouds: A survey,'' \emph{IEEE TPAMI}, vol.~43, no.~12, pp. 4338--4364, 2020.

\bibitem{naeem20223d}
M.~F. Naeem, E.~P. {\"O}rnek, Y.~Xian, L.~Van~Gool, and F.~Tombari, ``3d compositional zero-shot learning with decompositional consensus,'' in \emph{ECCV}.\hskip 1em plus 0.5em minus 0.4em\relax Springer, 2022, pp. 713--730.

\bibitem{cheraghian2022zero}
A.~Cheraghian, S.~Rahman, T.~F. Chowdhury, D.~Campbell, and L.~Petersson, ``Zero-shot learning on 3d point cloud objects and beyond,'' \emph{IJCV}, vol. 130, no.~10, pp. 2364--2384, 2022.

\bibitem{zhang2023sam3d}
D.~Zhang, D.~Liang, H.~Yang, Z.~Zou, X.~Ye, Z.~Liu, and X.~Bai, ``Sam3d: Zero-shot 3d object detection via segment anything model,'' \emph{arXiv}, 2023.

\bibitem{abdelreheem2023satr}
A.~Abdelreheem, I.~Skorokhodov, M.~Ovsjanikov, and P.~Wonka, ``Satr: Zero-shot semantic segmentation of 3d shapes,'' in \emph{ICCV}, 2023, pp. 15\,166--15\,179.

\bibitem{liu2023zero}
R.~Liu, R.~Wu, B.~Van~Hoorick, P.~Tokmakov, S.~Zakharov, and C.~Vondrick, ``Zero-1-to-3: Zero-shot one image to 3d object,'' in \emph{ICCV}, 2023.

\bibitem{liu2023stylerf}
K.~Liu, F.~Zhan, Y.~Chen, J.~Zhang, Y.~Yu, A.~El~Saddik, S.~Lu, and E.~P. Xing, ``Stylerf: Zero-shot 3d style transfer of neural radiance fields,'' in \emph{CVPR}, 2023, pp. 8338--8348.

\bibitem{xu2023dream3d}
J.~Xu, X.~Wang, W.~Cheng, Y.-P. Cao, Y.~Shan, X.~Qie, and S.~Gao, ``Dream3d: Zero-shot text-to-3d synthesis using 3d shape prior and text-to-image diffusion models,'' in \emph{CVPR}, 2023, pp. 20\,908--20\,918.

\bibitem{li2024genzi}
L.~Li and A.~Dai, ``Genzi: Zero-shot 3d human-scene interaction generation,'' in \emph{CVPR}, 2024, pp. 20\,465--20\,474.

\bibitem{jiang2024back}
Z.~Jiang, Z.~Zhou, L.~Li, W.~Chai, C.-Y. Yang, and J.-N. Hwang, ``Back to optimization: Diffusion-based zero-shot 3d human pose estimation,'' in \emph{WCACV}, 2024, pp. 6142--6152.

\bibitem{abdelreheem2023zero}
A.~Abdelreheem, A.~Eldesokey, M.~Ovsjanikov, and P.~Wonka, ``Zero-shot 3d shape correspondence,'' in \emph{SIGGRAPH}, 2023, pp. 1--11.

\bibitem{javed2024cplip}
S.~Javed, A.~Mahmood, I.~I. Ganapathi, F.~A. Dharejo, N.~Werghi, and M.~Bennamoun, ``Cplip: Zero-shot learning for histopathology with comprehensive vision-language alignment,'' in \emph{CVPR}, 2024, pp. 11\,450--11\,459.

\bibitem{li2023rs}
X.~Li, C.~Wen, Y.~Hu, and N.~Zhou, ``Rs-clip: Zero shot remote sensing scene classification via contrastive vision-language supervision,'' \emph{IJAEOG}, vol. 124, p. 103497, 2023.

\bibitem{li2024zero}
A.~Li, C.~Qiu, M.~Kloft, P.~Smyth, M.~Rudolph, and S.~Mandt, ``Zero-shot anomaly detection via batch normalization,'' \emph{NeurIPS}, vol.~36, 2024.

\bibitem{chen2021elaborative}
S.~Chen and D.~Huang, ``Elaborative rehearsal for zero-shot action recognition,'' in \emph{ICCV}, 2021, pp. 13\,638--13\,647.

\bibitem{wang2024content}
S.~Wang, J.~Chang, Z.~Wang, H.~Li, W.~Ouyang, and Q.~Tian, ``Content-aware rectified activation for zero-shot fine-grained image retrieval,'' \emph{IEEE TPAMI}, 2024.

\bibitem{hong2023fine}
M.~Hong, X.~Zhang, G.~Li, and Q.~Huang, ``Fine-grained feature generation for generalized zero-shot video classification,'' \emph{IEEE TIP}, vol.~32, pp. 1599--1612, 2023.

\bibitem{wang2024comprehensive}
L.~Wang, X.~Zhang, H.~Su, and J.~Zhu, ``A comprehensive survey of continual learning: theory, method and application,'' \emph{IEEE TPAMI}, 2024.

\bibitem{zhou2021multi}
F.~Zhou, S.~Huang, B.~Liu, and D.~Yang, ``Multi-label image classification via category prototype compositional learning,'' \emph{IEEE TCSVT}, vol.~32, no.~7, pp. 4513--4525, 2021.

\bibitem{guo2023parsnets}
J.~Guo, Q.~Zhou, R.~Li, X.~Lu, Z.~Liu, J.~Chen, X.~Xie, and J.~Zhang, ``Parsnets: A parsimonious orthogonal and low-rank linear networks for zero-shot learning,'' \emph{IJCAI}, 2023.

\bibitem{rahman2018unified}
S.~Rahman, S.~Khan, and F.~Porikli, ``A unified approach for conventional zero-shot, generalized zero-shot, and few-shot learning,'' \emph{IEEE TIP}, vol.~27, no.~11, pp. 5652--5667, 2018.

\bibitem{shafiee2022zero}
N.~Shafiee and E.~Elhamifar, ``Zero-shot attribute attacks on fine-grained recognition models,'' in \emph{ECCV}.\hskip 1em plus 0.5em minus 0.4em\relax Springer, 2022, pp. 262--282.

\end{thebibliography}
\begin{IEEEbiography}
[{\includegraphics[width=1in,height=1.25in,clip,keepaspectratio]{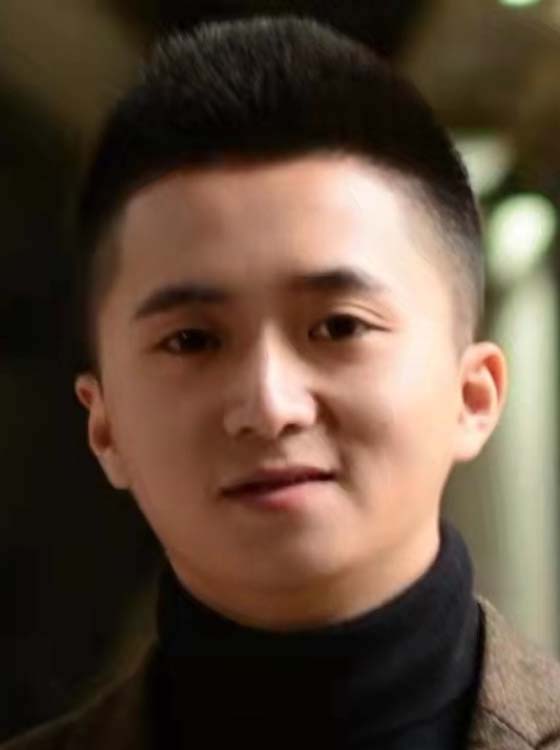}}]{Jingcai~Guo} (Member, IEEE) is currently a Research Assistant Professor with Department of Computing, The Hong Kong Polytechnic University, Hong Kong SAR, where he received his Ph.D. in Dec. 2020. Prior to that, he received his M.E. from Waseda University (2015), Japan, and his B.E. from Sichuan University (2013), China, all in Computer Science. He is interested in Low-shot AI, which targets learning/modeling with limited resources in terms of data, computing capability, and their derivative applications. Topics include zero/few-shot learning, representation learning, federated learning, and model compression. He is currently serving as Associate Editor for IEEE Open Journal of the Computer Society and Guest Editor for IEEE Transactions on Computational Social Systems. He has served as Area Chair, Senior PC, and Session Chair for prestigious conferences like ICML, ACM-MM, AAAI, IJCAI, ICME, and VTC.
\end{IEEEbiography}

\begin{IEEEbiography}[{\includegraphics[width=1in,height=1.25in,clip,keepaspectratio]{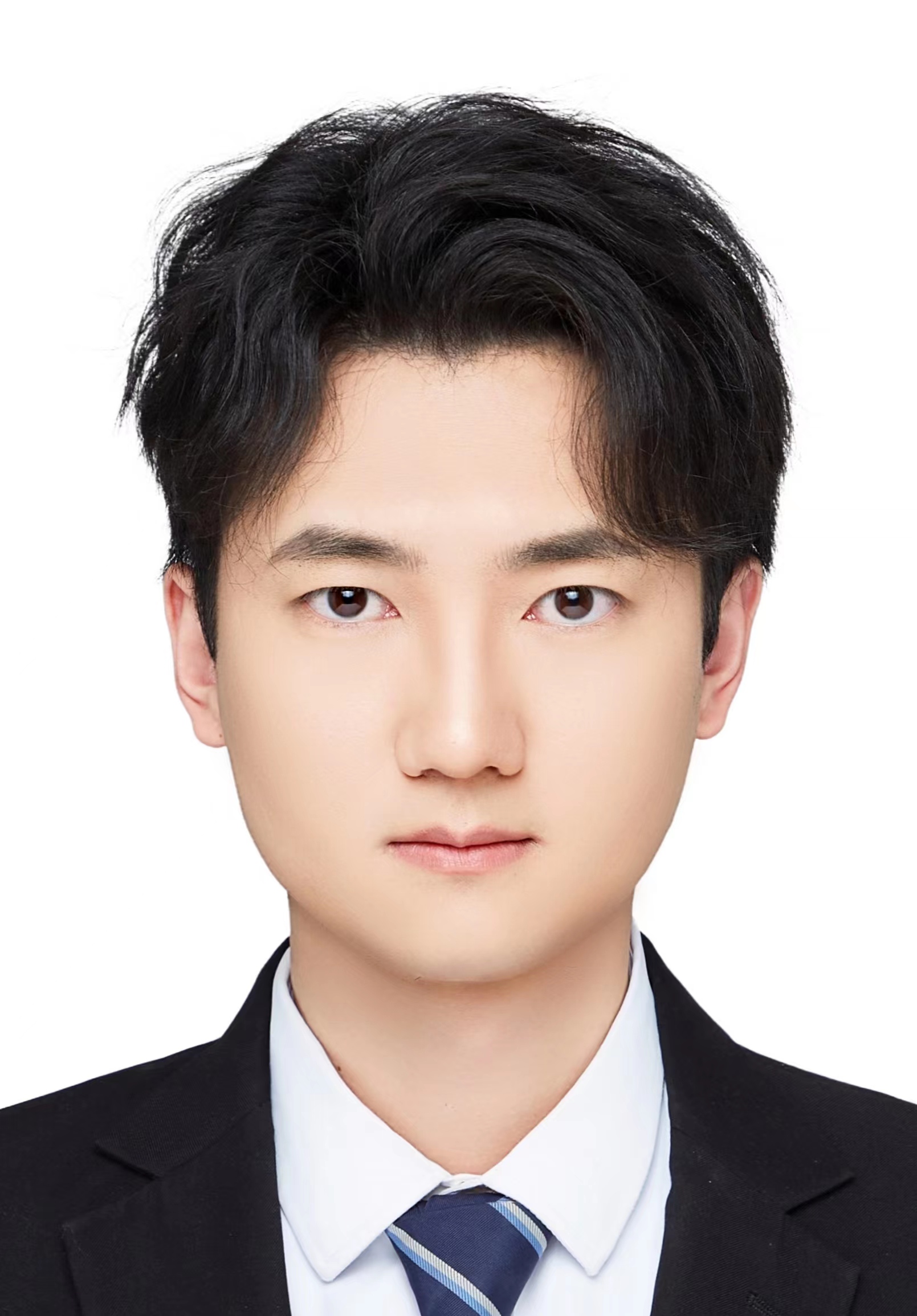}}]
{Zhijie Rao} (Student Member, IEEE) received the B.E. degree from Beijing University of Posts and Telecommunications, China, in 2019, and the M.S. degree from Xiamen University, China, in 2023. He is currently working towards the Ph.D. degree at Department of Computing, The Hong Kong Polytechnic University, Hong Kong SAR. His research interest spans transfer learning and edge AI, with a particular focus on zero/few-shot learning, domain adaptation/generalization, and deployment/training of foundation models.
\end{IEEEbiography}

\begin{IEEEbiography}[{\includegraphics[width=1in,height=1.25in,clip,keepaspectratio]{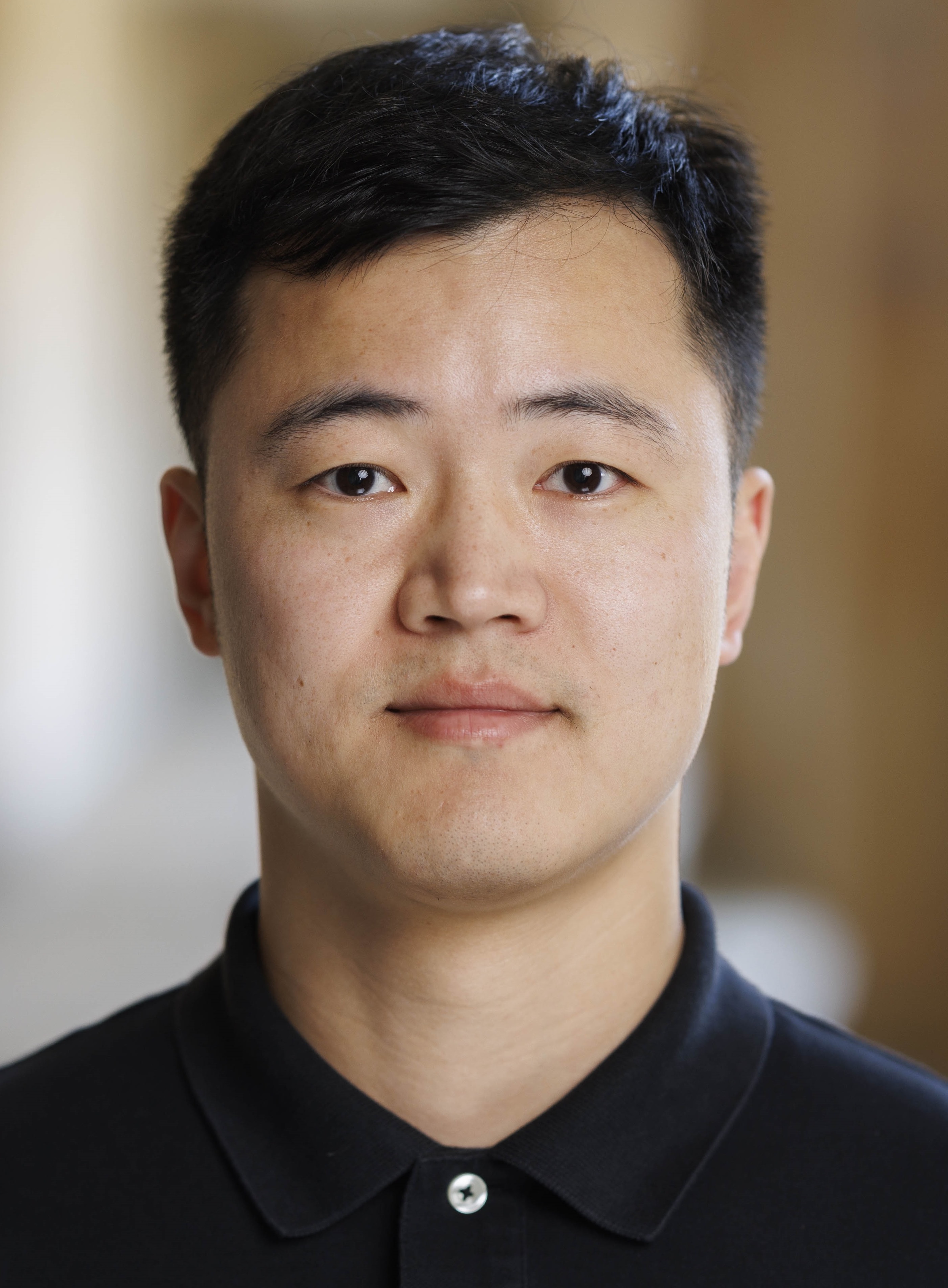}}]
{Zhi Chen} (Member, IEEE) received the M.S. and Ph.D. degrees from the University of Queensland, Australia, in 2018 and 2023, respectively. He is currently a Postdoctoral Research Fellow with the School of Electrical Engineering and Computer Science at the University of Queensland, Australia. His research interests include zero-shot learning, multimedia understanding, computer vision, and generative modeling.
\end{IEEEbiography}

\begin{IEEEbiography}[{\includegraphics[width=1in,height=1.25in,clip,keepaspectratio]{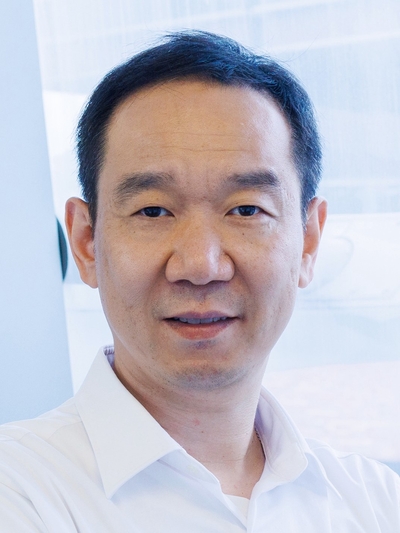}}]
{Song Guo} (Fellow, IEEE) is currently a Chair Professor in the Department of Computer Science and Engineering at the Hong Kong University of Science and Technology, Hong Kong SAR. His research interests are mainly in machine learning, edge AI, big data, mobile computing, and distributed systems. He is a Fellow of the Canadian Academy of Engineering, a Fellow of the IEEE, and a Member of Academia Europaea. He has served for IEEE Computer Society on Fellow Evaluation Committee, Transactions Operations Committee, Editor-in-Chief Search Committee, etc. He is the current Editor-in-Chief of IEEE Open Journal of the Computer Society and has been on the editorial board of various prestigious IEEE journals. He has also been chair of organizing and technical committees of numerous international conferences. Prof. Guo was the recipient of the 2020 Best Paper Award for IEEE Transactions on Computers, 2019 IEEE TCBD Best Conference Paper Award, 2018 IEEE TCGCC Best Magazine Paper Award, 2019 and 2017 IEEE Systems Journal Annual Best Paper Award, and other 8 Best Paper Awards from IEEE/ACM conferences. He was the recipient of 2024 Edward J. McCluskey Technical Achievement Award. 
\end{IEEEbiography}

\begin{IEEEbiography}[{\includegraphics[width=1in,height=1.25in,clip,keepaspectratio]{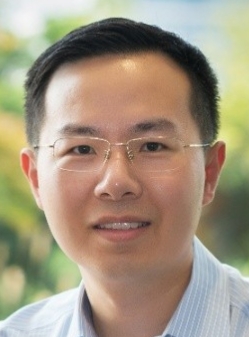}}]
{Jingren Zhou} (Fellow, IEEE) received the PhD degree in computer science from Columbia University. He is currently the senior vice president with Alibaba Group and Chief Technology Officer with Alibaba Cloud. He has authored or co-authored dozens of papers in top conferences and journals in his research areas, which include machine learning algorithm platforms, query processing, large-scale distributed systems, optimization of distributed databases, and data processing methods based on large-scale distributed systems, and holds several patented inventions of key technologies in the industry. He is a Fellow of IEEE.
\end{IEEEbiography}

\begin{IEEEbiography}[{\includegraphics[width=1in,height=1.25in,clip,keepaspectratio]{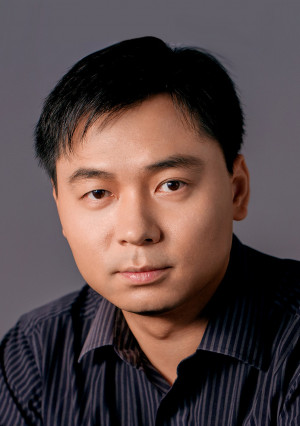}}]
{Dacheng Tao} (Fellow, IEEE) is currently a Distinguished University Professor in the College of Computing and Data Science at Nanyang Technological University, Singapore. He mainly applies statistics and mathematics to artificial intelligence and data science, and his research is detailed in one monograph and over 200 publications in prestigious journals and proceedings at leading conferences, with best paper awards, best student paper awards, and test-of-time awards. His publications have been cited over 131K times and have an h-index of 175 in Google Scholar. He received the 2015 and 2020 Australian Eureka Prize, the 2018 IEEE ICDM Research Contributions Award, and the 2021 IEEE Computer Society McCluskey Technical Achievement Award. He is a Fellow of the Australian Academy of Science, AAAS, ACM, and IEEE.
\end{IEEEbiography}

\end{document}